\documentclass[final,3p,times]{elsarticle}

\usepackage{amssymb}
\usepackage{rotating}
\usepackage{amsmath,amsfonts}
\usepackage{float}
\usepackage[ruled,lined]{algorithm2e}

\usepackage{graphicx}
\usepackage[export]{adjustbox}
\usepackage{textcomp}
\usepackage{xcolor}
\usepackage{mathtools}
\usepackage{array, multirow, booktabs}
\usepackage{makecell}
\usepackage{caption,subcaption}
\usepackage{url}
\usepackage{verbatim}
\usepackage{placeins}
\usepackage{booktabs} 
\usepackage{tabularx}
\usepackage{ltablex}
\usepackage{comment}
\usepackage{listings}
\usepackage{caption} 
\usepackage{hyperref}
\usepackage{subcaption}
\usepackage{placeins}
\usepackage{grffile}
\usepackage{xcolor,colortbl}
\usepackage{nicematrix}
\usepackage{soul} 
\usepackage[normalem]{ulem}
\usepackage[colorinlistoftodos, textwidth=2.5cm]{todonotes}

\usepackage{xspace}
\newcommand{\approach}{ExIFFI\xspace} 
 
\newcommand{\EIFplus}{\texorpdfstring{$\text{EIF}^+$}{EIFp}}
\newcommand{\EIFplusspace}{ \texorpdfstring{$\text{EIF}^+$\xspace}{EIFp}}

\newcommand{\davide}[1]{{\color{green} #1}}

\newcommand{\modifica}[1]{{\color{red} #1}}
\newcommand{\rebuttal}[1]{{\color{black} #1}}
\newcommand{\ale}[1]{{\color{black} #1}}

\journal{Engineering Applications of Artificial Intelligence}

\begin{document}

\begin{frontmatter}



\title{Enhancing Interpretability and Generalizability in Extended Isolation Forests}


\author[PD]{Alessio Arcudi} 
\author[PD]{Davide Frizzo}
\author[SDS]{Chiara Masiero}
\author[PD]{Gian Antonio Susto}

\affiliation[PD]{organization={University of Padova},
            country={Italy}}
\affiliation[SDS]{organization={Statwolf Data Science Srl},
            country={Italy}}

\begin{abstract}
Anomaly Detection (AD) involves identifying unusual behaviours within complex datasets and systems. While Machine Learning algorithms and Decision Support Systems (DSSs) offer effective solutions for this task, simply pinpointing anomalies \ale{may prove insufficient in real-world applications, particularly in engineering settings where maintenance and diagnostics are crucial. Users require insights into the rationale behind predictions to facilitate root cause analysis and foster trust in the model. However, the unsupervised nature of AD presents a challenge in developing interpretable tools.} \ale{This paper addresses this challenge by introducing Extended Isolation Forest Feature Importance (\approach), a novel interpretability approach specifically designed to explain the predictions made by Extended Isolation Forest (EIF), which works for all the Isolation Forest models with branching based on hyperplanes. \approach leverages feature importance to provide explanations at both global and local levels. This work also introduces Enhanced Extended Isolation Forest (\EIFplus), an enhanced variant of EIF, conceived to improve its generalization capabilities in the detection of unseen anomalies through a different splitting hyperplanes design strategy. We conduct a comprehensive comparative analysis using five synthetic and eleven real-world datasets to evaluate various unsupervised anomaly detection (AD) methods, using the Average Precision metric. Our proposed method, \EIFplus, consistently outperforms EIF across all datasets when trained without anomalies, demonstrating its superior generalization capabilities. To proof the effectiveness of the interpretation, we introduce a novel metric, the Area Under the Curve of the Feature Selection ($AUC_{FS}$), employing feature selection as a proxy task to assess performance. \approach proves highly effective, outperforming other unsupervised interpretation methods in 8 out of 11 real-world datasets and accurately identifying anomalous features in synthetic datasets. When trained only on inliers, \approach surpasses all other models on real-world datasets and successfully detects the correct anomalous features of the synthetic datasets. Lastly, we contribute to the research community by providing open-source code, promoting further exploration and reproducibility.}
\end{abstract}




\end{frontmatter}



\section{Introduction} \label{sec:intro}

\rebuttal{Machine Learning (ML) and Artificial Intelligence (AI) play a central role in the ongoing socio-economic change, revolutionizing various sectors such as manufacturing \cite{krafft2020challenges, xu2018fourth, industry5.0, C_RISE_survey_AD}, medicine \cite{houssein2021deep}, smart farming \cite{AIT,corn} and the Internet of Things \cite{klaib2021eye, AIoT}. With the increasing deployment of ML across various industries, new challenges have arisen, particularly due to the complexity and lack of transparency in these systems.} 

\rebuttal{Moreover, the diversity of end users, many of whom may not have expertise in data-driven methods, further complicates the effective use of these technologies. To address this, it is crucial to develop algorithms that can explain the structure and predictions of ML models, making them accessible and understandable to a broader audience.} 
Many scientific works identify explainability\footnote{While authors in the literature use the term 'interpretability' and 'explainability' associated with slightly different concepts when associated to Machine Learning/Artificial Intelligence as presented in \cite{gilpin2018explaining}, in this work we will use both terms interchangeably.} as a key factor to enable the successful adoption of ML-based systems \cite{confalonieri2021historical, doshi2017towards, linardatos2020explainable}. 

\rebuttal{For tabular data, several methods are available to explain a model's predictions \cite{molnar2020interpretable}}. One common approach is to calculate the importance of each feature in the model's predictions. This involves evaluating the influence of each feature on both individual predictions (referred to as ``Local Importance") and the overall dataset (``Global Importance"). \rebuttal{By evaluating feature importance, users gain a deeper understanding of how the model utilizes input data to generate predictions.}

Despite the remarkable recent advancements in eXplainable Artificial Intelligence (XAI), most approaches are designed for supervised tasks, leaving unsupervised tasks, like Anomaly Detection (AD), rarely discussed in the literature. 

AD, also referred to as Outlier Detection\footnote{In this paper, we will refer to 'Outlier Detection' and 'Anomaly Detection' alternatively, always referring to the same unsupervised task of finding anomalous data points.}, is a field of ML 
that focuses on identifying elements that live outside 
the standard ``normal" behavior observed in the majority of the dataset \cite{hawkins1980identification}. 

\rebuttal{Explainability is crucial for Anomaly Detection (AD) approaches, particularly in industrial settings. For instance, AD methods used to monitor industrial machinery can significantly benefit from clear interpretations of detected anomalies. Understanding the root cause of an anomaly can help reduce machine failures, minimize energy loss, conserve resources, and lower production costs. Additionally, the need for interpretable algorithms extends beyond technical benefits, lack of explanations can undermine trust in the model’s output, especially in high-stakes decision-making scenarios. In such cases, explainability is not just a preference but a legal requirement \cite{EuropeanCommission2020}.}

\rebuttal{Isolation Forest (IF) \cite{4781136} is a popular Anomaly Detection (AD) method due to its high accuracy, low computational cost, and relatively simple internal mechanism. It operates by recursively splitting the feature space using randomly chosen axis-aligned hyper-planes, allowing it to isolate anomalies with minimal partitions. To enhance the interpretability of IF, specialized approaches like Depth-based Isolation Forest Feature Importance (DIFFI) \cite{diffi} have been developed. DIFFI takes advantage of IF’s internal structure to provide both global and local explanations of the model’s decisions.}



\rebuttal{However, the one-dimensional partitioning process used by IF can introduce artifacts that degrade anomaly detection and reduce the quality of feature explanations. To address this limitation, the Extended Isolation Forest (EIF) has been proposed \cite{8888179}. EIF improves upon IF by utilizing oblique partitions, which help eliminate these artifacts and enhance both anomaly detection and feature explanation.} 

\rebuttal{According to the literature, the EIF is among the top-performing unsupervised anomaly detection (AD) methods \cite{bouman2023unsupervised}. However, EIF lacks built-in interpretability for features and struggles with robustness when encountering novel anomalies. To address these limitations and encourage the adoption of isolation-based unsupervised AD approaches, this paper makes three key contributions:}
\begin{enumerate}
    \item We propose the Extended Isolation Forest Feature Importance (\approach), the first (to the best of our knowledge) model-specific approach which generalize the explanations to both IF models and the extended version EIF. 
    \item We present \EIFplus, a refined version of the EIF designed to optimally model the space around the training data distribution, enhancing the algorithm’s performance on unseen data. This innovation aims to bolster the model’s generalization ability, reliably identifying anomalies even without prior knowledge of their potential locations or without being present in the training dataset. \approach applies to \EIFplus, too.
    \item \rebuttal{We benchmark the novel \EIFplus and \approach against state-of-the-art isolation based AD and explainability approaches on 16 public datasets and showcase their effectiveness and computational efficiency. 
    To facilitate investigation and reproducibility, we introduce a novel functionally-grounded quantitative evaluation, named $AUC_{FS}$ score, that measure the effectiveness of the interpretation exploiting Feature Selection as a proxy task.}
\end{enumerate}

\rebuttal{These contributions address critical gaps in the interpretability and robustness of the EIF, which are essential to bridge for its broader acceptance and adoption over the traditional IF, particularly given EIF's superior performance, enhanced robustness, and reduced bias. Moreover, this work paves the way for the wider adoption and application of feature importance methods across various domains by establishing a clear and systematic framework for comparing feature importance explainability approaches.}

\rebuttal{The paper is organized as follows. Section \ref{sec:related_work} surveys relevant research. Section \ref{IF} provides an introduction to the IF algorithm, followed by a discussion of the EIF algorithm in Section \ref{ExtendedIF}. The newly proposed \EIFplus model is introduced in Section \ref{sec:EIFplus}. Sections \ref{DIFFI} and \ref{ExIFFI} present the DIFFI and \approach interpretation algorithms, along with the graphical tools used to illustrate their results.

Section \ref{sec:experimental_results} describes the experimental setup for evaluating \EIFplus and \approach analyzing anomaly detection and interpretability performances. Section \ref{sec:Evaluation} details the experiments, including graphical evaluations of three datasets \rebuttal{in Sections \ref{sec:syn_perf}, \ref{sec:real_perf}}, while Section \ref{sec:perf-tab} presents the numerical AD performance results for 16 datasets. Section \ref{sec:imp-tab} covers the interpretation results, \rebuttal{Section \ref{sec:corr-tab} analyses the correlation between interpretation and AD models results,} and Section \ref{sec:time_scaling} provides an ablation study on scalability for large datasets.}

Finally, limitations, conclusions, and future research directions are discussed in Section \ref{sec:conclusions}. 

The Appendix includes an extensive description of datasets \ref{sec:appendix_datasets}, an analysis comparing EIF and IF \ref{sec:IFvsEIF}, \rebuttal{an analysis providing a deep explanation on why \approach is not depth-based as DIFFI \ref{sec:non-depth},} an ablation study on the \EIFplus hyperparameter $\eta$ \rebuttal{\ref{sec:eta}} \rebuttal{, and a comparison of different interpretations through the Normalized Discounted Cumulative Gain (NDCG) metric \ref{sec:ndcg_exp}}.

\section{Related Work}\label{sec:related_work}

\rebuttal{Root Cause Analysis (RCA) is essential for identifying the underlying causes of anomalies in model behavior, enhancing our understanding of system outputs. Papageorgiou et al. \cite{papageorgiou2022systematic} categorize RCA methods into two main types: probabilistic and deterministic.

Probabilistic RCA methods, such as Bayesian Networks \cite{kitson2023survey}, model relationships between variables to infer causal connections \cite{amin2021data}. These approaches are particularly effective in fields like medical diagnostics and risk assessment, where they can trace the origins of specific events by representing probabilistic dependencies {\cite{C_RISE_fault_isolation,C_RISE_dynamic_bayes_net,C_RISE_bayesian_net,C_RISE_bayes_net}}. However, they often require substantial domain expertise, are computationally intensive, and may struggle with scalability in large datasets with complex interdependencies.

In contrast, Deterministic RCA methods focus on evaluating the importance of individual features in contributing to specific outcomes without modeling the entire system. These methods are divided into ad-hoc and post-hoc approaches \cite{molnar2020interpretable}. For example, ad-hoc methods like the feature importance measure in Random Forests \cite{kursa2010feature} are straightforward to implement and interpret but may lack generalizability due to their dependence on specific models. Post-hoc methods, such as SHAP \cite{NIPS2017_7062} and permutation importance \cite{altmann2010permutation}, offer model-agnostic interpretability, though they often come at the cost of higher computational demands and do not inherently provide causal insights.

Thus, the choice between probabilistic and deterministic RCA methods hinges on the specific needs of the problem at hand. Probabilistic methods are suited for uncovering causal relationships but are more complex and resource-intensive, while deterministic methods offer greater flexibility and ease of use, especially when a model-agnostic solution is required, though they may provide less depth in causal inference.

Explainability in Unsupervised Anomaly Detection (AD) remains a significant challenge, where Explainable Anomaly Detection (XAD) focuses on extracting relevant knowledge from AD models regarding relationships either present in the data or learned by the model. Recent surveys highlight the importance of interpretability in AD, particularly as it becomes a regulatory requirement in various applications \cite{XADSurvey}.

Unsupervised AD methods can be categorized into parametric and non-parametric approaches, each presenting unique advantages and challenges, especially in terms of explainability. Parametric models express relationships through a finite set of parameters, facilitating straightforward predictions. However, these models often lack clear interpretability, necessitating additional tools for explanation. For instance, the widely-used Anomaly Autoencoder (AE) model \cite{AutoEncoder} faces interpretability challenges, leading to the development of methods like RXP (Residual eXPlainer) \cite{RXP}, which provides explanations by analyzing deviations in reconstructed features.

On the other hand, Non-parametric AD approaches do not rely on predefined parametric functions and instead detect anomalies by leveraging the inherent structure and patterns in the data. This often results in intrinsic interpretability. For example, ECOD (Empirical-Cumulative-Distribution-based Outlier Detection) \cite{ECOD} derives explanations directly from data distributions, and IF is enhanced by the DIFFI algorithm \cite{diffi}, which uses partitioning processes to explain outliers. This intrinsic interpretability makes non-parametric methods particularly valuable when understanding the decision-making process is crucial, with DIFFI showing effectiveness in industrial applications \cite{DIFFI2, DIFFIBRITO} by providing both local and global explanations.

In conclusion, the selection between parametric and non-parametric unsupervised AD methods should be guided by application needs. Parametric models offer simpler predictions but often require supplementary tools for interpretability, whereas non-parametric models provide greater flexibility and inherent interpretability by utilizing data patterns directly, making them ideal for scenarios where understanding the decision-making process is critical.

This paper addresses a significant gap in the literature by introducing a novel interpretability algorithm specifically tailored for models based on the IF framework, including the Extended Isolation Forest (EIF) model. While existing methods like DIFFI offer some interpretability of IF model, they do not fully accommodate the unique structure of EIF. Our proposed algorithm enhances the interpretability of EIF and similar models, offering both local and global insights into anomaly detection processes. This contribution is crucial for advancing Explainable Anomaly Detection (XAD) by providing more accessible and effective interpretability solutions for IF-based models, particularly in complex, unsupervised settings.}

\color{black}


\section{Isolation-based Approaches for Anomaly Detection}\label{sec:isolation_based_approaches}
Next, we provide some notions about isolation-based approaches for AD. This family of methodologies, stemming from the Isolation Forest \cite{4781136}, identifies outliers as samples that can be easily separated from the others, i.e., through a reduced number of splitting hyperplanes.
\subsection{Isolation Forest} \label{IF} 
Isolation Forest is a widely used ML model for AD \cite{4781136}. It generates a set of $N$ random trees, called isolation trees, that are able to identify anomalous elements in a dataset based on their position in the tree structure. The idea behind this approach is that anomalies, on average, are located at the beginning of the trees because they are easier to separate from the rest of the dataset.

Assume that we have $n$ training data $\mathbf{X} = \{\mathbf{x}_1,\dots,\mathbf{x}_n\}$, where $\mathbf{x}_i \in \mathbb{R}^d$. The IF algorithm chooses at random one dimension $q \in \{1,\dots,d\}$ and a split value $p\in [\min_{i \in \{1,\dots,n\}} x_{i,q},\max_{i \in \{1,\dots,n\}} x_{i,q}] $. The dataset is then divided into two subsets, the left one $L=\{\mathbf{x_i}|x_{i,q}\leq p\}$ and the right one $R=\{\mathbf{x_i}|x_{i,q}> p\}$.
This procedure is calculated iteratively until the whole forest is built. Suppose the size of the dataset is excessive, meaning that the number of samples makes the construction of the trees too slow; in that case, it is demonstrated by Liu et al. \cite{4781136} that it is better to build the forest using only a random subsample $\tilde{X}$ with $\tilde{n}$ elements for each tree. Not only does this keep the computational cost low, it also improves IF's ability to identify anomalies clearly. Once the model has built an isolation forest, to determine which data points live outside the dataset distribution, the algorithm computes an anomaly score for each of them. This value is based on the average depth among trees where each data point is isolated.

The depth of a point $x$ in a tree $t$, denoted by $h_{t}(x)$, is the cardinality of the set of nodes $\mathcal{P}^{t}_x$ that it has to pass to reach the leaf node, i.e.:
\begin{equation}\label{setnodesx}
h_{t}(x) = |\mathcal{P}^{t}_x|\quad\text{where}\quad\mathcal{P}^{t}_x=\{k_{s_1},k_{s_2},\dots,k_{s_{h}}\},
\end{equation}

Let $E(h(x))$ be the mean value of depths reached among all trees for a single data point $x$. Then, according to \cite{preiss2000data}, the anomaly score is defined by the function:
\begin{equation}\label{anom_score_if}
    s(x,\tilde{n})=2^{-\frac{E(h(x))}{c(\tilde{n})}},
\end{equation}
where $c(N)$ is the normalizing factor defined as the average depth of an unsuccessful search in a Binary Search Tree \cite{8888179}, i.e.:
\begin{equation}\label{normalized_factor}
    c(N) = 2H(N-1)-\frac{2(N-1)}{N}.
\end{equation}
and $H(i)$ is the harmonic number that can be estimated by:
\begin{equation}
    H(i) = \ln(i) + 0.5772156649 \mbox{ (Euler constant)}. 
\end{equation} 
IF's fast execution with low memory requirement is a direct result of building partial models and requiring only a significantly small sample size as compared to the given training set. This capability is due to the fact IF has the main goal to quickly isolate anomalies more than modelling the normal distribution, contrarily to what other detection methods do \cite{9347460}.
\subsection{Extended Isolation Forest} \label{ExtendedIF} 
Although IF is one of the most popular and effective AD algorithms, it has some drawbacks due to its partition strategy. One of them is related to considering only hyper-planes that are orthogonal with respect to the directions of the feature space, as Hariri et al. show in \cite{8888179}.
In some cases, this bias leads to the formation of some artifacts where points are associated with low anomaly scores even if they are clearly anomalous. These areas are in the intersection of the hyperplanes orthogonal to the dimensions associated with the detection of inliers (Figure \ref{fig:scoremap 2 IF vs EIF}), creating misleading score maps and, in some cases, wrong predictions.

As a consequence, Hariri et al. in \cite{8888179}  tried to improve the AD algorithm by correcting this bias with a different and more general algorithm called Extended IF. Instead of selecting a hyper-plane orthogonal to a single input dimension chosen at random, they suggested picking a random point $\mathbf{p}$ and a random vector $\mathbf{v}$ that are consequently used to build a hyperplane that will split the space at each node of the binary tree into two smaller subspaces, as the IF does. Therefore, they proposed a more generic and extended paradigm, while maintaining the fast execution and low memory requirements of the original IF.

In the following, we will briefly describe the working principle of the EIF model.  Let's consider a set $X$ of $n$ elements $\mathbf{x}_i \in \mathbb{R}^d$ with $i \in \{1,\dots,n\}$.
In order to evaluate the anomaly score of the samples, the EIF model \cite{8888179} generates a forest of $N$ binary trees:
\begin{equation}
\mathcal{T} = \{t_0,t_1,\dots,t_N\}.
\end{equation}

Every tree $t \in \mathcal{T}$ is built from a bootstrap sample $X_t \subset X$. 
At each node $k$ of $t$, a random hyperplane $\mathcal{H}^t_k$ is selected by picking a random point $\mathbf{p}$ inside the distribution space limits, and a normalized random vector $\mathbf{v}$ defined as

\begin{equation}\label{eq:vector_v}
\mathbf{v} \sim\frac{\pmb{Z}}{\|\pmb{Z}\|_2}, 
\quad \text{where} 
\quad
\pmb{Z} = (Z_1,\dots,Z_{d})^{\top} 
\quad \text{and} 
\quad
Z_i \sim \mathcal{N}(0,1) \, \forall \ i \in \{1,\dots,d\}.
\end{equation}

Lesouple et al. in \cite{LESOUPLE2021109} pointed out that this way of drawing a random hyperplane may lead to the creation of empty branches, that are merely artifacts of the algorithm. To avoid this problem, they propose a particular way of selecting random hyperplanes, that we also use in our implementation of the EIF model. 

As done in \cite{LESOUPLE2021109, 8888179}, we first select a random hyperplane by drawing a unit vector $\mathbf{v}$ defined as in \eqref{eq:vector_v}. which will serve as the normal vector of the hyperplane. Then, as done in \cite{LESOUPLE2021109}, the point where the hyperplane passes is obtained by a scalar $\alpha$ drawn uniformly between the minimum value and the maximum value of the points of the dataset $X$ in the direction of the random vector previously drawn. Therefore,
\begin{equation}
\alpha \in [\min{\{\mathbf{x}\cdot\mathbf{v} \,\,\forall \mathbf{x} \in X\}},\max{\{\mathbf{x}\cdot\mathbf{v} \,\,\forall \mathbf{x} \in X\}}],\quad
\mathbf{p} = \alpha \mathbf{v} \label{p_dist}.
\end{equation}

The algorithm starts by generating a random hyperplane $\mathcal{H}^t_{k_0}$ defined by the intercept point $\mathbf{p}$ and $\mathbf{v}$ as described in Equations \ref{eq:vector_v} and \ref{p_dist}. 

Thus the hyperplane splits the dataset $X_t$ into two subsets,
\begin{align}
\begin{split}
 L^{t}_{k_0}&=\{\pmb{x}|\pmb{x} \in X^{t}_{k_0},\pmb{v}^{t}_{k_0} \cdot \pmb{x}>\pmb{v}^{t}_{k_0} \cdot \mathbf{p}^{t}_{k_0}\} ,
\\
 R^{t}_{k_0}&=X_{t} \setminus L^{t}_{k_0} .
\end{split}
\end{align}
The root node of the tree, $k_0$, is the space splitting made by the hyperplane $\mathcal{H}_{k_0}^t$.
Then, this method is applied recursively to the subsets $L^{t}_{k_0}$ and $R^{t}_{k_0}$ until the max number of splits is reached, which corresponds to the preset max depth of the tree or when the set to split has only one element.\\

Thanks to this hierarchical tree structure, to evaluate if an element is an anomaly, the model extracts the path of the point $x \in X_t$ from the root to the leaf nodes down the tree. 
Then, as IF does \cite{9347460}, the EIF algorithm uses the average depth of the point in each tree to evaluate the anomaly score, according to the paradigm that the anomalies can be isolated with few partitions. 
The average depth of the point in the trees will be translated to an anomaly score according to Equation \eqref{anom_score_if}, as in IF.


\subsection{\texorpdfstring{$\text{EIF}^+$\xspace}: a novel enhancement of EIF algorithm}\label{sec:EIFplus}
Lesouple et al. \cite{LESOUPLE2021109} showed that the EIF algorithm presented by Hariri et al. in \cite{8888179} can create misleading empty branches. On the other hand, we observed that the solution proposed in \cite{LESOUPLE2021109} 
hinders the ability to generalize well in the space around the distribution. Actually, generalization ability is very important in the context of AD, since an anomaly is a point that is outside the normal distribution of the data.

\begin{figure}[!ht] 
    \centering 
    \begin{subfigure}[t]{0.47\textwidth}
      \includegraphics[width=\linewidth]{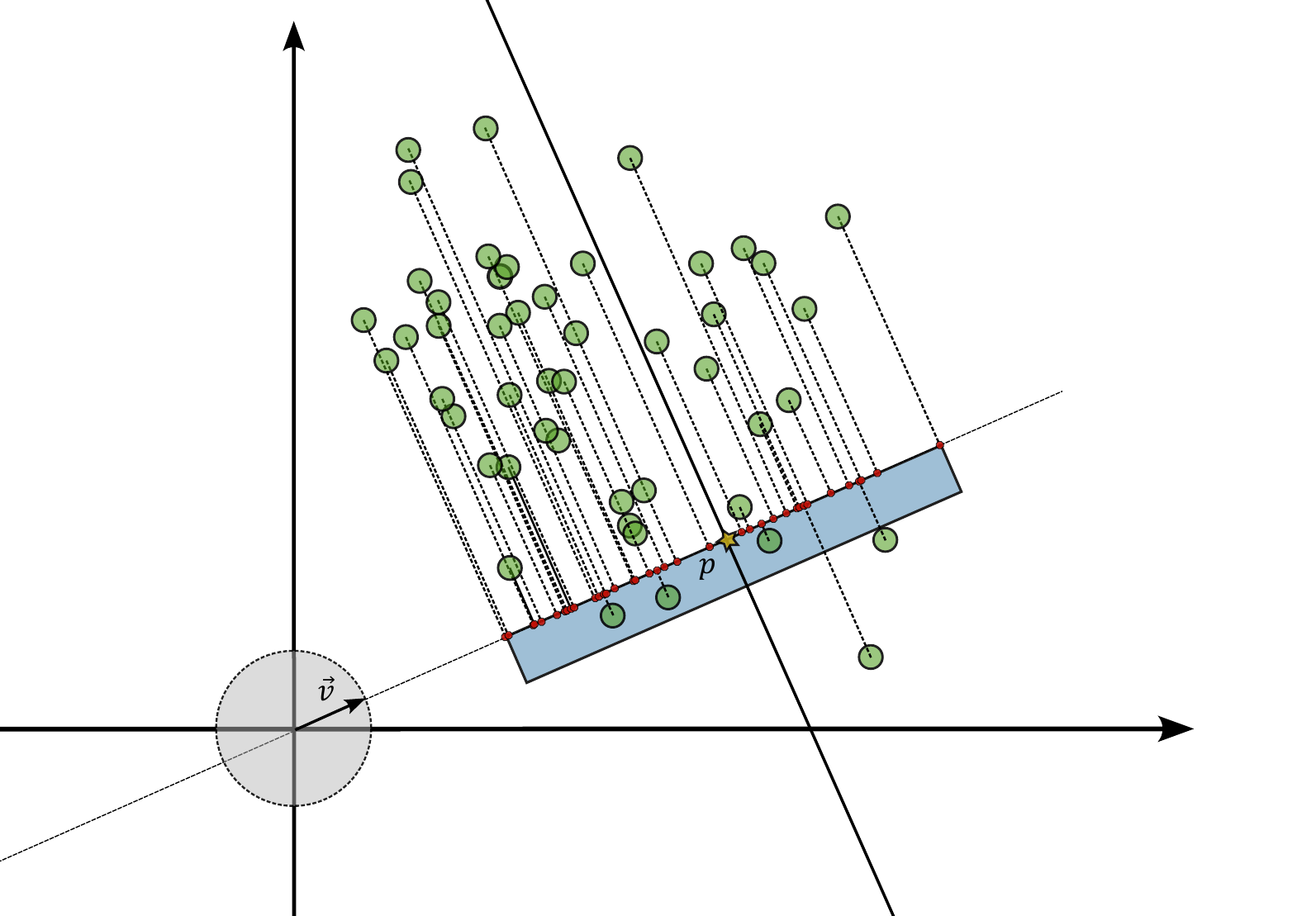}
      \caption{\rebuttal{Hyperplane selection EIF utilized by the algorithm developed by Lesouple et al. \cite{LESOUPLE2021109}. This algorithm selects the point $p$ by uniformly sampling it from an interval depicted in light blue within the figure.}}
      \label{fig:hyper-EIF}
    \end{subfigure}\hfil 
    \begin{subfigure}[t]{0.40\textwidth}
      \includegraphics[width=\linewidth]{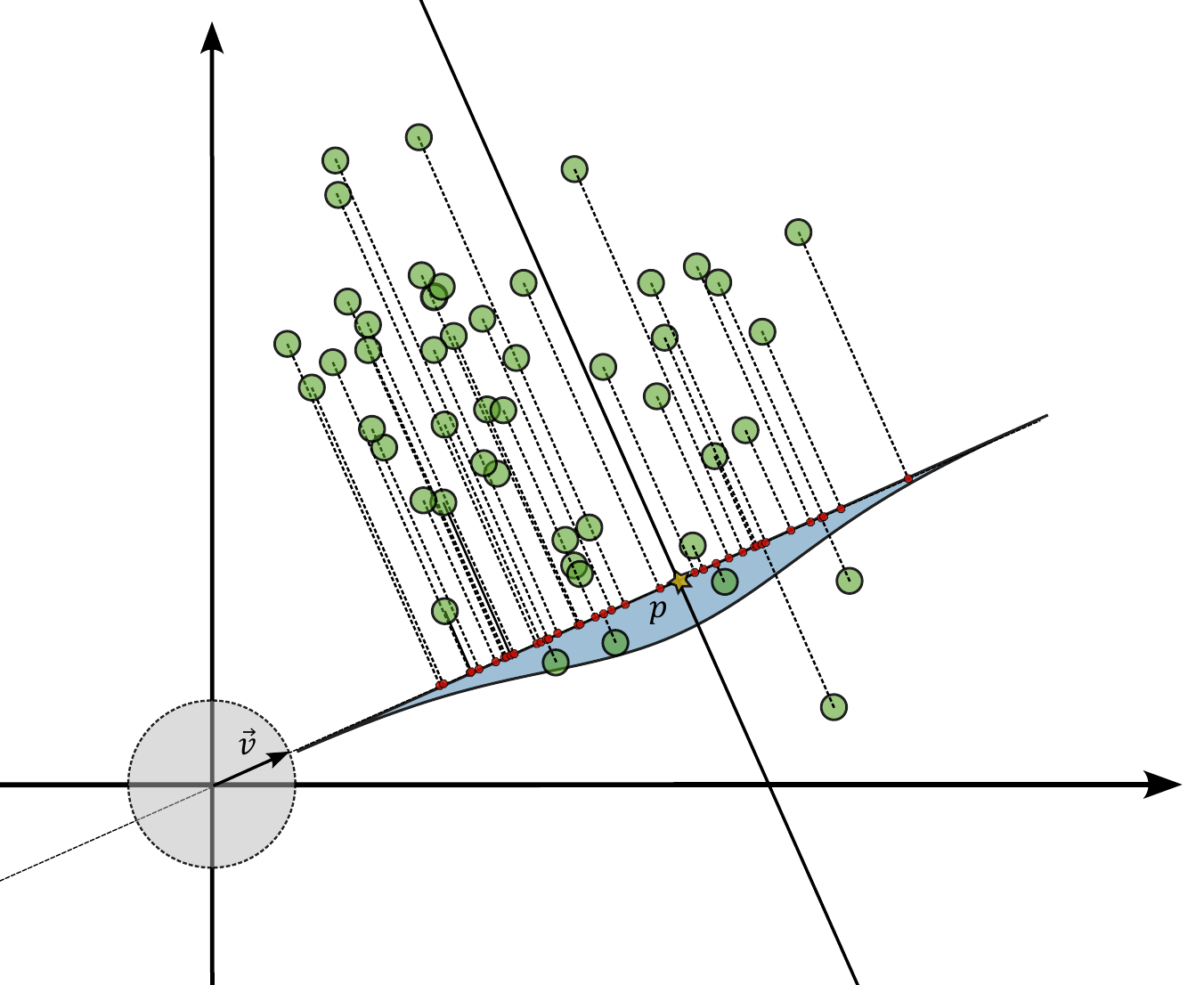}
      \caption{\rebuttal{Hyperplane selection \EIFplus, a novel approach for drawing the point $p$. This new method involves sampling the point $p$ from a normal distribution centered around the mean point of the dataset projection onto the direction described by vector $\mathbf{v}$ (represented as red dots in the figure). The distribution is visually depicted as a blue bell shape.}}
      \label{fig:hyper-EIFplus}
    \end{subfigure}\hfil 
\caption{\rebuttal{Comparison of hyperplane selection methods: (a) EIF method by Lesouple et al. and (b) the novel \EIFplus\ method.}}\label{fig:GFI-toy-2d}
\end{figure}

Therefore, we propose \EIFplus, a novel approach that enhances the EIF methodology, based on the modification introduced by Lesouple et al. \cite{LESOUPLE2021109}. \EIFplus aims to better adapt to the space surrounding the training data distribution, thus improving the performances of the EIF model when the model is fitted without anomalies in the training set. This goal is achieved by choosing splitting hyperplanes with an \textit{ad hoc} procedure, \rebuttal{in Figure \ref{fig:GFI-toy-2d} it is possible to observe the difference.}  

Let $\mathcal{A} = \{\mathbf{x}\cdot\mathbf{v} \,\,\forall \mathbf{x} \in X\}$ be the set of the point projections along the hyperplane's orthogonal direction $\mathbf{v}$. As in \cite{LESOUPLE2021109}, the point $\mathbf{p}$ is defined as  $\mathbf{p} = \alpha \mathbf{v}$.
However, instead of drawing $\alpha$ using the interval defined by the minimum and maximum values of $\mathcal{A}$  \rebuttal{as in Figure \ref{fig:hyper-EIF}}, \EIFplus draws it from a normal distribution $\mathcal{N}(\mathop{\mathbb{E}}[\mathcal{A}],\eta \sigma(\mathcal{A}))$ as shown in Figure \ref{fig:hyper-EIFplus}.
Even if \EIFplus can generate empty branches, 
they contribute to the formation of a model that exhibits enhanced generalizability, as will be shown in Section \ref{sec:Evaluation}. 

\section{Interpretability of Isolation-based Anomaly Detection models}

\rebuttal{In the context of anomaly detection models, interpretability is crucial because it allows us to understand and trust the model's decisions, especially in critical applications such as fraud detection, network security, and medical diagnostics. An interpretable model enables practitioners to verify whether the model is identifying meaningful patterns or if it is influenced by spurious correlations, ultimately ensuring that the anomalies detected are actionable and reliable.

With this in mind, we move on to the task of interpreting the predictions made by the isolation-based models introduced in Section \ref{sec:isolation_based_approaches}.} Interpretation algorithms aim to explain the latent patterns identified by the models, thereby enriching our comprehension of their outputs.
Drawing inspiration from the DIFFI algorithm developed by Carletti et al. \cite{diffi} for the IF, we introduce \approach, a novel model-specific algorithm designed to interpret the results generated by the IF, EIF and its variant, the \EIFplus.


\subsection{Depth-based Isolation Forest Feature Importance}\label{DIFFI}
\rebuttal{Depth-based Isolation Forest Feature Importance (DIFFI) was the first unsupervised model-specific method addressing the need to interpret the IF model \cite{diffi}. It exploits the structure of the trees in the IF algorithm to understand which features are the most relevant to discriminate whether a point is an outlier or not.
In particular, a meaningful feature should isolate the anomalies as soon as possible, and create a high imbalance when isolating anomalous points (the opposite being true for inliers).

We briefly explain how DIFFI works and we introduce the related notation, as we will leverage it when introducing the novel \approach approach to explain EIF and \EIFplus predictions.
From \cite{diffi}, we define:
\begin{description}
    \item[Induced Imbalance Coefficients $\lambda$]: given an internal node $k$  of an isolation tree, as defined in Section \ref{IF}, let $n(k)$ be the number of points that the node divides, being $n_l(k)$ and  $n_r(k)$  the number of points on the left and the right child, respectively. The coefficient measuring the induced imbalance of the node $v$ is:

    \begin{equation}
        \lambda(v)=    \begin{dcases}
      0  \,\,\,& \mbox{if }  n_l(k)=0   \mbox{ or }  n_r(k)=0 \\
      \tilde{\lambda}(k)  \,\,\,& \text{otherwise}
    \end{dcases}
    \end{equation}
    where
    \begin{equation}
        \tilde{\lambda}(k) = g\left(\frac{\max(n_l(k),n_r(k))}{n(k)}\right) \quad \text{and} \quad
        g(a)=\frac{a-\lambda_{min}(n)}{2(\lambda_{max}(n)-\lambda_{min}(n))} +0.5.
    \end{equation}
    In the previous equation, $\lambda_{min}$ and $\lambda_{max}$ denote the minimum and maximum scores that can be obtained a priori given the number of data $n(k)$. 
    
    \item[Cumulative feature importances $\mathbf{I}$]: it is a vector of dimension $d$ (i.e., the number of features) where the $j$-th component is the feature importance of the $j$-th feature. In 
    \cite{diffi}, authors distinguish between $\mathbf{I}_I$,  created based on predicted inliers, and $\mathbf{I}_O$, based on the outliers.
    Concerning $\mathbf{I}_I$, the procedure starts with the initialization $\mathbf{I}_I= \mathbf{0}_{d}$. Considering the subset of predicted inliers for the tree $t$, $\mathcal{P}_{I,t}$, for each predicted inlier $x_I \in \mathcal{P}_{I,t}$, DIFFI iterates over the internal nodes in its path $Path(x_I , t)$.
    If the splitting feature associated with the generic internal node $v$ is $f_j$, then we update the $j$-th component of $\mathbf{I}_I$ by adding the quantity:
    \begin{equation}\label{depth}
        \Delta = \frac{1}{h_t(x_I)}\lambda_I(v)
    \end{equation}
    The same procedure applies for $x_O \in \mathcal{P}_{O,t}$ and $\mathbf{I}_O$. Intuitively, at each point and each step, the vector accumulates the imbalance produced by each feature. The imbalance is measured by the previously defined $\lambda$ and weighted by the depth of the node. This means that, according to DIFFI, features that allowed to isolate the points sooner, are considered to be more useful.
    
    \item[Features counter $\mathbf{V}$]: it is used to compensate for uneven random feature sampling that might bias the calculated cumulative feature importance.  At each passage of a point through a node, the entry of the counter corresponding to the splitting feature is incremented by one.  As for the cumulative feature importance, two feature counters are calculated, the one for predicted inliers $\mathbf{V}_I$ and the one for outliers $\mathbf{V}_O$.

\end{description}

Based on the above-introduced quantities, DIFFI computes the Global Feature Importance by looking at the weighted ratio between outliers and inliers cumulative feature importance:
    \[ GFI = \frac{\mathbf{I}_O/\mathbf{V}_O}{\mathbf{I}_I/\mathbf{V}_I} \]
}

\subsection{Extended Isolation Forest Feature Importance (\approach)} \label{ExIFFI}
Drawing inspiration from DIFFI, we introduce \approach, the Extended Isolation Forest Feature Importance, a generalization of the DIFFI algorithm that is able to rank the importance of the features in deciding whether a sample is an anomaly or not for any of the IF models where the space is splitted by an hyperplane.

As seen in Section \ref{ExtendedIF}, a node $k \in \mathcal{K}_t$ in an EIF tree $t \in \mathcal{T}$ corresponds to an hyperplane $\mathcal{H}^{t}_{k}$, that splits the subset $X_k^t \subseteq X^t \subseteq X$. Using the Lesouple et al. \cite{LESOUPLE2021109} correction of the EIF, $\mathcal{H}^{t}_{k}$ is completely defined by means of a vector orthogonal to its direction $\mathbf{v}^{t}_{k}$, defined as in \eqref{eq:vector_v}, and a point $\mathbf{p}^{t}_{k}$ that belongs to it, defined as in \eqref{p_dist}.

The hyperplane $\mathcal{H}^{t}_{k}$ separates the points in a set of elements on the left side of the hyperplane and a set of elements on the right side of the hyperplane. 
\begin{align}
\begin{split}
 L^{t}_{k}&=\{\mathbf{x}|\mathbf{x} \in X^{t}_{k},\mathbf{v}^{t}_{k} \cdot \mathbf{x}>\mathbf{v}^{t}_{k} \cdot \mathbf{p}^{t}_{k}\} ,
\\
 R^{t}_{k}&=X^{t}_{k} \setminus L^{t}_{k} .
\end{split}
\end{align}

\approach computes a vector of feature importances for each node of the tree, based on two intuitions:
\begin{itemize}
    \item The importance of the node $k$ for a point $x$ is higher when $k$ creates a greater inequality between the number of elements on each side of the hyperplane, and $x$ is in the smaller subset. Indeed, greater inequality means a higher grade of isolation for the points in the smaller set.
    \item For node $k$, the relative importance of the $j$-th feature is determined by the projection of the normal vector of the splitting hyperplane onto that feature. If the split occurs along a single feature, that feature will receive the entire importance score.
    If the splitting hyperplane is oblique, the importance scores of multiple features will be calculated based on their respective projections onto the normal vector of the hyperplane.
\end{itemize}

Thus, we assign an importance value function to every node of the trees that is the projection on the normal vector of the splitting hyperplane of the quotient between the cardinality of the sample before a particular node and after it, following the path of a sample $\mathbf{x}$.
Thus, knowing that the splitting hyperplane $\mathcal{H}^t_k$ of that node is determined by the pair $\{\mathbf{v}^{t}_{k},\mathbf{p}^{t}_{k}\}$ \footnote{With $abs(\mathbf{v})$ we refer to the positive part of every element of the vector $\mathbf{v}$}:
\begin{equation}\label{importance}
\mathbf{\lambda}^{t}_{k}(\mathbf{x}) = \begin{cases} \left(\frac{|X^{t}_{k}|}{|L^{t}_{k}|}\right) abs(\mathbf{v}^{t}_{k}), & \mbox{if } \mathbf{v}^{t}_{k} \cdot \mathbf{x}>\mathbf{v}^{t}_{k} \cdot \mathbf{p}^{t}_{k}\\
\left(\frac{|X^{t}_{k}|}{|R^{t}_{k}|}\right) abs(\mathbf{v}^{t}_{k}),& \mbox{otherwise} \end{cases}
\end{equation}

\rebuttal{
For example, consider a node \( k \) with a hyperplane defined by \( \mathbf{v}^{t}_{k} = [0.5, 0.7, 0.2] \) and \( \mathbf{p}^{t}_{k} = [2.0, 2.5, 1.0] \). Suppose a point \( \mathbf{x} = [2.5, 3.0, 1.5] \). We first calculate the dot products:

\[
\mathbf{v}^{t}_{k} \cdot \mathbf{x} = 0.5 \times 2.5 + 0.7 \times 3.0 + 0.2 \times 1.5 = 1.25 + 2.1 + 0.3 = 3.65
\]

\[
\mathbf{v}^{t}_{k} \cdot \mathbf{p}^{t}_{k} = 0.5 \times 2.0 + 0.7 \times 2.5 + 0.2 \times 1.0 = 1.0 + 1.75 + 0.2 = 2.95
\]

Since \( \mathbf{v}^{t}_{k} \cdot \mathbf{x} = 3.65 \) is greater than \( \mathbf{v}^{t}_{k} \cdot \mathbf{p}^{t}_{k} = 2.95 \), the point \( \mathbf{x} \) lies in \( R^{t}_{k} \). If the subsets contain \( |L^{t}_{k}| = 90 \) and \( |R^{t}_{k}| = 10 \) points out of \( |X^{t}_{k}| = 100 \), the importance vector \( \mathbf{\lambda}^{t}_{k}(\mathbf{x}) \) would be:

\[
\mathbf{\lambda}^{t}_{k}(\mathbf{x}) = \left(\frac{100}{10}\right) \times \text{abs}([0.5, 0.7, 0.2]) = [10 \times 0.5, 10 \times 0.7, 10 \times 0.2] = [5, 7, 2]
\]}

The vector of importances evaluated in the tree $t$ for a point $x$ is the sum of all the importance vectors of all the nodes that the element $x$ passed through on its path to the leaf node in the tree $t$ defined in Equation (\ref{setnodesx}):
\begin{equation}\label{sumimp}
\mathbf{I}_{t}(x)=\sum_{k \in \mathcal{P}^{t}_x} \mathbf{\lambda}_{t,k}(x).
\end{equation}

We then calculate the sum of the importance vector of the point $x$ for all the trees in $\mathcal{T}$:
\begin{equation}\label{equaimport_2}
\mathbf{I}(x) = \sum_{t \in T} \mathbf{I}_{t}(x)
\end{equation}

We define $\mathbf{V}(x)$ as the sum of the vectors orthogonal to the hyperplanes of the nodes that an element $x$ passes in a tree, then we calculate the sum of the values in all the trees:
\begin{equation}
\mathbf{V}(x) = \sum_{t \in T}\sum_{k \in \mathcal{P}^{t}_x} \mathbf{v}_k^t
\end{equation}

\subsubsection{\approach: Global Feature Importance}\label{sec: GFI}
To globally evaluate the importance of the features, we divide $X$ into the subset of predicted inliers $\mathcal{Q}_I=\{\mathbf{x}_i \in X|\hat{y}_i=0\}$  and the one of predicted outliers $\mathcal{Q}_O=\{\mathbf{x}_i \in X|\hat{y}_i=1\}$ where $\hat{y}_i \in \{0,1\}$ is the binary label produced by the thresholding operation indicating whether the corresponding data point $\mathbf{x}_i$ is anomalous $(\hat{y}_i =1)$ or not $(\hat{y}_i =0)$. 
    
We define the global importance vectors for the inliers and for the outliers by summing out the importance values introduced in Equation (\ref{equaimport_2}):
    \begin{align}
        \mathbf{I}_I=\sum_{\mathbf{x}\in \mathcal{Q}_I} \mathbf{I}(\mathbf{x}),\quad
        \mathbf{I}_O=\sum_{\mathbf{x}\in \mathcal{Q}_O} \mathbf{I}(\mathbf{x}).
    \end{align}
    Likewise the sum of the orthogonal vectors:
    \begin{align}
        \mathbf{V}_I=\sum_{\mathbf{x}\in \mathcal{Q}_I} \mathbf{V}(\mathbf{x}),\quad
        \mathbf{V}_O=\sum_{\mathbf{x}\in \mathcal{Q}_O} \mathbf{V}(\mathbf{x}).
    \end{align}
    
    Due to stochastic sampling of hyperplanes, in order to avoid the bias generated by the fact that it is possible that some dimensions are sampled more often than others, the vectors of importance have to be divided by the sum of the orthogonal vectors.
    \begin{align}\label{impvectfinal}
        \hat{\mathbf{I}}_I=\frac{\mathbf{I}_I}{\mathbf{V}_I},\quad
        \hat{\mathbf{I}}_O=\frac{\mathbf{I}_O}{\mathbf{V}_O}.
    \end{align}

    
    To evaluate which are the most important features to discriminate a data as an outlier we divide the importance vector of the outliers by the one of the inliers. Equation (\ref{impvectfinal}), and we obtain the global feature importance vector in the same vein as in the DIFFI algorithm \cite{diffi}:
    \begin{equation} \label{gfi_eq}
        \mathbf{GFI}=\frac{\hat{\mathbf{I}}_O}{\hat{\mathbf{I}}_I}.
    \end{equation}

\rebuttal{For clarity, the pseudocode of the \approach algorithm for the computation of the Global Feature Importance is reported in \ref{alg:exiffi}}. 


\begin{algorithm}[ht!]
\caption{ExIFFI Algorithm}\label{alg:exiffi}

\SetKwInOut{Input}{Input}
\SetKwInOut{Output}{Output}

\Input{Isolation based AD model $F$, Input dataset $\mathcal{D}$}
\Output{Global Feature Importance vector $GFI \in \mathbb{R}^p$}

$I(\mathbf{x}) \gets (0)_p \quad \forall \mathbf{x} \in \mathcal{D}$\;
\tcp{Initialize all importance vectors to zero}

\ForEach{$\mathbf{x}$ in $\mathcal{D}$}{
    \ForEach{Isolation Tree $t$ in $F$}{
        $\mathcal{P}_x^t= \{k \in t \mid \mathbf{x} \text{ traverses node } k \text{ in tree } t\}$\;
    
        $I_t(x) \gets \sum_{k \in \mathcal{P}_x^t} \lambda_{k,t}(x)$
        \tcp{Aggregate the importance scores of all nodes in the path from the root to the leaf node containing $\mathbf{x}$}
        $V_t(x) \gets \sum_{k \in \mathcal{P}_x^t} v_{k,t}(x)$\;
        \tcp{Aggregate the vectors orthogonal to hyperplanes of the nodes in the path from the root to the leaf node containing $\mathbf{x}$}
        $I(\mathbf{x}) \gets \sum_{t \in F} I_t(\mathbf{x})$\;
        \tcp{Aggregate the importance scores for $x$ over all the trees in the forest $F$}
        $V(\mathbf{x}) \gets \sum_{t \in F} \sum_{k \in \mathcal{P}_x^t} \mathbf{v}_k^t$\;
        \tcp{Aggregate the vectors of the direction of the cut for $x$ over all the trees in the forest $F$}
        }
    }
    $\mathcal{P}_{I}, \mathcal{P}_{O} \gets \text{Predict}(\mathcal{D},F)$\;
    \tcp{Get the predicted inliers and outliers }
    $\mathbf{I}_I \gets \sum_{\mathbf{x} \in \mathcal{P}_I} \mathbf{I}(\mathbf{x})$\;
    $\mathbf{I}_O \gets \sum_{\mathbf{x} \in \mathcal{P}_O} \mathbf{I}(\mathbf{x})$\;
    \tcp{Sum together the importance vectors of all inliers and outliers}

    $\mathbf{V}_I \gets \sum_{\mathbf{x} \in \mathcal{P}_I} \mathbf{V}(\mathbf{x})$\;
    $\mathbf{V}_O \gets \sum_{\mathbf{x} \in \mathcal{P}_O} \mathbf{V}(\mathbf{x})$\;
    \tcp{An analogous operation is done for the orthogonal vectors}

    $\hat{\mathbf{I}}_I \gets \frac{\mathbf{I}_I}{\mathbf{V}_I}$\;
    $\hat{\mathbf{I}}_O \gets \frac{\mathbf{I}_O}{\mathbf{V}_O}$\;
    \tcp{Compute the final normalized importance vectors separately for inliers and outliers}
    
    $GFI \gets \frac{\hat{\mathbf{I}}_O}{\hat{\mathbf{I}}_I}$\;

\Return{$GFI$}\;

\end{algorithm}

\subsubsection{\approach: Local Feature Importance}
    The Local Feature Importance assumes significance primarily within the context of anomalous data points, especially from the point of view of applications. Indeed, providing explanations of samples deemed anomalous eases decision-making by domain experts, who can subsequently tailor their responses based on the salient features driving the anomaly of a single point.
    Let's take into account an element $x$, the Equation (\ref{equaimport_2}) gives a vector of importances $\mathbf{I}(x)$ of the sample $x$ for each feature. Then the vector $\mathbf{V}(x)$ is the normalization factor of the feature importance.
    Thus, the Local Feature Importance ($\mathbf{LFI}$) of an element $x$ is the quotient: 
    \begin{equation} \label{lfi_eq}
        \mathbf{LFI}(x) = \frac{\mathbf{I}(x)}{\mathbf{V}(x)}.
    \end{equation}

    The pseudocode for the $\mathbf{LFI}$ computation is analogous to the one reported in \ref{alg:exiffi} with the difference that equation \ref{lfi_eq} is used in place of equation \ref{gfi_eq} in the last step.

\subsubsection{Visualizing Explanations}\label{sec:viz-explanations}
Miller \cite{MILLER20191} defines interpretability in AI models as the extent to which a human can comprehend the rationale behind a decision. 
To be effective, interpretability should deliver a clear and comprehensible representation of how inputs influence outputs, even for individuals who are not experts in the field.

To bolster users' trust in the model, relying solely on a series of numerical Scores is insufficient. Providing a series of summary scores and comprehensible graphical representations of these scores may help the evaluation of the model outputs and bolster its interpretational efficacy.

To achieve these goals, two distinct graphical representations are proposed 

\begin{figure}[htb] 
    \centering 
    \begin{subfigure}{0.50\textwidth}
      \includegraphics[width=\linewidth]{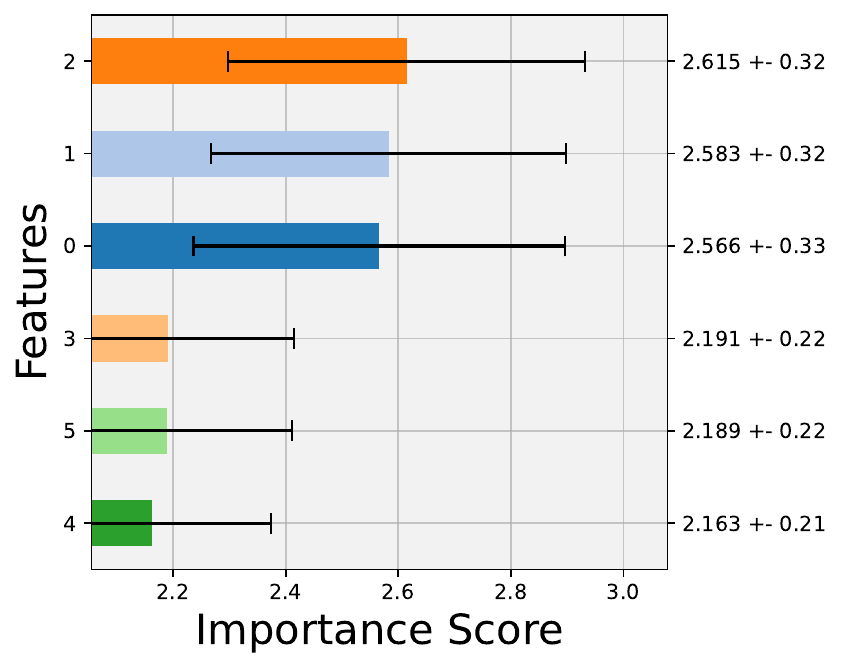}
      \caption{Score Plot}
      \label{fig:Feat-imp-Bisect-3d}
    \end{subfigure}\hfil 
    \begin{subfigure}{0.50\textwidth}
        \centering
        \includegraphics[width=\linewidth]{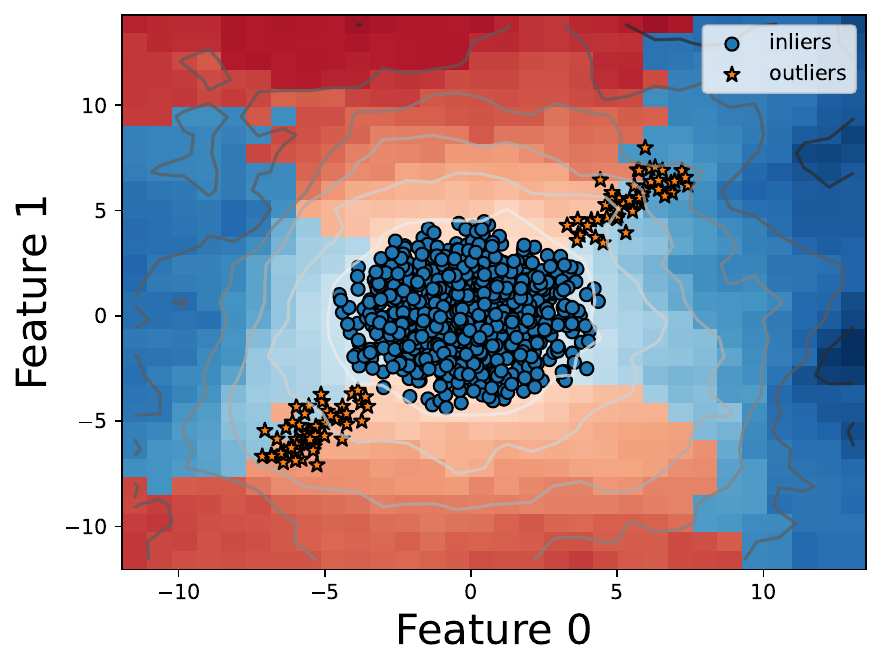}
        \caption{Scoremap}
        \label{fig:Imp-scoremap-Bisect-3d}
    \end{subfigure}\hfil 
\caption{Score Plot \ref{fig:Feat-imp-Bisect-3d}, and Scoremap \ref{fig:Imp-scoremap-Bisect-3d} of the dataset {\ttfamily Bisect3d}}
\label{fig:Bisect-3d}
\end{figure}

\begin{description}
    \item[GFI Score Plot]  The Score Plot is the average importance score across runs of the algorithm. For clarity, we introduce the Score Plot, a horizontal bar plot displaying average Global Importance Scores across training executions (Figure \ref{fig:Feat-imp-Bisect-3d}).
    
    \item[LFI Scoremap] The Scoremap provides local interpretability by computing Local Importance Scores for feature pairs. Each grid point represents the feature with the highest score, shown in red or blue for the first and second features respectively. Darker shapes indicate higher Feature Importance Scores. Scatter plots overlay data points, distinguishing inliers (blue dots) and outliers (red stars). Contour plots offer contextual insights of the anomaly score regions. Focusing on the most important feature pair is common practice, facilitating anomaly dispersion analysis. For instance, in the {\ttfamily Bisec3D} dataset (Figure \ref{fig:Imp-scoremap-Bisect-3d}), this strategy was employed. 
\end{description}

\rebuttal{
The GFI Score Plot represents the outcome of multiple model runs, 40 in this paper (Section \ref{sec:Evaluation}), to account for the stochastic nature of IF-based models like EIF and \EIFplus. This approach recognizes the inherent randomness of these models and leverages multiple runs to capture variability, enhancing the reliability of interpretability results and providing deeper insight into the influence of individual features on model output.}

To illustrate the interpretation of the graphical representation, we use the {\ttfamily Bisect3D} dataset introduced in \ref{sec:appendix_datasets}. In this dataset, anomalies are placed along the bisector of Features 0, 1, and 2, while inliers are centered around the origin, following a normal distribution across six dimensions. Consequently, Features 0, 1, and 2 play a key role in detecting outliers, and due to model stochasticity, any of these three features can be top-ranked in different executions.

The Score Plot in Figure \ref{fig:Feat-imp-Bisect-3d} shows nearly identical Importance Scores for these three features, with the rest having consistently lower scores. The feature ranking may vary with additional model runs.

The scatter plot in Figure \ref{fig:Imp-scoremap-Bisect-3d} highlights that outliers, represented by stars, deviate from the ball of inliers. The color distribution in the Scoremap indicates how the Local Importance Score varies across feature space. Lighter shades dominate the center, while the further from the distribution of inliers, the darker the colour are, we observe that the anomalies lay in between the change of the colour of the most important feature, underlying how the outliers are distributed in the intersection of the two features.

\rebuttal{To sum up, ExIFFI serves as the feature importance mechanism for the EIF, extending the ideas from DIFFI, which focuses on the IF. While DIFFI evaluates feature importance by assessing the imbalance created by single-feature splits in IF trees, ExIFFI adapts this approach to the more complex hyperplane splits in EIF. By leveraging the tree structure, ExIFFI determines how each feature contributes to creating imbalanced splits, ranking features based on their importance in detecting anomalies within the EIF framework. This structural method allows for understanding feature importance both globally, across the entire dataset, and locally, for individual data points. } 

\section{Experimental Setup}\label{sec:experimental_results}
\rebuttal{
To evaluate the performance of the \EIFplus model and the interpretability of \approach, we conducted experiments using various benchmark datasets, detailed in Section \ref{sec:appendix_datasets}. The implementation for \EIFplus, \approach, and the experimental setup are publicly available at \url{https://github.com/alessioarcudi/ExIFFI}. 
In Section \ref{sec:performance_evaluation}, we present in detail how we will conduct the experiments about the model's ability to detect anomalies, especially under low contamination conditions, and provide a comparative analysis against other AD models. Additionally, in Section \ref{sec:interpr_evaluat}, we present instead the proxy tasks used in order to evaluate the accuracy of \approach's interpretability and comparing it with the ad-hoc method DIFFI and ECOD and various post-hoc interpretation techniques.
}

\ale{
Following the evaluation of the datasets, we analyze \rebuttal{the soundness} and computational efficiency of these algorithms. \rebuttal{In Section \ref{sec:corr-tab} we compute the correlation between the local importance scores provided by different interpretation algorithms with the anomaly scores provided by the explained AD model with the aim of identifying a strong linear relationship between these two quantities, which serves as a proof of the correctness of the explanations.} \rebuttal{Finally,} in Section \ref{sec:time_scaling}  we compare the algorithms considering the time needed to fit the model, predict the results and computing feature importances. This examination highlights the efficiency of the proposed models.
}

\subsection{Performance Evaluation}\label{sec:performance_evaluation}
    \ale{We assess the performance of the \EIFplus model by benchmarking it against several anomaly detection (AD) models. This evaluation includes both well-established models from the literature and a state-of-the-art unsupervised AD model. The models compared are:
    \begin{itemize}
        \item Traditional Isolation-based approaches such as Isolation Forest (IF) and Extended Isolation Forest (EIF).
        \item The novel Deep Isolation Forest (DIF), an advanced Isolation-based model integrating deep learning techniques for enhanced AD \cite{xu2023deep}.
        \item The Anomaly AutoEncoder \cite{AutoEncoder}, a deep learning-based AD model, recognized for its robust performance in unsupervised AD scenarios.
        \item \rebuttal{The Empirical Cumulative distribution based Outlier Detection (ECOD) \cite{ECOD}, a non-parametric AD method which uses the Empirical Cumulative Distribution function (ECDF) of each single feature to compute the anomaly score, following the intuition that anomalous points should be placed on the tails of distributions.}
    \end{itemize}
}
     
\rebuttal{
In our experiments, we will evaluate the performance of anomaly detection (AD) models across different levels of contamination within the training dataset. This analysis is vital, as highlighted by Kim et al. \cite{kim2023iterative}, who stress the importance of examining the robustness of AD models when the "normality assumption" (that all training data are inliers) does not hold.

It is crucial to assess how AD models respond to varying levels of contamination in the training data, where normal and anomalous samples are mixed without labels. This evaluation not only reveals the models' effectiveness in real-world scenarios but also highlights the impact of anomalies concentrated in specific regions of the data space. When anomalies are misclassified as normal, the model may adapt to these outliers, distorting its representation of the overall data distribution. As a result, the model’s ability to correctly identify and isolate true anomalies can be significantly diminished.}



Specifically, our analysis studies anomaly contamination's effect on average precision, focusing on EIF versus \EIFplusspace performance differences. This thorough assessment benchmarks \EIFplus against traditional methods and showcases its generalization ability.

With reference to the analysis of the performances varying the contamination, we consider two distinct experimental scenarios chosen to mirror the varied conditions under which AD models might be deployed, offering a comprehensive understanding of their adaptability and effectiveness:
\begin{itemize}
    \item Scenario I: We fit and evaluate the models using the entire dataset.
    \item Scenario II: We fit the models exclusively on the inliers within the dataset, that correspond to the scenario of $0$ contamination of outliers. Subsequently, we assess the Average Precision scores of these models when applied to the entirety of the dataset.
\end{itemize}

\rebuttal{Finally in Section \ref{sec:perf-tab} the performances of the AD models are analyzed in terms of average precision, precision and AUC-ROC score. Results are detailed in the Table\ref{tab:ComprehensiveConsolidated}.}


\ale{
\subsection{Interpretability Evaluation}\label{sec:interpr_evaluat}
    Following the performance assessment, we shift our focus to the interpretability of AD models, evaluating the \approach algorithm, an ad-hoc interpretation algorithm for the isolation based anomaly detections such as IF, EIF and \EIFplus.
}

\rebuttal{Our evaluation is carried out in two stages. Initially, we qualitatively examine the \approach algorithm's interpretive method through the plots detailed in section \ref{sec:viz-explanations}, observing how the resulting visualizations provide insights on the nature of the anomalies. 
}

\rebuttal{    
    We then proceed to assess our model interpretive effectiveness by comparing it with alternative interpretation algorithms. 
    
    \begin{itemize}
    \item Ad-hoc algorithm named DIFFI, introduced in Section \ref{DIFFI}.
    \item Leveraging the ad-hoc interpretation of ECOD model \cite{ECOD} we developed a local and global interpretation on its feature outlier scores to create a common environment of comparison. The model assigns an outlier score \( s_i \) to each feature \( i \) given a sample $x$, and the paper analyzes feature outliers by comparing each feature's score \( s_i(x) \) to the 99th percentile \( P_{99} \) saying that the closer \( s_i \) is to \( P_{99} \), the more outlier-like the feature. Than we defined the LFI as:

    \[I_{\text{local}}(x) = \frac{1}{1 + (s_i(x) - P_{99})^2}\]
    
    The GFI, similar to ExIFFI, is the mean importance of the outliers $\mathcal{P}_O$ features divided by the mean importance of the inliers $\mathcal{P}_O$:      $\mathbf{I}_I = \frac{\sum_{\mathbf{x} \in \mathcal{P}_O} \mathbf{I}(\mathbf{x})}{|\mathcal{P}_O|}\frac{|\mathcal{P}_I|}{\sum_{\mathbf{x} \in \mathcal{P}_I} \mathbf{I}(\mathbf{x})}$

    \item Post-hoc method that leverages Random Forest as a surrogate model. This approach involves using an inherently interpretable model to estimate the predictions of the initial Isolation Forest models (IF, EIF, or \EIFplus), with interpretations derived from the feature importance scores that the surrogate model provides. 
    \end{itemize}
}

\ale{
    In the absence of a one-size-fits-all metric or fully comprehensible prior knowledge on the features, assessing the performances of such algorithms is challenging due to the dependency on various factors, including input data complexity, model intricacies, and end-user interpretability, then the evaluation becomes a delicate balancing act between quantifiable measures and qualitative insights. 
}

\ale{
    To quantitatively evaluate model interpretability, a proxy task is often utilized as an indirect means to assess model performance. In this paper the Feature Selection acts as such a proxy, where Importance Scores from interpretability methods prioritize input features. Then we assess the interpretation algorithm performance based on how it prioritizes key features versus less important ones offering insight into its effectiveness using the Feature Selection. Additionally, we add to these results those from a casual feature selection to gauge the overall performance trends more broadly.}

We then introduce a metric to quantify the goodness of a Feature Selection, the Area Under the Curve of Feature Selection ($AUC_{FS}$), taking inspiration from the methodological evaluation used by Turbe et al. in \cite{turbe2023evaluation}. $AUC_{FS}$ is calculated as the difference between the area under the curve when shrinking the dimensions of the dataset on the most important features ($AUC_M$) and the area under the curve when considering the least important features ($AUC_L$), \(AUC_{FS} = AUC_M - AUC_L\). This metric reflects the distinction in selecting significant features or dropping them, with a larger gap indicating more effective prioritization, as on one hand losing vital information leads to a noticeable performance decline when important features are excluded early on, on the other hand maintaining the most important information will result in maintaining the performances or even increasing them.
\rebuttal{The pseudocode \ref{alg:feature_selection_proxy} presents in details how the $AUC_{FS}$ score is calculated.}


\begin{algorithm}[!ht]
\caption{Feature Selection Proxy Task} \label{alg:feature_selection_proxy}
\SetKwInOut{Input}{Input}
\SetKwInOut{Output}{Output}

\Input{Isolation based AD model $F$, $GFI$ ranking $\mathcal{R}$, dataset $X$}
\Output{Average Precision vector $AP$}

$AP_{least} \gets [0,\dots,0]$\;
$AP_{most} \gets [0,\dots,0]$\;
$X_{least} \gets X$\;
$X_{most} \gets X$\;

\For{$i \gets 1$ \KwTo $p$}{
        $AP_{least}[i] \gets \text{average\_precision}(F, X_{least})$\;
        $AP_{most}[i] \gets \text{average\_precision}(F, X_{most})$\;

        $X_{least} \gets X_{least} \setminus X_{least}[:, \mathcal{R}[-i]]$\;
        \tcp{Inverse Approach: Remove the least important feature}

        $X_{most} \gets X_{most} \setminus X_{most}[:, \mathcal{R}[i]]$\;
        \tcp{Direct Approach: Remove the most important feature}
}

\Return{$(AP_{least}, AP_{most})$}\;

\end{algorithm}

There exist various post-hoc interpretation algorithms like SHAP \cite{NIPS2017_8a20a862} that could be integrated as a comparison in the proxy task. However, due to their computational burden, these algorithms are often less suitable for tree-based models, which are widely employed in industrial settings due to their speed and low memory requirements. \rebuttal{Our experiments confirmed the unsuitability of SHAP in our evaluation \ref{sec:time_scaling}}, even considering the faster variants of SHAP \cite{lundberg2018consistent}, the required computational time necessary to apply it to the benchmark datasets is excessive, making it not viable in practice.

Our interpretability evaluation bypasses human experiments due to their high cost in terms of required time and effort, focusing instead on metric-based assessments. Like other studies \cite{turbe2023evaluation,wojtas2020feature}, we adopt, according to the Doshi-Velez's et al. taxonomy \cite{doshi2017towards}, a Functionally-grounded evaluation by using a proxy task (Feature Selection). This method is neither specific nor expensive, making it ideal for our purposes. It provides a practical framework for assessing interpretability without the significant resources required for human-based evaluations, aligning with our research constraints and objectives.


\section{Experimental Evaluation of \EIFplus and \approach}\label{sec:Evaluation}

This section evaluates \EIFplus and \approach, demonstrating their improvements over existing models. We focus on performance, particularly interpretability and anomaly detection. The analysis also considers their practicality in various settings by examining computational time.

\rebuttal{We use a benchmark of 16 datasets with labeled anomalies to evaluate the performances of the models. This benchmark includes 5 synthetic datasets, which were designed to highlight the differences in model performance and to provide a ground truth about anomalies and model interpretation, as well as 11 open source datasets based on real applications. In the rest of the paper, the datasets will be indicated using a {\ttfamily typewriter} font. Table \ref{tab:dataset_used} summarizes the key characteristics of the datasets that were examined. A detailed description can be found in \ref{sec:appendix_datasets}.

\begin{table*}[!ht]
\centering
\caption{Experimental datasets overview: The first column lists the dataset names; the second specifies the total number of instances; the third details the count of outliers; the fourth shows contamination rates; the fifth indicates the total features; the final column reveals the dataset's dimensionality.}
\label{tab:dataset_used}
\footnotesize
\begin{tabular}{|r|rrrrrr|}
\toprule
\hline
{} &  data  &  anomalies & contam  &  features & Size &  Dataset Type \\
{} &  $n$  &    & \%  &  $d$ & &\\
\hline
\texttt{Xaxis} & 1100 & 100 & 9.09 & 6 & (Low) & Synthetic \\
\texttt{Bisect} & 1100 & 100 &  9.09 & 6 & (Low) & Synthetic \\
\texttt{Bisec3D} & 1100 & 100 &  9.09 & 6 & (Low) & Synthetic \\
\texttt{Bisec3D\_Skewed} & 1100 & 100 &  9.09 & 6 & (Low) & Synthetic \\
\texttt{Bisec6D} & 1100 & 100 & 9.09 & 6 & (Low) & Synthetic \\
\texttt{Annthyroid}   & 7200 & 534 & 7.56               & 6                                   & (Low) & Real \\
\texttt{Breastw}      & 683 & 239  & 52.56             & 9                                     & (Middle) & Real \\
\texttt{Cardio}       &	1831 & 176 & 9.60               & 21                                  & (High) & Real \\
\texttt{Glass} & 214& 29 & 13.55& 9& (Middle) & Real \\
\texttt{Ionosphere}   & 351 & 126  &  35.71              & 33                                   & (High) & Real\\
\texttt{Pendigits}    & 6870 & 156  &  2.27             & 16                                 & (Middle) & Real\\
\texttt{Pima}         & 768  & 268  &  34.89             & 8                                    & (Middle) & Real\\
  \texttt{Shuttle}     &   49097 &         3511  &   7.15 & 9                 & (Middle) & Real \\
   \texttt{Wine} & 129	&  10 &  7.75	& 13   & (Middle) & Real \\ 
   \texttt{Diabetes} & 85916 & 8298 & 9.65 & 4 & (Low) & Real \\
   \texttt{Moodify} & 276260 & 42188 & 15.27 & 11 & (Middle) & Real \\
\hline
\end{tabular}
\end{table*}}

\rebuttal{In the following sections, we present the qualitative results for two synthetic datasets: {\ttfamily Xaxis} and {\ttfamily Bisect3D\_Skewed} (Section \ref{sec:synthetic}), along with the real-world dataset {\ttfamily glass} (Section \ref{sec:real-dataset-experiments}). Next, we provide the quantitative results of the AD performance and its interpretation in Sections \ref{sec:perf-tab} and \ref{sec:imp-tab}. Additional graphical results can be found in the repository at \url{https://github.com/alessioarcudi/ExIFFI}. }

\subsection{Synthetic Datasets} \label{sec:synthetic}
We evaluate the performance of \EIFplus and \approach in detecting anomalies and identifying the main features for the detection of anomalous points through the graphical representations illustrated in Section \ref{sec:viz-explanations}.

\rebuttal{Synthetic datasets are designed to showcase the functionality and behavior of the models and explanation algorithms, both in situations where a single feature drives the detection of an anomaly and in cases where multiple features contribute to anomaly identification. By constructing these datasets, we can replicate specific conditions and assess how effectively the model interprets anomalies when the underlying anomaly-generating mechanism is fully understood. This method enables a precise evaluation of the model’s interpretative abilities.}

\rebuttal{The construction of each synthetic dataset is inspired by from the research paper authored by Carletti et al. \cite{diffi}. The inliers are points that lie within a $p$-dimensional sphere centered at the origin with a radius $r$. These inliers are sampled from a uniform distribution in $p$ dimensions, denoted as $\mathcal{U}_p([-r,r])$, and are included in the dataset if their L2 norm is less than $r$. Mathematically, this set of inliers is represented as \(\mathcal{D}_I = \{ x \sim \mathcal{U}_p([-r,r]) \mid ||x||^2 \leq r\}\). This ensures that all inlier points are uniformly distributed within the defined boundary, providing a well-defined structure for the normal data.

Outliers, on the other hand, are generated using a different approach to simulate anomalies. These outliers are uniformly distributed along one or more axes, characterized by \( k \) anomalous features. The remaining \( p-k \) features act as noise, sampled from a Gaussian distribution \(\mathcal{N}(0,1)\). The anomalous points are determined by the following parameters:

\begin{itemize}
    \item A unit vector \( \mathbf{u} \in \mathbb{S}^{n-1} \), where \( \mathbb{S}^{n-1} \) represents the unit sphere in \( \mathbb{R}^n \), and \( u_i \neq 0 \) if \( i \) is an anomalous feature,
    \item A scalar distance \( d \) from the origin,
    \item A random variable \( x \), sampled from a uniform distribution \( \mathbf{x} \sim \mathcal{U}([\text{min}, \text{max}]) \),
    \item A noise vector \( \boldsymbol{\epsilon} \), where each anomalous feature is perturbed by Gaussian noise: \( \epsilon_i \sim \mathcal{N}(0,1) \) if \( i \) is anomalous.
\end{itemize}

Thus, the outlier points are generally defined as:

\[
\mathcal{D}_O = \left\{ d \cdot \mathbf{u} + x \cdot \mathbf{u} + \boldsymbol{\epsilon} \,\middle|\, d \in \mathbb{R}, \, \mathbf{u} \in \mathbb{S}^{n-1}, \, x \sim \mathcal{U}([\text{min}, \text{max}]), \, \epsilon_i \sim \mathcal{N}(0,1) \, \text{for each anomalous feature} \, i \right\}
\]

This equation captures the generation of outliers as a combination of a scaled unit vector, a random displacement, and added noise, reflecting the anomalous behavior in the dataset.
}

\rebuttal{In this section, we examine anomalies in datasets with \( p = 6 \) dimensions, varying the number of anomalous features, \( k \). Two datasets are constructed to represent different scenarios: {\ttfamily{Xaxis}}, where \( k = 1 \), $r=5$, {$d=5$}, min$=0$, max=$5$ and anomalies occur along a single dimension thus \( u = [1,0,0,0,0,0] \), and {\ttfamily{Bisect3D\_Skewed}}, where \( k = 3 \), $r=5$, {$d=5$}, min$=0$, max=$5$ and anomalies are distributed along a skewed vector (\( u = [4,3,2,0,0,0]/\|[4,3,2,0,0,0]\| \)). These datasets are used to illustrate how feature importance scores change as anomalies along one feature or along multiple feature with an order of how much each feature is important to determine the anomalies.}

\rebuttal{For a deeper explanation on the synthetic datasets generation process please refer to the pseudocode presented in the Appendix \ref{alg:syn_data}. The other dataset used in the quantitative evaluation are presented in the \ref{sec:appendix_datasets} and are named {\ttfamily Bisect}, {\ttfamily Bisect3D} and  {\ttfamily Bisect6D}}

\subsection*{Performance} \label{sec:syn_perf}

\rebuttal{
In the {\ttfamily Xaxis} dataset anomalies are restricted to specific intervals along a single feature (feature 0), while the rest of the data points are uniformly spread within a 6-dimensional space. Detecting these anomalies poses a challenge for algorithms like ECOD and Isolation Forest, as they evaluate features individually. In particular in high-dimensional spaces, anomalies confined to one dimension appear normal across others, making them harder to detect. This challenge is particularly evident in the well-known bias of Isolation Forest, pointed out by Hariri et al.\cite{8888179}, where there is a lower anomaly score in the bundles orthogonal to all the axis but one around the data distribution. Since anomalies in this case are aligned with one axis, they are inside the bundle so they have a lower anomaly scores than anomalies with the same distance from the data distribution but in a combination of features. 

This occurs because Isolation Forest as well as ECOD isolates each feature independently, leading to a skewed perception when other features behave normally. This bias motivated the development of the EIF algorithm, which enhances anomaly detection by partitioning the data space based on linear combinations of multiple features, rather than treating them individually. As a result, EIF achieves substantially better performance in detecting anomalies in datasets where the anomalies are spread across a single dimension, such as the {\ttfamily Xaxis} dataset, which Isolation Forest and ECOD struggle to identify.

In {\ttfamily Bisect3D\_Skewed} the anomalies are aligned along the direction \(u = [4,3,2,0,0,0]\), then they are correlated to multiple features. This improves the detection performance of both ECOD and Isolation Forest, as the anomalies exhibit more distinct behavior in the full feature space.

In the two synthetic datasets presented, and in the results of the synthetic datasets in the Table \ref{tab:ComprehensiveConsolidated}, the DIF perform poorly. This is due to the fact that the anomalies are distributed in a specific manner that does not leverage the strengths of DIF. DIF excels in detecting anomalies when there are complex, non-linear relationships in the data. However, when anomalies are confined to a specific feature (like in the feature 0 case) or are linearly aligned (along the direction \([4,3,2,0,0,0]\)), the model may not effectively distinguish them from the normal data because these scenarios might not exhibit the complex interactions between features that DIF is designed to exploit. This leads to the model's inability to properly isolate and detect these anomalies.

{Finally, for what concerns the EIF and \EIFplus models they showcase excellent performances with Average Precision values close to the perfect score of 1, as detailed in Table \ref{tab:ComprehensiveConsolidated}}}


\begin{figure}[!ht] 
    \centering 
    \begin{subfigure}[t]{0.50\textwidth}
      \includegraphics[width=\linewidth]{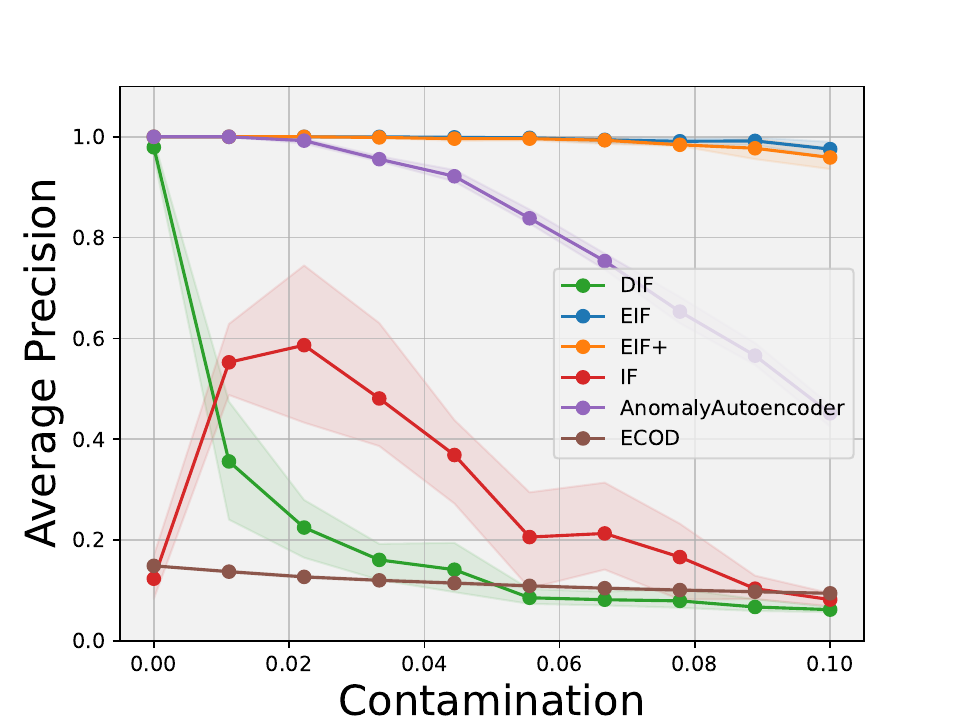}
      \caption{{\ttfamily Xaxis}}
      \label{fig:Xaxis_contamination_full}
    \end{subfigure}\hfil 
    \begin{subfigure}[t]{0.50\textwidth}
      \includegraphics[width=\linewidth]{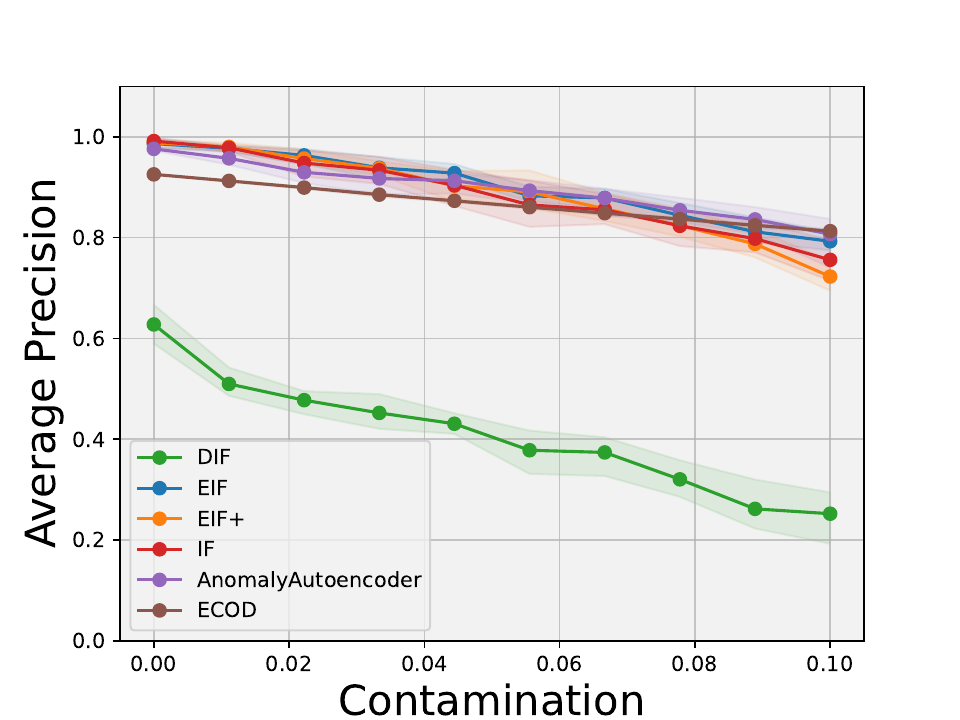}
      \caption{{\ttfamily Bisect3D\_Skewed}}
      \label{fig:Bisect3D_prop_contamination_full}
    \end{subfigure}\hfil 
\caption{\rebuttal{Figures \ref{fig:Xaxis_contamination_full},\ref{fig:Bisect3D_prop_contamination_full} showcase the behavior of the average precision metric as the contamination level in the training set increases for different AD models. In particular Figure \ref{fig:Xaxis_contamination_full} is dedicated to {\ttfamily Xaxis} while \ref{fig:Bisect3D_prop_contamination_full} refers to {\ttfamily Bisect3D\_Skewed}}}
\label{fig:synthetic_contamination}
\end{figure}

\subsection*{Interpretation} 

\begin{figure}[!ht] 
    \centering 
    \begin{subfigure}[t]{0.50\textwidth}
      \includegraphics[width=\linewidth]{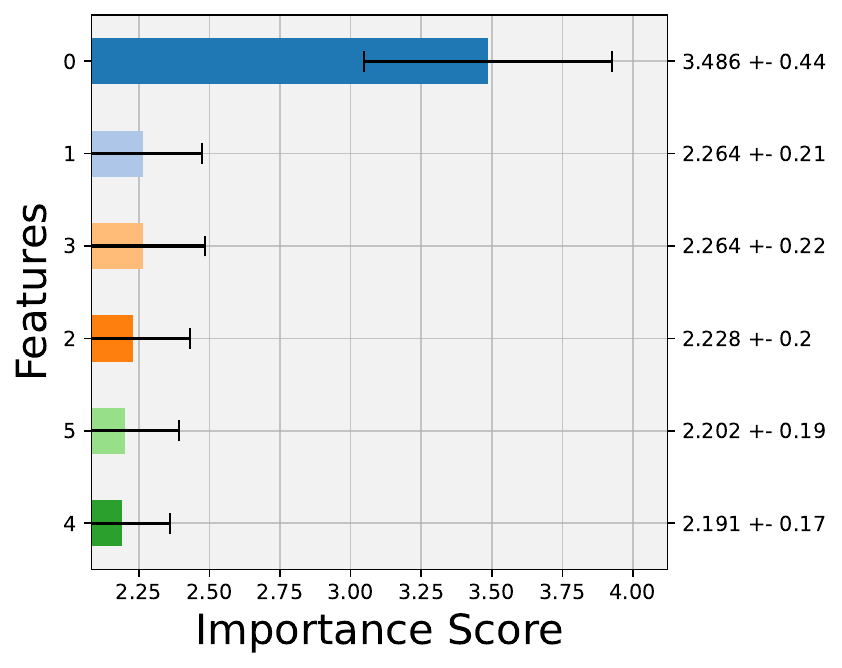}
      \caption{{\ttfamily Xaxis}}
      \label{fig:Xaxis_score_bars}
    \end{subfigure}\hfil 
    \begin{subfigure}[t]{0.50\textwidth}
      \includegraphics[width=\linewidth]{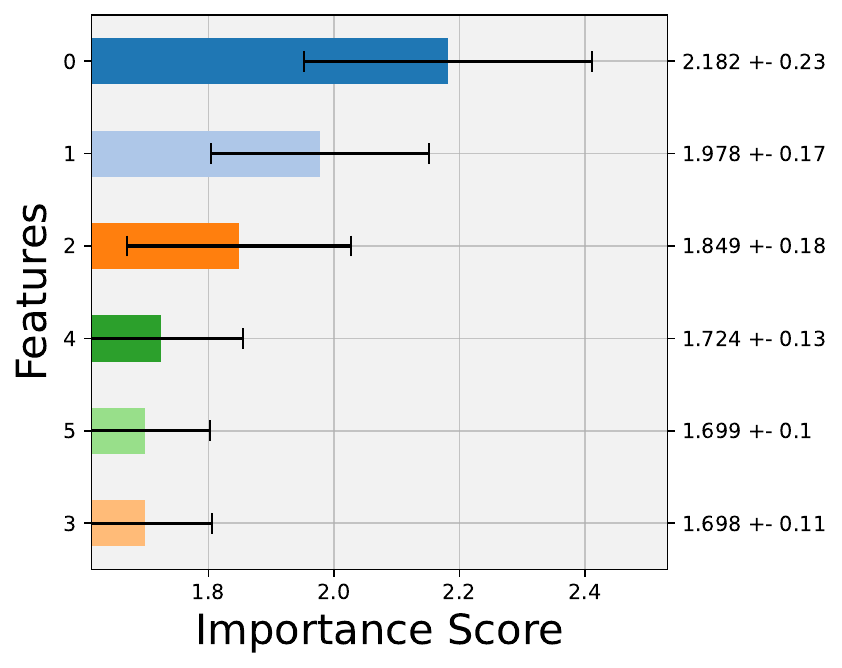}
      \caption{{\ttfamily Bisect3D\_Skewed}}
      \label{fig:Bisect3D_prop_score_bars}
    \end{subfigure}\hfil 

    \begin{subfigure}[t]{0.50\textwidth}
      \includegraphics[width=\linewidth]{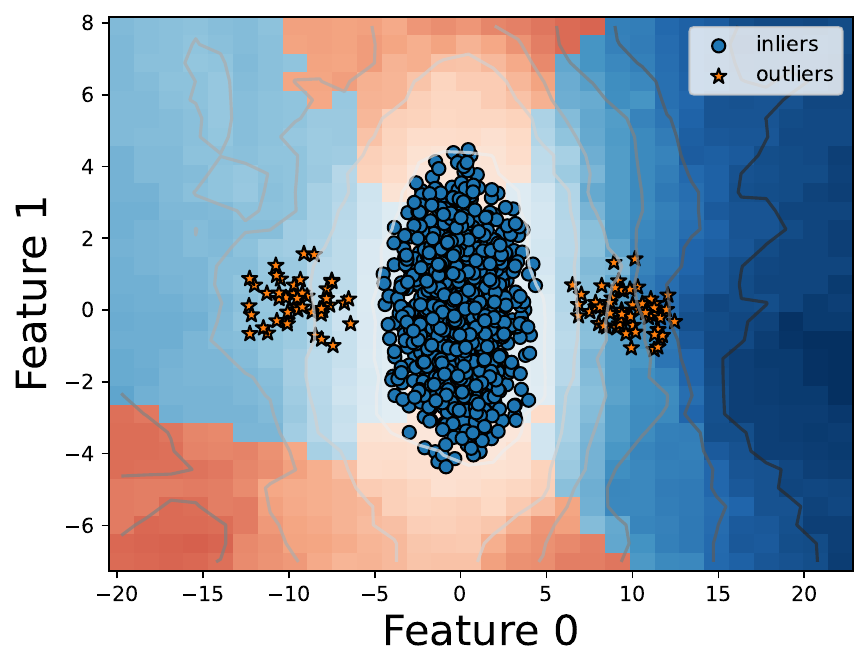}
      \caption{{\ttfamily Xaxis}}
      \label{fig:Xaxis_local_scoremap}
    \end{subfigure}\hfil 
    \begin{subfigure}[t]{0.50\textwidth}
      \includegraphics[width=\linewidth]{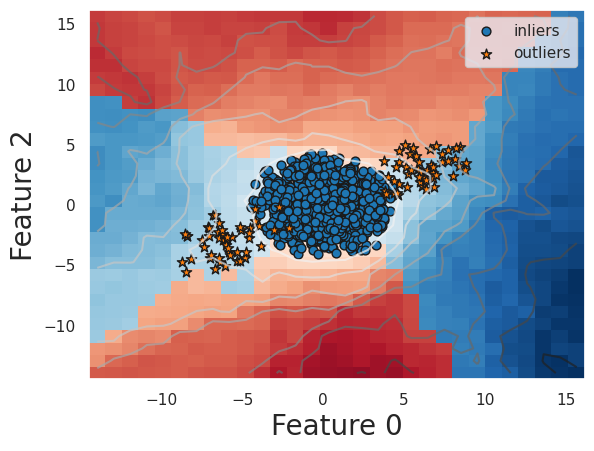}
      \caption{{\ttfamily Bisect3D\_Skewed}}
      \label{fig:Bisect3D_prop_local_scoremap}
    \end{subfigure}\hfil 
\caption{\rebuttal{Figures \ref{fig:Xaxis_score_bars}, \ref{fig:Bisect3D_prop_score_bars} display the Score Plots for the {\ttfamily Xaxis} and {\ttfamily Bisect3D\_Skewed} datasets, respectively. Figures \ref{fig:Xaxis_local_scoremap} and \ref{fig:Bisect3D_prop_local_scoremap} outline, instead, the Local Scoremaps of these two datasets. The feature importances used to generate these plots were obtained using the \approach algorithm applied to the \EIFplus AD model.}}
\label{fig:Xaxis_interpretation}
\end{figure}

\rebuttal{The Figure \ref{fig:Xaxis_interpretation} shows the importance scores of \approach applied to \EIFplus in Scenario II. The bar charts in the top row in Figures \ref{fig:Xaxis_score_bars} and \ref{fig:Bisect3D_prop_score_bars} illustrate the importance of each feature of the two datasets:{\ttfamily Xaxis} and {\ttfamily Bisect3D\_Skewed}. The importance scores not only demonstrate the model's ability to accurately identify the key features contributing to the anomalies, assigning the highest scores to the most relevant features, but also highlight its capability to distinguish between features with varying scores when anomalies are unevenly distributed among them. This nuanced differentiation underscores the model’s effectiveness in capturing the relative significance of each feature based on the specific distribution of anomalies.

In the bottom row, the local importance score maps depict the relative importance of Feature $0$ versus Feature $1$ across different points in a 2D space. The color gradient indicates which feature is more important at each location (red for Feature $1$ and blue for Feature $0$). The maps demonstrate that the model consistently identifies the correct dominant feature across different regions, with clear transitions between blue and red areas, indicating precise differentiation between the influence of Feature $0$ and Feature $1$ in different parts of the data space. This ability to distinguish local importance highlights \approach’s robustness and effectiveness in feature selection under Scenario II.}

\rebuttal{The Table \ref{tab:AUC_FS_tab} presents an evaluation of feature selection methods applied to the two Scenarios presented in Section \ref{sec:performance_evaluation}, highlighting key insights into their performance. The IF-based feature selection (IF\_\approach) yields poor interpretation results in the case of {\ttfamily Xaxis} dataset, which is consistent with the model's overall bad performance, indicating a direct link between the quality of the model and the reliability of the feature importance it generates. Similarly, the Random Forest-based method (IF\_RF) also shows low interpretation results, reinforcing the notion that the accuracy of a model's performance is crucial for effective feature importance evaluation.

On the other hand, the other feature selection methods, including those based on EIF and \EIFplus, demonstrate very similar and strong performance across both scenarios. This consistency suggests what we have seen in Figures \ref{fig:Xaxis_contamination} that these methods successfully identified the most important features, contributing to their reliable performance across different conditions. Notably, the \EIFplus\_\approach and EIF\_RF methods performed particularly well, especially in Scenario II, highlighting their superior generalization ability to unseen anomalies.}

\subsection{Real-World Datasets}\label{sec:real-dataset-experiments}
Real-world data presents a multifaceted benchmarking challenge, mainly due to the intricate distributions and the subtle distinction between inliers and outliers. These complexities often complicate the task of reliably identifying anomalies, which in turn can affect the interpretability of results.
Our experiments aimed to shed light on the efficacy of our models in discerning the features that affect the most the anomalies.

\rebuttal{We analyzed in details the results of the dataset named {\ttfamily Glass} due to its unique characteristics and the insights they provide into anomaly detection and feature interpretation. The {\ttfamily Glass} dataset offers clear definitions of features and anomalous classes, supported by literature that helps validate the correctness of interpretation results. Moreover, this dataset's particular distribution, where some inliers are far from the normal distribution and some outliers are close to inliers, highlights the significant impact of contamination levels on model performance. }

\rebuttal{For a detailed view of the quantitative results of the datasets examined, refer to the Tables \ref{tab:ComprehensiveConsolidated} and \ref{tab:AUC_FS_tab}.}

\subsection*{Performance} \label{sec:real_perf}

\begin{figure}[!ht] 
    \centering 

      \includegraphics[width=0.4\linewidth]{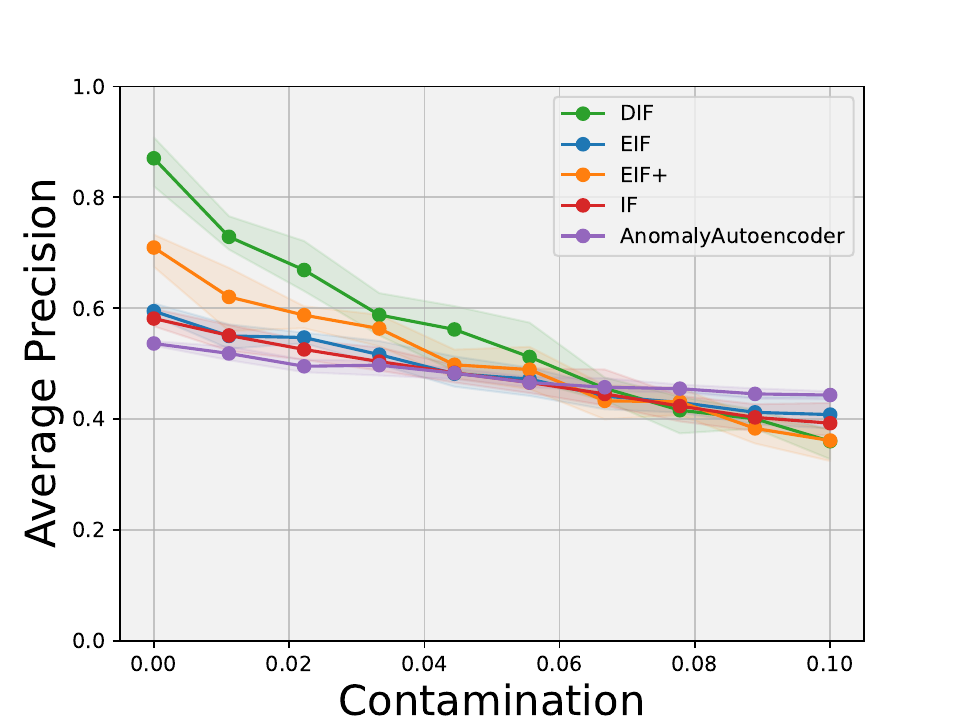}

\caption{\rebuttal{Figure showcases the behavior of the average precision metric as the contamination level in the training set increases for different AD models trained on the {\ttfamily glass} dataset.}}
\label{fig:glass_contamination}
\end{figure}

\rebuttal{Figure \ref{fig:glass_contamination} shows that \EIFplus significantly outperforms EIF when contamination levels are low, highlighting its superior generalization in scenarios where the model is fitted primarily on normal data. This improvement is evident in Figure \ref{fig:glass_scoremap}, where some inliers (blue dots) are far from the distribution and some outliers (marked with orange star) are inside the normal distribution, making it difficult for the model to distinguish between normal and anomalous data when fitted to the entire dataset. By focusing on normal data, \EIFplus better discerns the two, leading to its marked advantage over EIF in these conditions.

A similar pattern can be observed for DIF, which also experiences substantial performance gains when contamination levels are low. While DIF struggles in high contamination scenarios, much like other models, its deep learning architecture allows it to better map data into informative subspaces when contamination decreases. This enables DIF to capture the underlying structure of the normal data more effectively, thereby isolating the remaining anomalies more accurately. However, as noted in another table, DIF lacks robustness; despite achieving high scores with some datasets in particular in Scenario II as it is possible to see in Table \ref{tab:ComprehensiveConsolidated}, its performance drastically drops in other datasets such as the synthetic ones, {\ttfamily breastw}, {\ttfamily diabetes} and {\ttfamily annthyroid}. Additionally, DIF's fit and prediction times analyzed in Section \ref{sec:time_scaling} are significantly longer compared to simpler models like IF and EIF, making it computationally expensive. In contrast, \EIFplus not only improves upon EIF's accuracy but also maintains relatively robust results compared to the benchmark models and very low fit and prediction times, offering a balanced trade-off between performance and computational efficiency.}

\subsection*{Interpretation} 
\rebuttal{
In this study, we chose to present the results from Scenario II due to the significant overlapping of anomalies with inliers in Scenario I, which led to poor performance. The strong overlap in Scenario I made it challenging for the models to accurately distinguish between normal and anomalous instances, resulting in less reliable feature importance interpretations. As a result, the findings from Scenario II, where the model was trained solely on inliers, provide a clearer and more robust analysis. For completeness.}

\begin{figure}[!ht] 
    \centering 
    \begin{subfigure}[t]{0.50\textwidth}
      \includegraphics[width=\linewidth]{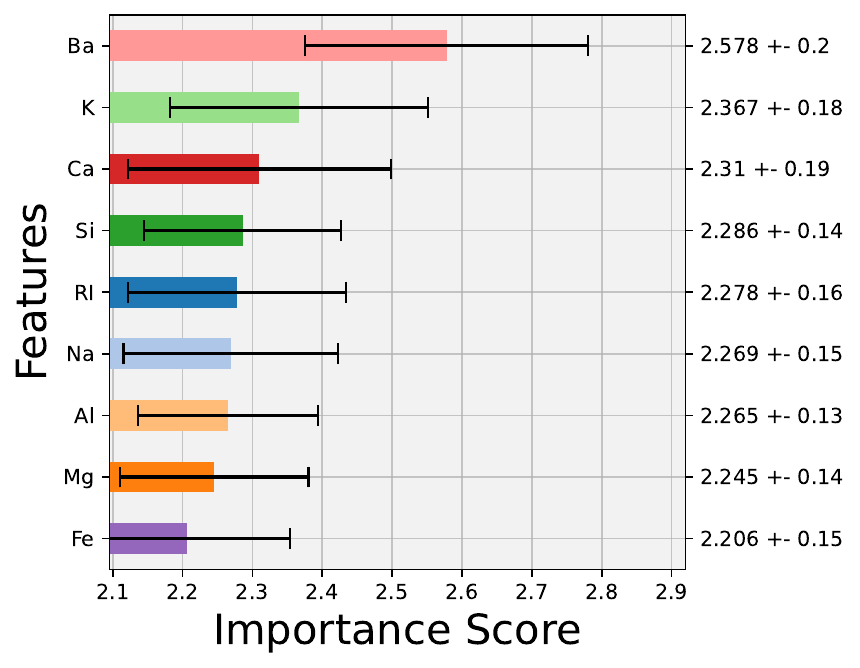}
      \caption{{\ttfamily glass}}
      \label{fig:glass_score_bars}
    \end{subfigure}\hfil 
    \begin{subfigure}[t]{0.50\textwidth}
      \includegraphics[width=\linewidth]{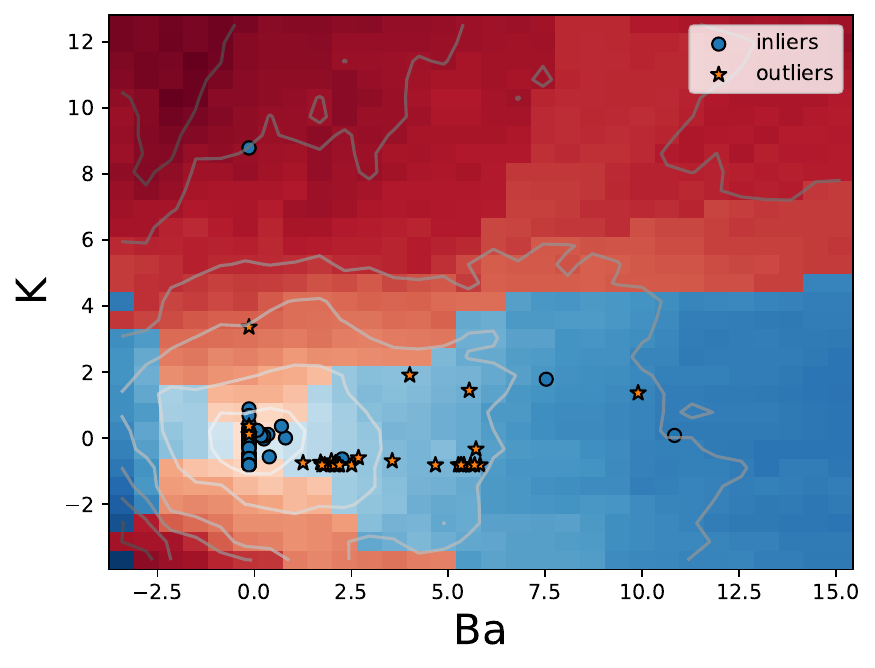}
      \caption{{\ttfamily glass}}
      \label{fig:glass_scoremap}
    \end{subfigure}\hfil 
\caption{\rebuttal{Figures \ref{fig:glass_score_bars}, \ref{fig:glass_scoremap} display the Score Plot and Local Scoremaps for the {\ttfamily glass} dataset. The feature importances used to generate these plots were obtained using the \approach algorithm applied to the \EIFplus AD model.}}
\label{fig:glass_imp_plots}
\end{figure}

\rebuttal{{As observable in Figure \ref{fig:glass_score_bars},} Barium (Ba) is emerged as the most important feature, followed by Potassium (K). This aligns with domain knowledge that we have on the dataset since the anomalous points represent the chemical composition of the headlamp glass, confirming the correctness of the approach, as Barium is known to be the key differentiator between headlamp and window glass \cite{D3RA08015C}. 

The scoremap reveals that deviations in Potassium levels are not strongly correlated with the anomaly distribution. In fact, as shown in Figure \ref{fig:glass_scoremap}, the anomalous points (represented by orange stars) do not deviate from the normal distribution along the Potassium feature. However, some inlier points clearly deviate in their Potassium composition and influence the interpretation output.} 

\subsection{Performance Report} \label{sec:perf-tab}

\begin{table}[ht!]
\centering
\caption{Precision Report: Scenario 1 (S1) and Scenario 2 (S2) performance metrics for various Anomaly Detection models across different datasets.}
\label{tab:ComprehensiveConsolidated}
\resizebox{\textwidth}{!}{
\begin{tabular}{|l|l|l|m{0.9cm} m{0.9cm} m{0.9cm} m{0.9cm} m{0.9cm}|m{0.9cm} m{0.9cm} m{0.9cm} m{0.9cm} m{0.9cm} m{0.9cm} m{0.9cm} m{0.9cm} m{0.9cm} m{0.9cm} m{0.9cm}| m{0.9cm}|}
\hline
\multicolumn{3}{|l|}{}& \multicolumn{5}{c|}{Synthetic}& \multicolumn{11}{c|}{ Real World}&\multicolumn{1}{c|}{ }\\
\hline
\rotatebox[origin=c]{90}{Scenario  }&\rotatebox[origin=c]{90}{Model  }&\rotatebox[origin=c]{90}{Measure  }&
 \rotatebox[origin=c]{90}{{\ttfamily Xaxis}} & \rotatebox[origin=c]{90}{{\ttfamily Bisec}} & \rotatebox[origin=c]{90}{{\ttfamily Bisec3D}} & \rotatebox[origin=c]{90}{{\ttfamily Bisec6D}} & 
 \rotatebox[origin=c]{90}{  {\ttfamily Bisec3D\_Skewed}  } & \rotatebox[origin=c]{90}{{\ttfamily  Annthyroid }} & \rotatebox[origin=c]{90}{{\ttfamily Breastw}} & \rotatebox[origin=c]{90}{{\ttfamily Cardio}} & \rotatebox[origin=c]{90}{{\ttfamily Diabetes}} & \rotatebox[origin=c]{90}{{\ttfamily Glass}} & \rotatebox[origin=c]{90}{{\ttfamily Ionosphere }} & \rotatebox[origin=c]{90}{{\ttfamily Moodify}} & \rotatebox[origin=c]{90}{{\ttfamily Pendigits}} & \rotatebox[origin=c]{90}{{\ttfamily Pima}} & \rotatebox[origin=c]{90}{{\ttfamily Shuttle}} & \rotatebox[origin=c]{90}{{\ttfamily Wine}} & \rotatebox[origin=c]{90}{mean} \\
\hline
\multirow{18}{*}{S1}& \multirow{3}{*}{IF}       & Avg Prec & 0.09 & 0.75 & 0.95 & 0.99 &  0.6& \cellcolor{lightgray} \textbf{0.30}     & \cellcolor{lightgray} \textbf{0.96}     & 0.53     & 0.46     & 0.36     & 0.79     & 0.35     & 0.28     & \cellcolor{lightgray} \textbf{0.51 }    & \cellcolor{lightgray} \textbf{0.98}     & \cellcolor{lightgray} \textbf{0.26} &0.57  \\
                    &                           & Prec     & 0.11 & 0.74 & 0.85 & 0.98 &  0.51& 0.32     & 0.90     & 0.44     & 0.46     & 0.34     & 0.65     & 0.30     & 0.31     & 0.52     & 0.96     & 0.20 & 0,54 \\
                    &                           & ROC AUC  & 0.51 & 0.85 & 0.91 & 0.98 &  0.63& 0.63     & 0.90     & 0.69     & 0.70     & 0.62     & 0.73     & 0.59     & 0.64     & 0.63     & 0.97     & 0.56 &  0.72 \\
                   \cline{2-20}
                    & \multirow{3}{*}{EIF}      & Avg Prec & \cellcolor{lightgray} \textbf{0.97} & 0.97 & 0.96 & 0.96 &  0.77& 0.19     & 0.91     & 0.55     & \cellcolor{lightgray} \textbf{0.47}     & 0.35     & 0.82     & 0.33     & 0.23     & 0.49     & 0.90     & 0.21 & \cellcolor{lightgray} \textbf{0.63} \\
                    &                           & Prec     & 0.89 & 0.91 & 0.89 & 0.92 &  0.71& 0.24     & 0.86     & 0.59     & 0.48     & 0.27     & 0.69     & 0.37     & 0.24     & 0.49     & 0.87     & 0.20 & 0.60 \\
                    &                           & ROC AUC  & 0.93 & 0.95 & 0.93 & 0.95 &  0.84& 0.59     & 0.86     & 0.77     & 0.71     & 0.58     & 0.76     & 0.63     & 0.61     & 0.61     & 0.93     & 0.56 & 0.76 \\
                    \cline{2-20}
                    & \multirow{3}{*}{\EIFplus} & Avg Prec & 0.91 & 0.94 & 0.93 & 0.93 &  0.75& 0.20     & 0.88     & 0.50     & 0.44     & 0.31     & 0.84     & 0.28     & 0.21     & 0.49     & 0.70     & 0.17 & 0.59 \\
                    &                           & Prec     & 0.89 & 0.83 & 0.71 & 0.81 &  0.64& 0.28     & 0.85     & 0.52     & 0.49     & 0.24     & 0.72     & 0.27     & 0.41     & 0.50     & 0.71     & 0.00 & 0.55  \\
                    &                           & ROC AUC  & 0.93 & 0.90 & 0.84 & 0.89 &  0.8& 0.61     & 0.84     & 0.73     & 0.72     & 0.56     & 0.78     & 0.57     & 0.69     & 0.61     & 0.84     & 0.45 & 0.73 \\
                    \cline{2-20}
                    & \multirow{3}{*}{DIF}      & Avg Prec & 0.06 & 0.16 & 0.30 & 0.19 &  0.25& 0.21     & 0.44     & 0.58     & 0.10     & 0.30     & \cellcolor{lightgray} \textbf{0.87}     & 0.26     & \cellcolor{lightgray} \textbf{0.34}     & 0.41     & 0.54     & 0.07 & 0.31 \\
                    &                           & Prec     & 0.06 & 0.20 & 0.28 & 0.16 &  0.32& 0.26     & 0.49     & 0.52     & 0.90     & 0.20     & 0.76     & 0.26     & 0.39     & 0.41     & 0.60     & 0.00 & 0.36 \\
                    &                           & ROC AUC  & 0.48 & 0.56 & 0.60 & 0.53 &  0.62& 0.60     & 0.46     & 0.73     & 0.49     & 0.54     & 0.81     & 0.56     & 0.68     & 0.54     & 0.78     & 0.45 & 0.59 \\
                    \cline{2-20}
                    & \multirow{3}{*}{AE}       & Avg Prec & 0.52 & 0.89 &\cellcolor{lightgray} \textbf{ 0.97} & 0.97 &  \cellcolor{lightgray} \textbf{0.83}& 0.21     & 0.61     & 0.62     & 0.45     & \cellcolor{lightgray} \textbf{0.41}     & 0.76     & 0.45     & 0.22     & 0.43     & 0.90     & 0.14 & 0.59 \\
                    &                           & Prec     & 0.46 & 0.82 & 0.92 & 0.95 &  0.78& 0.25     & 0.44     & 0.62     & 0.47     & 0.44     & 0.60     & 0.53     & 0.33     & 0.44     & 0.95     & 0.10 & 0.57 \\
                    &                           & ROC AUC  & 0.70 & 0.90 & 0.95 & 0.97 &  0.87& 0.59     & 0.41     & 0.79     & 0.70     & 0.68     & 0.68     & 0.72     & 0.65     & 0.57     & 0.97     & 0.51 & 0.73 \\
                    \cline{2-20}
                    &\multirow{3}{*}{ECOD}       & Avg Prec &  0.1&  \cellcolor{lightgray} \textbf{0.99}&  \cellcolor{lightgray} \textbf{0.97}&  \cellcolor{lightgray} \textbf{1.00}&  0.82&  0.26&  0.85&  \cellcolor{lightgray} \textbf{0.65}&  0.29&  0.38&  0.56&  \cellcolor{lightgray} \textbf{0.5}&  0.26&  0.46&  0.91&   0.19& 0.57 \\
                    &                           & Prec     &  0.1&  0.99&  0.9&  1.00&  0.75&  0.3&  0.75&  0.53&  0.35&  0.41&  0.54&  0.48&  0.35&  0.45&  0.87&   0.1& 0.55 \\
                    &                           & ROC AUC  &  0.5&  0.99&  0.94&  1.00&  0.86&  0.62&  0.74&  0.74&  0.64&  0.66&  0.64&  0.69&  0.67&  0.58&  0.93&   0.51& 0.73 \\
\hline
\hline
\multirow{18}{*}{S2}& \multirow{3}{*}{IF}       & Avg Prec & 0.10 & 0.99 &\cellcolor{lightgray} \textbf{1.00} &\cellcolor{lightgray} \textbf{1.00} &  \textbf{0.98}& 0.45     &\cellcolor{lightgray} \textbf{0.99}& 0.68     & 0.26     & 0.65     & 0.89     &\cellcolor{lightgray} \textbf{0.72}& 0.37     &\cellcolor{lightgray} \textbf{0.58}& 0.98     & 0.61 &  0.70\\
                    &                           & Prec     & 0.00 & 0.95 & 1.00 & 1.00 &  0.96& 0.43     & 0.94     & 0.60     & 0.31     & 0.68     & 0.79     & 0.73     & 0.42     & 0.55     & 0.97     & 0.50 & 0.68 \\
                    &                           & ROC AUC  & 0.45 & 0.97 & 1.00 & 1.00 &  0.97& 0.69     & 0.94     & 0.78     & 0.62     & 0.82     & 0.83     & 0.84     & 0.70     & 0.66     & 0.98     & 0.72 &0.81\\
                    \cline{2-20}
                    & \multirow{3}{*}{EIF}      & Avg Prec & 0.99 & 0.99 & 0.99 & 0.99 &  0.97& 0.45     & 0.98     & 0.74     & 0.55     & 0.57     & 0.90     & 0.65     & 0.27     & 0.54     & 0.97     & 0.57 & 0.76 \\
                    &                           & Prec     & 1.00 & 1.00 & 1.00 & 0.99 &  0.98& 0.41     & 0.93     & 0.72     & 0.61     & 0.68     & 0.82     & 0.64     & 0.36     & 0.57     & 0.95     & 0.60 & 0.77 \\
                    &                           & ROC AUC  & 1.00 & 1.00 & 1.00 & 0.99 &  0.98& 0.68     & 0.93     & 0.84     & 0.78     & 0.82     & 0.86     & 0.79     & 0.67     & 0.67     & 0.97     & 0.78 & 0.86 \\
                    \cline{2-20}
                    & \multirow{3}{*}{\EIFplus} & Avg Prec & 0.99 &\cellcolor{lightgray} \textbf{1.00} & 0.99 & 0.99 &  0.97&\cellcolor{lightgray} \textbf{0.46}&\cellcolor{lightgray} \textbf{0.99}& 0.76     &\cellcolor{lightgray} \textbf{0.61}& 0.69     &\cellcolor{lightgray} \textbf{0.96}& 0.66     & 0.36     &\cellcolor{lightgray} \textbf{0.58}& 0.91     &\cellcolor{lightgray} \textbf{0.79} &\cellcolor{lightgray} \textbf{0.80} \\
                    &                           & Prec     & 1.00 & 1.00 & 0.99 & 1.00 &  0.95& 0.46     & 0.95     & 0.73     & 0.56     & 0.79     & 0.87     & 0.66     & 0.45     & 0.57     & 0.96     & 0.70 & 0.79 \\
                    &                           & ROC AUC  & 1.00 & 1.00 & 0.99 & 1.00 &  0.97& 0.70     & 0.95     & 0.85     & 0.76     & 0.88     & 0.90     & 0.80     & 0.72     & 0.67     & 0.97     & 0.83 & 0.87 \\
                    \cline{2-20}
                    & \multirow{3}{*}{DIF}      & Avg Prec & 0.97 & 0.85 & 0.79 & 0.36 &  0.6& 0.39     & 0.60     & 0.79     & 0.13     &\cellcolor{lightgray} \textbf{0.85}&\cellcolor{lightgray} \textbf{0.96}&\cellcolor{lightgray} \textbf{0.72}&\cellcolor{lightgray} \textbf{0.46}& 0.49     &\cellcolor{lightgray} \textbf{0.99}& 0.72 & 0.67\\
                    &                           & Prec     & 0.90 & 0.78 & 0.72 & 0.32 &  0.51& 0.39     & 0.66     & 0.70     & 0.17     & 0.82     & 0.89     & 0.73     & 0.53     & 0.46     & 0.97     & 0.70 & 0.64 \\
                    &                           & ROC AUC  & 0.94 & 0.87 & 0.84 & 0.62 &  0.73& 0.67     & 0.64     & 0.83     & 0.54     & 0.90     & 0.90     & 0.84     & 0.76     & 0.59     & 0.98     & 0.83 & 0.78 \\
                    \cline{2-20}
                    & \multirow{3}{*}{AE}       & Avg Prec &\cellcolor{lightgray} \textbf{1.00} &\cellcolor{lightgray} \textbf{1.00} &\cellcolor{lightgray} \textbf{1.00} &\cellcolor{lightgray} \textbf{1.00} &  0.95& 0.44     & 0.98     &\cellcolor{lightgray} \textbf{0.83}& 0.50     & 0.53     & 0.87     & 0.65     & 0.24     & 0.50     & 0.90     & 0.45 & 0.74 \\
                    &                           & Prec     & 1.00 & 1.00 & 1.00 & 1.00 &  0.9& 0.42     & 0.94     & 0.70     & 0.52     & 0.62     & 0.75     & 0.68     & 0.33     & 0.54     & 0.94     & 0.50 & 0.74 \\
                    &                           & ROC AUC  & 1.00 & 1.00 & 1.00 & 1.00 &  0.94& 0.69     & 0.93     & 0.76     & 0.73     & 0.78     & 0.80     & 0.81     & 0.66     & 0.64     & 0.97     & 0.72 & 0.84 \\
                    \cline{2-20}
                    &\multirow{3}{*}{ECOD}       & Avg Prec &  0.14&  0.97&  0.99&  \textbf{1.00}&  0.92&  0.32&  0.97&  0.64&  0.4&  0.45&  0.7&  0.58&  0.28&  0.5&  0.95&   0.25& 0.63 \\
                    &                           & Prec     &  0.15&  0.9&  0.99&  1.00&  0.86&  0.32&  0.91&  0.7&  0.44&  0.55&  0.56&  0.55&  0.35&  0.51&  0.93&   0.2& 0.62 \\
                    &                           & ROC AUC  &  0.53&  0.94&  0.99&  1.00&  0.92&  0.63&  0.91&  0.67&  0.69&  0.74&  0.66&  0.73&  0.67&  0.62&  0.96&   0.56& 0.76 \\
\hline
\end{tabular}
}
\end{table}

\ale{Table \ref{tab:ComprehensiveConsolidated} examinates the perfomance achieved by different AD models across various datasets. The analysis focuses on Average Precision (Avg Prec), Precision (Prec), and ROC-AUC score metrics, under the two different scenarios; with outliers in the training dataset (S1) and without outliers (S2).}

\rebuttal{In S1 even if for each dataset the most performing model vary, the EIF showcases superior Avg Prec, Prec, and ROC AUC scores on average. This improvement underscores EIF's enhanced anomaly detection capability, particularly in handling complex data structures, it is a model that robustly have high performances across all the possible datasets. However, the \EIFplus model does not outperform EIF within this scenario, indicating its modifications do not significantly impact performance under standard test conditions, it remains the second most performing model, toghether with AE.
S2's findings mark a distinct improvement in \EIFplus's performance over EIF, notably in real-world datasets. In fact \EIFplus achieves parity or surpasses EIF in Avg Prec and Prec metrics in all the dataset but {\ttfamily Shuttle}, demonstrating its refined ability to detect anomalies. This enhanced performance is attributable to \EIFplus's algorithmic adjustments, optimized to grasp the unseen outliers within complex data distributions more effectively than EIF.
Moreover \EIFplus demonstrate enhanced performances by figuring as the most performing model in 6 out of 11 real world datasets and reaching very high performances in synthetic ones. Thus it results as the most performing model on average across the datasets.}

\begin{figure}[!ht] 
    \centering 
      \includegraphics[width=0.95\linewidth]{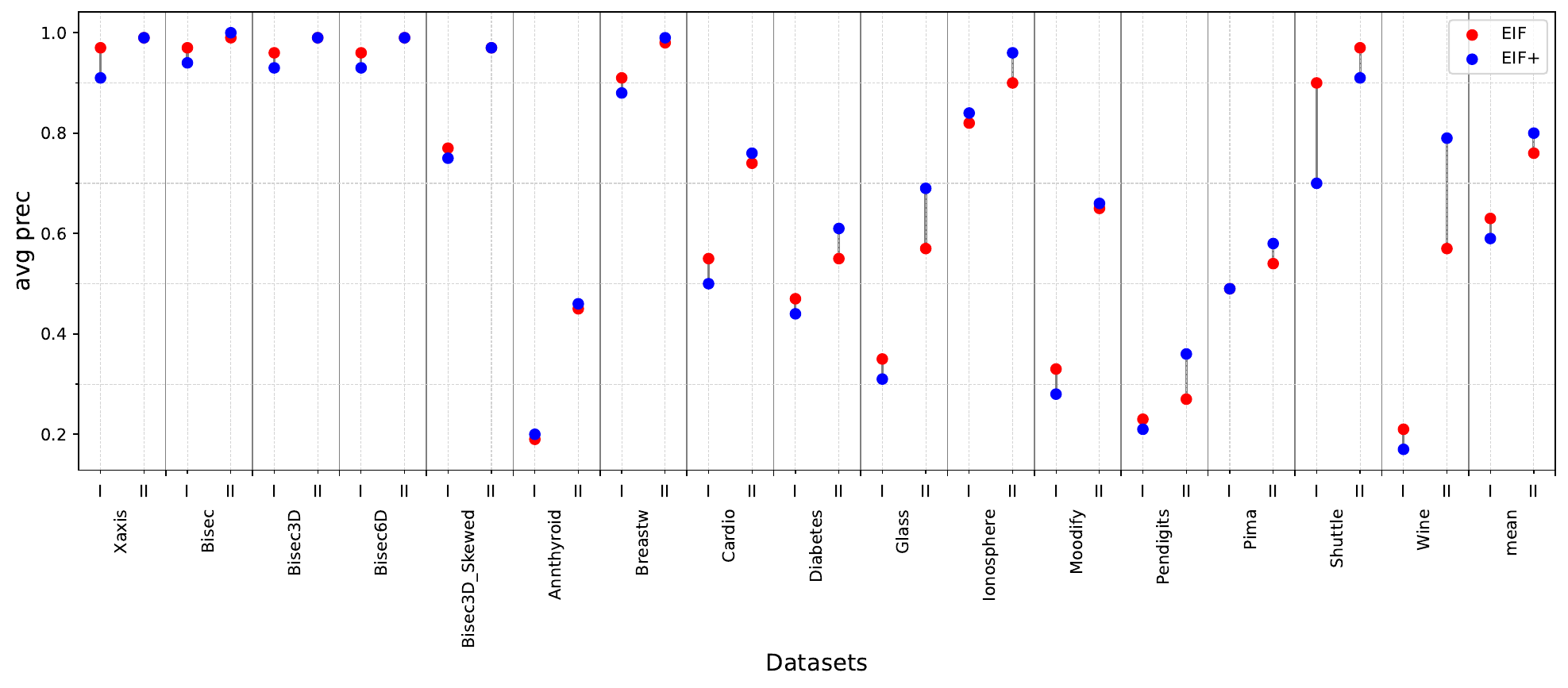}
      \label{fig:glass_contamination_full}
\caption{Average precision values for EIF and \EIFplus across multiple datasets in two scenarios and the mean precision over all datasets. Results for both Scenario I and Scenario II are displayed for each dataset.}
\label{fig:eifvseifplus}
\end{figure}

\rebuttal{In Figure \ref{fig:eifvseifplus} we observe {how in} Scenario I the results indicate that EIF performs slightly better than \EIFplus across most datasets. This is reflected by marginally higher average precision values for EIF in the majority of the datasets. Although the difference is not substantial, it suggests that EIF has a small performance edge under these specific conditions. Both algorithms appear to be competitive in this scenario, with only slight variations in their performance.

In contrast, Scenario II demonstrates a significant improvement in the performance of \EIFplus, which outperforms EIF across almost all datasets except for the {\ttfamily shuttle} dataset, where EIF maintains a slight advantage. The mean results summarize this trend, showing higher average precision values for \EIFplus in Scenario II. This suggests that \EIFplus exhibits greater generalization capability when the anomalies are not in the training set, as reflected by its overall better performance across the majority of datasets in this scenario.}

\subsection{Importance Score} \label{sec:imp-tab}

\begin{table}[ht!]
\caption{$AUC_{FS}$ metric values for different interpretation algorithms with Average Precision evaluated with EIF and \EIFplus in Scenario I (S1) and Scenario II (S2) across different datasets. The highest values are highlighted in bold.}
\label{tab:AUC_FS_tab}
\resizebox{\textwidth}{!}{
\begin{tabular}{|l|l|l|m{0.9cm} m{0.9cm} m{0.9cm} m{0.9cm} m{0.9cm}|m{0.9cm} m{0.9cm} m{0.9cm} m{0.9cm} m{0.9cm} m{0.9cm} m{0.9cm} m{0.9cm} m{0.9cm} m{0.9cm} m{0.9cm}|m{0.9cm}|}
\hline
\multicolumn{3}{|l|}{}& \multicolumn{5}{c|}{Synthetic}& \multicolumn{11}{c|}{ Real World}& \multicolumn{1}{c|}{ }\\
\hline
\rotatebox[origin=c]{90}{Scenario  }&\rotatebox[origin=c]{90}{Model  }&\rotatebox[origin=c]{90}{Measure  }&
 \rotatebox[origin=c]{90}{{\ttfamily Xaxis}} & \rotatebox[origin=c]{90}{{\ttfamily Bisec}} & \rotatebox[origin=c]{90}{{\ttfamily Bisec3D}} & \rotatebox[origin=c]{90}{{\ttfamily Bisec6D}} & 
 \rotatebox[origin=c]{90}{  {\ttfamily Bisec3D\_Skewed}  } & \rotatebox[origin=c]{90}{{\ttfamily  Annthyroid }} & \rotatebox[origin=c]{90}{{\ttfamily Breastw}} & \rotatebox[origin=c]{90}{{\ttfamily Cardio}} & \rotatebox[origin=c]{90}{{\ttfamily Diabetes}} & \rotatebox[origin=c]{90}{{\ttfamily Glass}} & \rotatebox[origin=c]{90}{{\ttfamily Ionosphere }} & \rotatebox[origin=c]{90}{{\ttfamily Moodify}} & \rotatebox[origin=c]{90}{{\ttfamily Pendigits}} & \rotatebox[origin=c]{90}{{\ttfamily Pima}} & \rotatebox[origin=c]{90}{{\ttfamily Shuttle}} & \rotatebox[origin=c]{90}{{\ttfamily Wine}}& \rotatebox[origin=c]{90}{ Mean} \\
\hline
\multirow{16}{*}{S1} & \multirow{8}{*}{EIF} & DIFFI     & \cellcolor{lightgray}\textbf{4.73}& 4.12& 3.42      & 0.02 &   2.92& 0.96     & 0.11& 4.65& -0.36&  \cellcolor{lightgray}\textbf{1.59}& 2.36& -1.48& 0.11& -0.48& -0.51& 0.17&1.40\\
    &  & IF\_\approach     &  -4.72&  4.13& 3.45&  0.18 &  3.62& \cellcolor{lightgray}\textbf{1.12}& \cellcolor{lightgray}\textbf{0.25}&  1.93&  0.84&  0.30& \cellcolor{lightgray}\textbf{3.29}&  -1.68&2.19&  -0.31& \cellcolor{lightgray}\textbf{2.76}& \cellcolor{lightgray}\textbf{0.68}&1.12\\
     &        & EIF\_\approach     &  4.71& \cellcolor{lightgray}\textbf{4.15}&  3.39&  0.08&  \cellcolor{lightgray}\textbf{3.67}&  0.98     &  0.04     & \cellcolor{lightgray}\textbf{5.41}&  0.96&  0.46&  1.82&  -1.78&  1.58&  -0.31&  -1.43&  0.45&\cellcolor{lightgray}\textbf{1.51}\\
    &         & \EIFplus\_\approach & 4.71& 4.07& 3.42& 0.16&  3.63& 0.98    & 0.07& 4.42&\cellcolor{lightgray}\textbf{1.05}& 0.40& 0.61&  -1.51& 1.86& -0.30& 0.45& 0.12&\cellcolor{lightgray}\textbf{1.51}\\
    &         &  IF\_RF   & -2.82& 4.01& 3.41& \cellcolor{lightgray}\textbf{0.19}&   3.39& 0.45     & -0.14     & 1.48& -0.30& -1.73& -1.91&  -0.80& 1.48& -0.36& 0.44& -3.09&0.23\\
    &         &  EIF\_RF   & 4.72& 4.05& 3.44& 0.16&   3.59& -0.68     & -0.18     & 1.35& 0.49& -1.71& -1.97& -0.84& -0.70& -0.44& 1.69& -2.41&0.66\\
     &        &\EIFplus\_RF   & 4.72& 4.02& 3.39& 0.16&   3.58& -1.19     & -0.16     & 0.32& 0.46& -1.46& -1.87& -1.88& -1.40& -0.28& 1.90& -2.81&0.46\\
    &        &ECOD   &  -4.7&  3.35&  \cellcolor{lightgray}\textbf{3.58}&  0.21&  3.63&  -0.10&  0.09&  3.09&  -1.01&  -0.42&  -0.61&  \cellcolor{lightgray}\textbf{2.50}&  \cellcolor{lightgray}\textbf{2.56}&   \cellcolor{lightgray}\textbf{-0.07}&  2.20& -0.03&0.89\\ 
\cline{2-20}
                    & \multirow{8}{*}{\EIFplus}& DIFFI     & 4.54& \cellcolor{lightgray}\textbf{4.16}& 3.32& -0.02&   2.75& 0.91     & 0.08    & 4.39& -0.34& \cellcolor{lightgray}\textbf{1.55}& -1.13& -1.11& 0.23& -0.49& -0.69& 0.43&1.16\\
                    & & IF\_\approach &  -4.49 &  4.12 &  3.46 & \cellcolor{lightgray}\textbf{0.18} &  3.11& \cellcolor{lightgray}\textbf{1.10} & \cellcolor{lightgray}\textbf{0.28} &  2.40&  0.81 &  0.48 & \cellcolor{lightgray}\textbf{2.68} &  -1.16 &  2.17 & \cellcolor{lightgray}\textbf{-0.20} & \cellcolor{lightgray}\textbf{2.78} & \cellcolor{lightgray}\textbf{0.62} &1.15\\
                    & & EIF\_\approach &  4.55 &  4.04 &  3.38 &  0.05 &  \cellcolor{lightgray}\textbf{3.23}&  0.93 &  0.02 & \cellcolor{lightgray}\textbf{5.12} & \cellcolor{lightgray}\textbf{0.82} &  0.56 &  1.61 &  -1.37 &  1.76 &  -0.32 &  1.22 &  0.45 &\cellcolor{lightgray}\textbf{1.63}\\
                    & & \EIFplus\_\approach &  4.52 &  4.05 & \cellcolor{lightgray}\textbf{3.48} &  .004 &  3.16&  0.97 &  0.08 &  4.25 &  0.74 &  0.52 &  0.45 &  -0.99 &  1.92 &  -0.27 &  0.73 &  0.18 &1.49\\
                    &         &  IF\_RF   & -2.68& 3.93& 3.44&0.07&   3.00& 0.42     & -0.14     & 1.28& -0.30& -1.31& 1.76& -0.72& 1.37& -0.30& 0.36& -2.33&0.49\\
                    &         &  EIF\_RF   & 4.56& 3.95& 3.43& 0.05&   3.11& -0.67     & -0.16     & 1.35& 0.38& -1.51& -1.85& -0.71& -0.36& -0.41& 1.73& -1.57&0.70\\
                     &        &\EIFplus\_RF   & \cellcolor{lightgray}\textbf{4.57}& 4.01& 3.40& 0.08&   3.16& -1.16     & -0.14     & 0.36& 0.40& -1.30& -1.56& -1.43& -0.97& -0.28& 1.62& -2.05&0.54\\
                    &        &ECOD   &  -4.47&  3.31&  3.42&  -0.04&  3.19&  -0.12&  - 0.01&  2.82&  -0.71&  -0.11&  -0.34&  \cellcolor{lightgray}\textbf{2.07}&  \cellcolor{lightgray}\textbf{2.71}&   \cellcolor{lightgray}\textbf{-0.12}&  2.72& 0.20&0.91\\
\hline
\multirow{16}{*}{S2} & \multirow{8}{*}{EIF} & 
                            DIFFI     & 2.83& 2.92& 2.79   & -0.04&   1.50& 2.04     & 0.06    & 4.76& -1.45& 2.11& -1.33& -3.66& -0.38& 0.25& -1.48& -3.79&0.45\\
                            & & IF\_\approach &  -4.73 & \cellcolor{lightgray}\textbf{3.84} &  3.01 &  0.06 &  \cellcolor{lightgray}\textbf{4.07}& \cellcolor{lightgray}\textbf{2.25} &  0.08 &  3.40 &  -1.01 & \cellcolor{lightgray}\textbf{3.45} &  0.89 & \cellcolor{lightgray}\textbf{4.39} & 2.68 & \cellcolor{lightgray}\textbf{0.37} & \cellcolor{lightgray}\textbf{2.88} &  8.07 &2.10\\
                            & & EIF\_\approach &  4.72 &  \cellcolor{lightgray}\textbf{3.84} &  3.01 & 0.11 &  4.05&  2.03 &  0.06 &  8.28 &  0.35 &  2.55 & \cellcolor{lightgray}\textbf{4.31} &  -0.35 &  1.88 &  -0.21 &  0.76 &  7.12 &\cellcolor{lightgray}\textbf{2.65}\\
                            & & \EIFplus\_\approach & \cellcolor{lightgray}\textbf{4.74} &  3.82 &  3.02 &  0.07 &  3.99&  2.18 &  0.02 & \cellcolor{lightgray}\textbf{8.60} & \cellcolor{lightgray}\textbf{0.82} &  1.76 &  2.26 &  -0.23 &  1.22 &  -0.03 &  0.01 & \cellcolor{lightgray}\textbf{9.00} &2.58\\
                            &         &  IF\_RF   & -2.85& 3.82& 3.05& 0.10&   4.04& 2.05     & 0.09     & 7.69& -0.05& 2.18& -1.23& 2.07& 1.28& 0.19& 0.90& 4.41&1.73\\
                            &         &  EIF\_RF   & 4.73& 3.83& 3.03& 0.06&   4.05& 2.03     & \cellcolor{lightgray}\textbf{0.10}   & 7.78& 0.31& -1.63& 1.33& 1.63& -0.84& 0.12& 2.74& 3.94&2.06\\
                             &        &\EIFplus\_RF   & \cellcolor{lightgray}\textbf{4.74}& 3.83& 3.04& 0.04&   4.06& 2.09     & \cellcolor{lightgray}\textbf{0.10}    & 8.24& 0.33& 1.94& 3.58& 1.47& -0.24& 0.06& 2.13& 4.87&2.52\\
                            &        &ECOD   &  -4.73&  3.02&  \cellcolor{lightgray}\textbf{3.05}&  \cellcolor{lightgray}\textbf{0.12}&  \cellcolor{lightgray}\textbf{4.07}&  -0.25&  0.07&  6.26&  -0.69&  -0.71&  -2.54&  4.29&  \cellcolor{lightgray}\textbf{2.82}&   0.01&  2.84& -1.59&1.00\\
\cline{2-20}
                   & \multirow{8}{*}{\EIFplus}& 
                            DIFFI     & 2.728    & 2.92& 2.83& -0.12&   1.52& 2.00& 0.06     & 4.33& -1.48& 2.44& 1.92& -3.36& -0.76& 0.30& -1.45& -3.49&0.65\\
                            & & IF\_\approach &  - 4.74&  3.82 &  3.01 &  0.09 & 4.03& \cellcolor{lightgray}\textbf{2.23} &  0.08 &  3.40 &  -0.93 & \cellcolor{lightgray}\textbf{3.41} &  0.26 & 4.23 & \cellcolor{lightgray}\textbf{3.18} & \cellcolor{lightgray}\textbf{0.36} & \cellcolor{lightgray}\textbf{3.07} & \cellcolor{lightgray}\textbf{9.40} &2.18\\
                            & & EIF\_\approach &  \cellcolor{lightgray}\textbf{4.74} &  3.83 &  3.02 &  \cellcolor{lightgray}\textbf{0.13} & \cellcolor{lightgray}\textbf{4.04}&  1.92 & 0.08 &  8.38 &  0.31 &  2.43 & \cellcolor{lightgray}\textbf{2.84} &  -0.20 &  2.23 &  -0.20 &  0.31 &  8.12 &\cellcolor{lightgray}\textbf{2.62}\\
                            & & \EIFplus\_\approach & \cellcolor{lightgray}\textbf{4.74} &  3.83 &  \cellcolor{lightgray}\textbf{3.03} &  0.08 & 4.03&  2.17 &  0.06 & \cellcolor{lightgray}\textbf{8.73} & \cellcolor{lightgray}\textbf{1.17} &  2.26 &  1.41 &  -0.04 &  1.32 &  -0.02 &  -0.13 &  9.25 &\cellcolor{lightgray}\textbf{2.62}\\
                            &         &  IF\_RF   & -2.85& 3.83& \cellcolor{lightgray}\textbf{3.03}& 0.11&    \cellcolor{lightgray}\textbf{4.04}& 1.99     & 0.06     & 7.88& -0.12& 2.28& -1.08& 2.48& 1.38& 0.21& 0.64& 4.53&1.78\\
                            &         &  EIF\_RF   & \cellcolor{lightgray}\textbf{4.74}& \cellcolor{lightgray}\textbf{3.84}& 3.01& 0.06&    4.02& 1.97     & \cellcolor{lightgray}\textbf{0.09}     & 7.85& 0.35& -1.64& 0.69& 2.17& -1.28& 0.17& 2.11& 4.36&2.03\\
                             &        &\EIFplus\_RF   & \cellcolor{lightgray}\textbf{4.74}& 3.83& 3.01& 0.07&    4.01& 2.02     & 0.07     & 8.30 & 0.29& 1.63& 2.17& 2.08& -0.74& 0.14& 2.11& 5.64&2.46\\
                            &        &ECOD   &  -4.74&  3.03&  \cellcolor{lightgray}\textbf{3.03}&  0.07&  \cellcolor{lightgray}\textbf{4.04}&  -0.16&  0.07&  6.04&  -0.37&  0.17&  -2.06&  \cellcolor{lightgray}\textbf{4.57}&  2.75&  0.05&  2.90& -0.60&1.17\\
\hline
\end{tabular}}
\end{table}

\rebuttal{Table \ref{tab:AUC_FS_tab} presents the performance of the newly introduced $AUC_{FS}$ metric on 16 benchmark datasets in the two different Scenarios and the interpretations are evaluated with the EIF model or the \EIFplus. As detailed in \ref{sec:experimental_results}, \approach is benchmarked against DIFFI, an ad-hoc interpretability algorithm of the IF, post-hoc interpretability approach utilizing feature importance from a Random Forest surrogate model and the intrinsically interpretable model ECOD.

For synthetic datasets, the $AUC_{FS}$ scores are generally high across all models showing that all the models found the right anomalous features, with the notable exception of the {\ttfamily Xaxis} dataset in the cases of IF\_\approach, IF\_RF and ECOD present wrong results due to the lack of precision of the underlying models.}

In the case of real-world datasets, the unique definitions of inliers and outliers complicate the interpretability performance assessment of the models. This issue is especially evident in the {\ttfamily Breastw}, {\ttfamily Pima}, and {\ttfamily Moodify} datasets, where the $AUC_{FS}$ values are similar and predominantly negative.

For the other eight datasets, the data supports the effectiveness of \approach, which consistently records the highest $AUC_{FS}$ values in nearly all datasets and scenarios, except for the {\ttfamily glass} dataset in Scenario 1. \rebuttal{In \approach applied to EIF or \EIFplus have the highest average score in any context.}

The analysis also reveals that the most effective interpretations by \approach are generally those associated to the most performing isolation forest model, indicating that the $AUC_{FS}$ metric values are strongly correlated with the efficacy of a specific prediction in a given dataset and scenario (i.e., Scenario I or Scenario II).

In conclusion, this comprehensive analysis confirms the effectiveness of the combined use of \EIFplus and \approach, achieving optimal results in both anomaly detection and interpretability.


\subsection{Correlation Table} \label{sec:corr-tab}

\rebuttal{In this section we want to verify the soundness of the results produced by the interpretation algorithms analyzed in this study. This is achieved by computing the correlation between the Local Importance Scores produced by an interpretation algorithm and the Anomaly Scores assigned by the explained AD models to different samples. Intuitively if the interpretation model is able to correctly detect anomalous points then, given an input data point $x$, high LFI scores should correspond to an high value on the Anomaly Score (i.e. $x$ is an outlier). Contrarily low importance values will be paired with a reduced Anomaly Score (i.e. $x$ is an inlier). In order to compute the correlation between the LFI score and the Anomaly Scores the former were aggregated trough the sum in order to obtain a single value. 

In Table \ref{tab:corr_tab} most of the correlation values exceed 0.9 proving the effectiveness of the proposed interpretation methods which are able to provide sufficiently high relevance scores to features describing anomalous samples. An exception is represented by the ECOD and IF\_\approach algorithms which perform poorly in correlation terms (i.e. lower than 0.75) on the {\ttfamily Xaxis} dataset. The reason under these poor performances can be found on the limitations of these two AD models which fail to detect outliers on {\ttfamily Xaxis} as already noted in Section \ref{sec:syn_perf}. 
}

\begin{table}[ht!]
\caption{Correlation values between Local Importance Scores and Anomaly Scores for different interpretation algorithms in Scenario I (S1) and Scenario II (S2) across different datasets. The highest values are highlighted in bold.}
\label{tab:corr_tab}
\resizebox{\textwidth}{!}{
\begin{tabular}{|l|l|m{0.9cm}|m{0.9cm}|m{0.9cm}|m{0.9cm}|m{0.9cm}|m{0.9cm}|m{0.9cm}|m{0.9cm}|m{0.9cm}|m{0.9cm}|m{0.9cm}|m{0.9cm}|m{0.9cm}|m{0.9cm}|m{0.9cm}|m{0.9cm}|}
\hline
\multicolumn{2}{|l|}{} & \multicolumn{5}{c|}{Synthetic} & \multicolumn{11}{c|}{Real World} \\
\hline
\rotatebox[origin=c]{90}{Scenario} & \rotatebox[origin=c]{90}{Model} &
 \rotatebox[origin=c]{90}{{\ttfamily Xaxis}} & \rotatebox[origin=c]{90}{{\ttfamily Bisec}} & \rotatebox[origin=c]{90}{{\ttfamily Bisec3D}} & \rotatebox[origin=c]{90}{{\ttfamily Bisec6D}} & 
 \rotatebox[origin=c]{90}{{\ttfamily Bisec3D\_Skewed}} & \rotatebox[origin=c]{90}{{\ttfamily  Annthyroid}} & \rotatebox[origin=c]{90}{{\ttfamily Breastw}} & \rotatebox[origin=c]{90}{{\ttfamily Cardio}} & \rotatebox[origin=c]{90}{{\ttfamily Diabetes}} & \rotatebox[origin=c]{90}{{\ttfamily Glass}} & \rotatebox[origin=c]{90}{{\ttfamily Ionosphere}} & \rotatebox[origin=c]{90}{{\ttfamily Moodify}} & \rotatebox[origin=c]{90}{{\ttfamily Pendigits}} & \rotatebox[origin=c]{90}{{\ttfamily Pima}} & \rotatebox[origin=c]{90}{{\ttfamily Shuttle}} & \rotatebox[origin=c]{90}{{\ttfamily Wine}} \\
\hline
\multirow{5}{*}{S1} & DIFFI & 0.83& 0.91& 0.92& 0.92& 0.91& 0.89& 0.84& 0.91& 0.88& 0.85& 0.9& 0.88& 0.92& 0.92& 0.91& 0.93\\
& IF\_ExIFFI & \cellcolor{lightgray} 0.76& \cellcolor{lightgray} 0.9& \cellcolor{lightgray} 0.91& \cellcolor{lightgray} 0.92& \cellcolor{lightgray} 0.87& \cellcolor{lightgray} 0.91& \cellcolor{lightgray} 0.91& \cellcolor{lightgray} 0.92& \cellcolor{lightgray} 0.85& \cellcolor{lightgray} 0.92& \cellcolor{lightgray} 0.97& \cellcolor{lightgray} 0.87& \cellcolor{lightgray} 0.91& \cellcolor{lightgray} 0.91& \cellcolor{lightgray} 0.94& \cellcolor{lightgray} 0.9\\
& EIF\_ExIFFI & \cellcolor{lightgray} 0.95& \cellcolor{lightgray} 0.95& \cellcolor{lightgray} 0.95& \cellcolor{lightgray} 0.95& \cellcolor{lightgray} 0.94& \cellcolor{lightgray} 0.91& \cellcolor{lightgray} 0.95& \cellcolor{lightgray} 0.93& \cellcolor{lightgray} 0.91& \cellcolor{lightgray} 0.92& \cellcolor{lightgray} 0.96& \cellcolor{lightgray} 0.75& \cellcolor{lightgray} 0.95& \cellcolor{lightgray} 0.89& \cellcolor{lightgray} 0.88& \cellcolor{lightgray} 0.93\\
& \EIFplus\_ExIFFI & \cellcolor{lightgray} 0.93& \cellcolor{lightgray} 0.93& \cellcolor{lightgray} 0.94& \cellcolor{lightgray} 0.93& \cellcolor{lightgray} 0.92& \cellcolor{lightgray} 0.92& \cellcolor{lightgray} 0.92& \cellcolor{lightgray} 0.94& \cellcolor{lightgray} 0.95& \cellcolor{lightgray} 0.94& \cellcolor{lightgray} 0.96& \cellcolor{lightgray} 0.93& \cellcolor{lightgray} 0.92& \cellcolor{lightgray} 0.94& \cellcolor{lightgray} 0.79& \cellcolor{lightgray} 0.96\\
& ECOD & 0.55& 0.85& 0.85& 0.94& 0.78& 0.88& 0.98& 0.93& 0.86& 0.91& 0.98& 0.87& 0.93& 0.86& 0.88& 0.91\\
\hline
\multirow{5}{*}{S2} & DIFFI & 0.84& 0.95& 0.95& 0.96& 0.93& 0.9& 0.92& 0.93& 0.91& 0.9& 0.91& 0.92& 0.92& 0.92& 0.91& 0.94\\
& IF\_ExIFFI & \cellcolor{lightgray} 0.74& \cellcolor{lightgray} 0.93& \cellcolor{lightgray} 0.92& \cellcolor{lightgray} 0.93& \cellcolor{lightgray} 0.89& \cellcolor{lightgray} 0.9& \cellcolor{lightgray} 0.94& \cellcolor{lightgray} 0.92& \cellcolor{lightgray} 0.88& \cellcolor{lightgray} 0.94& \cellcolor{lightgray} 0.88& \cellcolor{lightgray} 0.91& \cellcolor{lightgray} 0.9& \cellcolor{lightgray} 0.91& \cellcolor{lightgray} 0.92& \cellcolor{lightgray} 0.93\\
& EIF\_ExIFFI & \cellcolor{lightgray} 0.95& \cellcolor{lightgray} 0.97& \cellcolor{lightgray} 0.96& \cellcolor{lightgray} 0.96& \cellcolor{lightgray} 0.95& \cellcolor{lightgray} 0.92& \cellcolor{lightgray} 0.97& \cellcolor{lightgray} 0.95& \cellcolor{lightgray} 0.93& \cellcolor{lightgray} 0.94& \cellcolor{lightgray} 0.94& \cellcolor{lightgray} 0.76& \cellcolor{lightgray} 0.95& \cellcolor{lightgray} 0.88& \cellcolor{lightgray} 0.91& \cellcolor{lightgray} 0.94\\
& \EIFplus\_ExIFFI & \cellcolor{lightgray} 0.92& \cellcolor{lightgray} 0.95& \cellcolor{lightgray} 0.95& \cellcolor{lightgray} 0.94& \cellcolor{lightgray} 0.93& \cellcolor{lightgray} 0.9& \cellcolor{lightgray} 0.97& \cellcolor{lightgray} 0.93& \cellcolor{lightgray} 0.89& \cellcolor{lightgray} 0.94& \cellcolor{lightgray} 0.92& \cellcolor{lightgray} 0.92& \cellcolor{lightgray} 0.92& \cellcolor{lightgray} 0.94& \cellcolor{lightgray} 0.89& \cellcolor{lightgray} 0.95\\
& ECOD & 0.58& 0.84& 0.84& 0.94& 0.77& 0.88& 0.9& 0.93& 0.88& 0.9& 0.99& 0.86& 0.93& 0.85& 0.88& 0.89\\
\hline
\end{tabular}}
\end{table}

\subsection{Time Scaling Experiments}\label{sec:time_scaling}

\begin{figure}[!ht] 
\centering
\hspace{0.13\textwidth}\textbf{Fit} \hspace{0.24\textwidth} \textbf{Predict} \hspace{0.16\textwidth} \textbf{Importances}
\begin{minipage}{\textwidth}
    \rotatebox{90}{\parbox{15mm}{\centering \textbf{Samples}}} 
    \begin{minipage}{0.95\textwidth}
        \begin{subfigure}[t]{0.30\textwidth}
            \centering
            \includegraphics[trim={0 0 0 0}, clip, width=\linewidth]{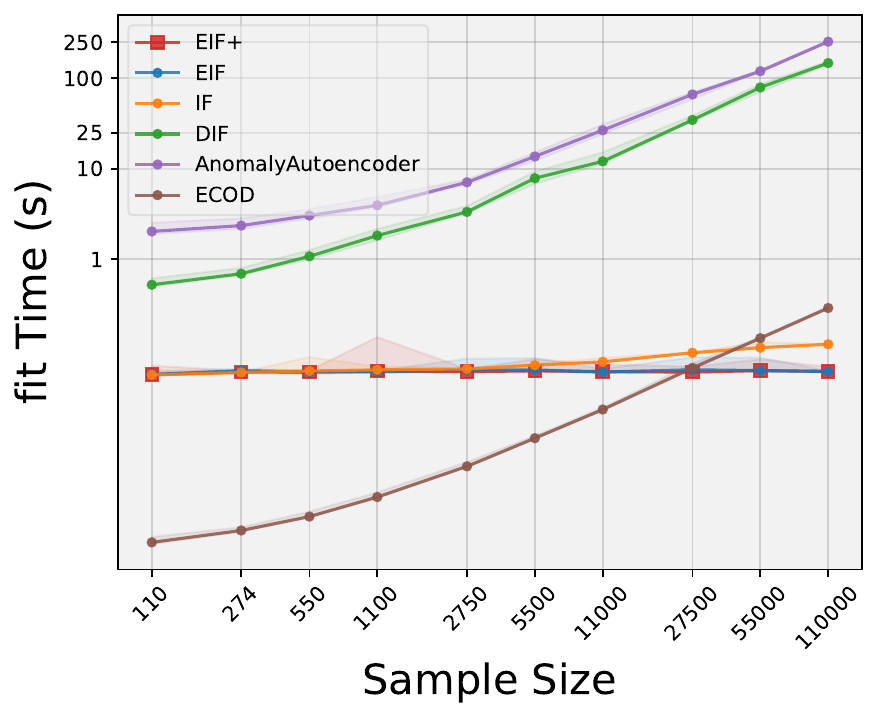}
            \caption{Fit time vs. sample size}
            \label{fig:Fit-time-samples}
        \end{subfigure}\hfill
        \begin{subfigure}[t]{0.30\textwidth}
            \centering
            \includegraphics[trim={0 0 0 0}, clip, width=\linewidth]{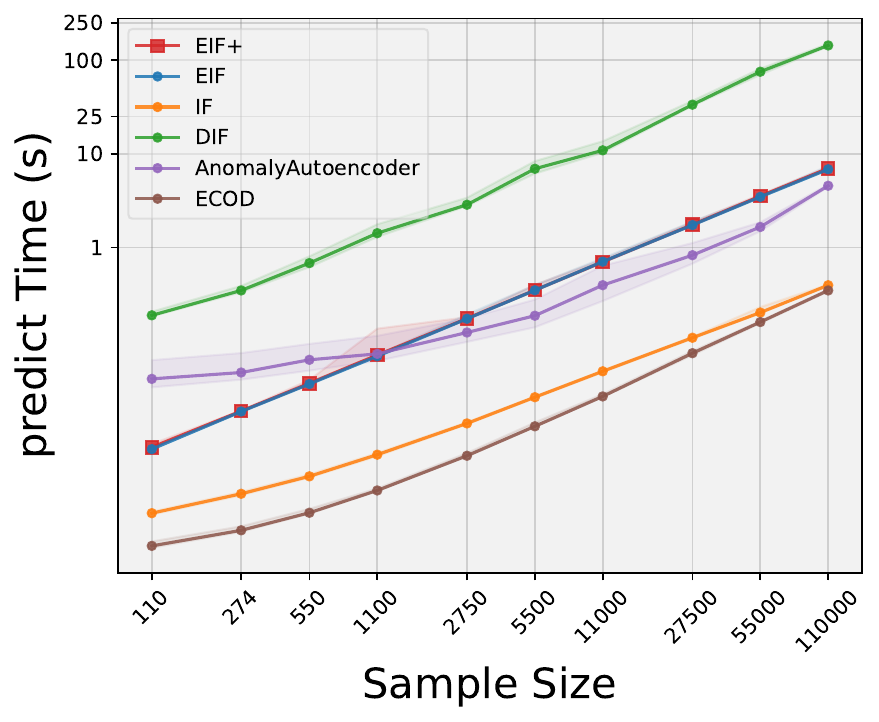}
            \caption{Predict time vs. sample size}
            \label{fig:Predict-time-samples}
        \end{subfigure}\hfill
        \begin{subfigure}[t]{0.30\textwidth}
            \centering
            \includegraphics[trim={0 0 0 0}, clip, width=\linewidth]{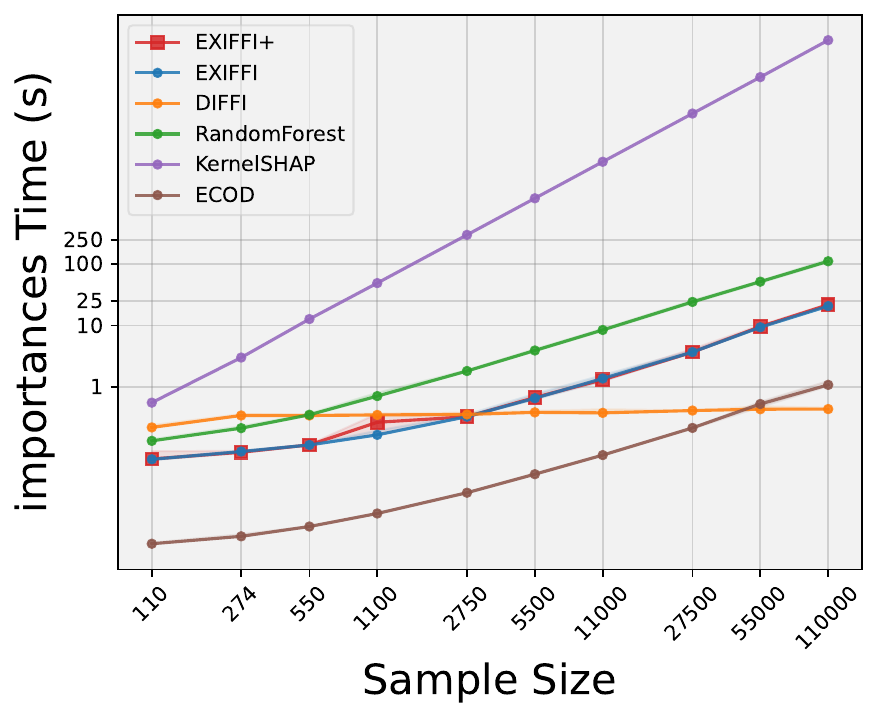}
            \caption{Importance time vs. sample size}
            \label{fig:Importance-time-samples}
        \end{subfigure}
    \end{minipage}
\end{minipage}

\vspace{5mm} 

\begin{minipage}{\textwidth}
    \rotatebox{90}{\parbox{15mm}{\centering \textbf{Features}}} 
    \begin{minipage}{0.95\textwidth}
        \begin{subfigure}[t]{0.30\textwidth}
            \centering
            \includegraphics[trim={0 0 0 0}, clip, width=\linewidth]{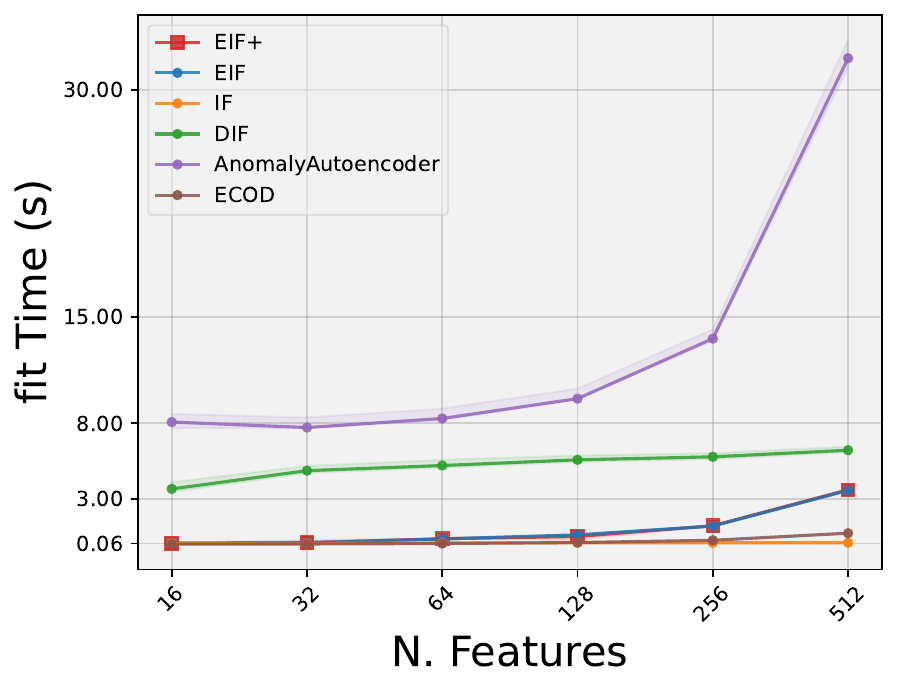}
            \caption{Fit time vs. N. of features}
            \label{fig:Fit-time-features}
        \end{subfigure}\hfill
        \begin{subfigure}[t]{0.30\textwidth}
            \centering
            \includegraphics[trim={0 0 0 0}, clip, width=\linewidth]{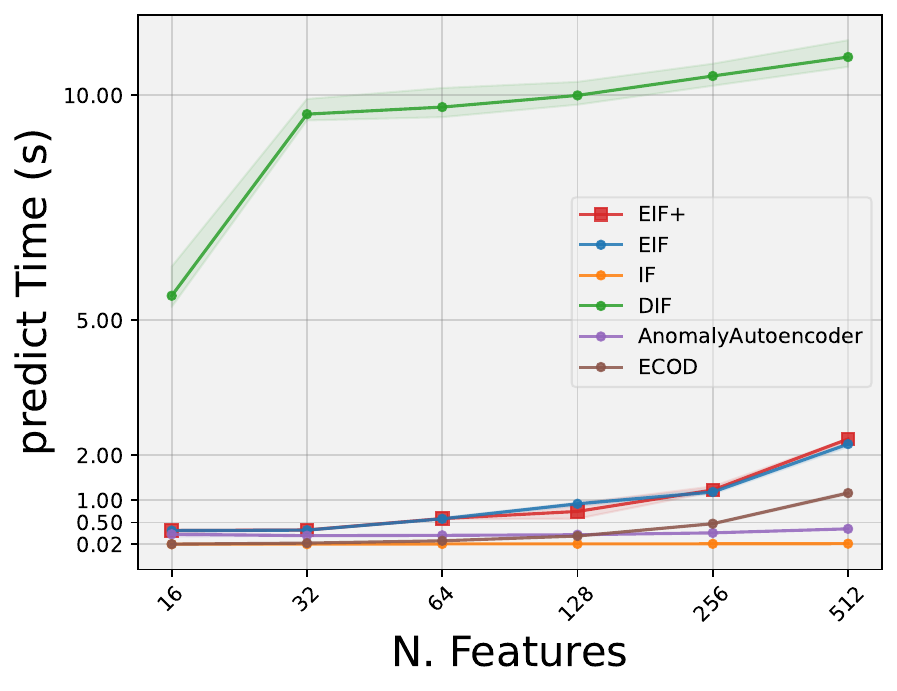}
            \caption{Predict time vs. N. of features}
            \label{fig:Predict-time-features}
        \end{subfigure}\hfill
        \begin{subfigure}[t]{0.30\textwidth}
            \centering
            \includegraphics[trim={0 0 0 0}, clip, width=\linewidth]{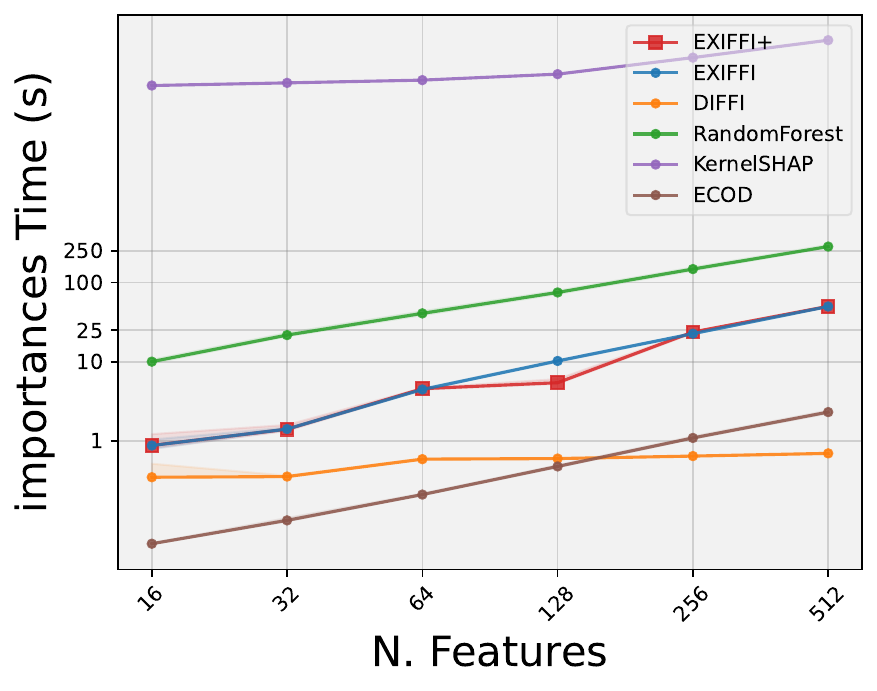}
            \caption{Importance time vs. N. of features}
            \label{fig:Importance-time-features}
        \end{subfigure}
    \end{minipage}
\end{minipage}

\caption{Time Scaling experiments display the execution times trend of various AD and XAD algortihms varying the sample size (fixing the number of features to 6) and the cardinality of the feature space (fixing the sample size to 5000).}\label{fig:time-plots}
\end{figure}

\ale{In this section, we compare the computational efficiency of different algorithms, evaluating the time required for model fitting, making predictions, and interpreting results. Our examination, depicted in Figures \ref{fig:time-plots}, presents how these durations adjust with dataset size increases—both by sample number and feature count, on a logarithmic scale. This investigation is crucial for understanding each algorithm's practical use in varying scenarios, such as industrial-sized datasets, where the sample size and the feature space are generally larger than the ones encountered in the benchmark datasets used for evaluation in \ref{sec:experimental_results}. Figures \ref{fig:Fit-time-samples}, \ref{fig:Fit-time-features} details fitting times varying the sample size and number of features respectively, Figures \ref{fig:Predict-time-samples}, \ref{fig:Predict-time-features} outline prediction times, and Figures \ref{fig:Importance-time-samples}, \ref{fig:Importance-time-features} the time to determine importance scores, providing insight into the scalability and efficiency of these computational processes.}

\rebuttal{
Analyzing computational efficiency, our observations reveal significant differences in the fitting, prediction, and interpretation speeds of various algorithms, as depicted in Figures \ref{fig:Fit-time-samples}, \ref{fig:Fit-time-features}, \ref{fig:Predict-time-samples}, \ref{fig:Predict-time-features}, and \ref{fig:Importance-time-samples}, \ref{fig:Importance-time-features}. Specifically, the fitting time, as shown in Figure \ref{fig:Fit-time-samples}, demonstrates exponential increases for both the DIF and Anomaly Autoencoder models. The Anomaly Autoencoder exhibits a steeper rise due to weight optimization requirements, contrasting with DIF, which utilizes deep neural networks with random weights for nonlinear input data transformation, hence avoiding lengthy NN training periods.

Additionally, ECOD scales exponentially as the sample size of the dataset increases. Although ECOD is faster than IF-based models when the sample size is small, it becomes slower when the dataset contains more than 30k points. Interestingly, the number of features does not significantly affect the efficiency of ECOD, making it less sensitive to feature-dimensionality compared to other models.

For the experiment in which time effectiveness is tested by varying the number of features in the dataset, the Anomaly Autoencoder remains the most time-consuming method, while the trend of DIF is closer to that of the Isolation-based methods (i.e., EIF, \EIFplus, IF), as shown in \ref{fig:Fit-time-features}.}

\ale{The prediction time analysis, referenced in Figure \ref{fig:Predict-time-samples}, highlights DIF as the most time-intensive model. This is attributed to DIF's necessity to aggregate predictions across an ensemble of neural networks and representation spaces, significantly extending the prediction process. In \ref{fig:Predict-time-features} , instead, while DIF is still the slowest model , the Anomaly Autoencoder, ECOD and IF models prediction time can be considered independent on the number of features but only dependent on the sample size. Finally, EIF and \EIFplus depict an increasing trend with, however, much smaller values than the ones of DIF.}

\rebuttal{In the case of interpretation, Figure \ref{fig:Importance-time-samples} identifies DIFFI as the fastest algorithm for calculating importance scores. EIF and \EIFplus experience a marked increase in computation time for this task beyond 11,000 samples. Nonetheless, their performance remains markedly better compared to the RandomForest post-hoc interpretability method, which suffers from exponential growth in execution time with larger sample and feature sizes. KernelSHAP, on the other hand, produces out-of-scale time results, making it unsuitable as an interpretation algorithm for large datasets due to its inefficiency.

ECOD, in contrast, is in line with the time complexity of \approach, with a constant factor of difference since ECOD does not need to reconstruct the importance scores because they are intrinsically embedded in its output. This analysis illustrates the efficiency and adaptability of \EIFplus and ECOD in providing rapid interpretation across varying dataset scales.}


\section{Conclusions}\label{sec:conclusions}

\rebuttal{This paper delves into the important area of unsupervised AD, a critical task for identifying unusual patterns or behaviors in data. While detecting anomalies is fundamental, the paper emphasizes that this is often not enough for practical use cases. Users not only need to know that an anomaly has occurred, but also understand why the model made a certain prediction. This understanding is key for root cause analysis and for building trust in the model’s decisions.

The primary contribution of this work is the introduction of \approach, an interpretability method inspired by DIFFI that aims to provide both local and global interpretability, particularly for Isolation Forest (IF)-based models. The qualitative results demonstrate that the approach produces comprehensible results, as shown through its application on three datasets: {\ttfamily Xaxis}, {\ttfamily Bisect3D\_skewed}, and {\ttfamily glass}. Additionally, the work presents \EIFplus, a variant of the Extended Isolation Forest (EIF) model, which is designed to improve generalization performance. A thorough comparative analysis of Isolation-based and deep learning-based anomaly detection approaches is provided, which represents one of the most comprehensive studies in this domain.

The experimental results support the utility of \EIFplus and \approach, showing that \EIFplus yields robust performance across multiple datasets, outperforming benchmark models in certain scenarios. Specifically, \EIFplus was the top-performing model in 6 out of 11 real-world datasets with the highest average score across the datasets in Scenario II and the second-best model after EIF in Scenario I. The \approach method applied to IF-based models consistently showed superior performance across all scenarios, as evaluated using the $AUC_{FS}$ metric, which is based on the Unsupervised Feature Selection proxy task. 

Furthermore, the applicability of \EIFplus and \approach to large-scale datasets, such as those encountered in industrial settings, was assessed through time-scaling experiments, with results presented in the paper's corresponding section.

Looking ahead, future research could explore innovative ways to utilize the information encoded in the splitting nodes of decision trees, particularly when multiple variables compete for relevance in the model. An additional research direction could investigate the development of a more general interpretability algorithm that could be applied to all types of Isolation Forest models, including those with different splitting functions, such as Hypersphere IF {\cite{hypersphere_if}}.

Finally, despite the promising results of \approach, a key limitation of the current study is the lack of evaluation in a real-world industrial scenario with field-specific expertise. While \approach provides interpretability and facilitates understanding of anomaly detection models, its true impact on human comprehension and decision-making processes has not yet been fully explored. The study did not include experiments involving human feedback, where experts in the field could interact with the model and assess how well the interpretability methods aid in understanding the data. Without such evaluations, the extent to which \approach improves human understanding of model predictions and supports more informed decisions remains uncertain. Future work should focus on assessing the practical value of \approach by conducting experiments that involve domain experts and gathering insights on how interpretability impacts human interaction with data-driven systems.}


\bibliographystyle{elsarticle-num} 
\bibliography{main}

\newpage
\appendix
\section{Appendices}

\subsection{Datasets}\label{sec:appendix_datasets}

\subsubsection{Synthetic datasets}\label{sec:appendix-synthetic-data}
\rebuttal{
We generated three additional datasets, following the same procedure for the inliers and outliers definition as the one described in Section \ref{sec:Evaluation}: {\ttfamily Bisect}, {\ttfamily Bisect3D} and {\ttfamily Bisect6D}, with the aim to explore how performance is affected when anomalies are defined by multiple features. These new datasets position the anomalies along bisectors in higher-dimensional subspaces (i.e. 2,3 and 6 dimensions respectively), incorporating multiple features to define the outliers. By doing so, we are able to assess the robustness of the algorithms in detecting anomalies that span multiple dimensions.

The mathematical construction of these synthetic datasets ensures a controlled and systematic variation in the number of anomalous features, providing valuable insights into the behavior of the algorithms in complex, high-dimensional settings.

\begin{itemize}
    \item In the {\ttfamily Bisect} dataset, the inliers are uniformly distributed within a six-dimensional sphere. The outliers are positioned along the bisector of the subspace formed by Feature 0 and Feature 1, using the direction vector $v = [1, 1, 0, 0, 0, 0]$. Noise is added to the remaining four dimensions to introduce variability, and the outliers are scaled and perturbed accordingly. The graphical results of the \approach explanations for this dataset are accessible at \url{https://github.com/alessioarcudi/ExIFFI} and they showcase how the most important features, with similar scores, are Feature 0 and 1 as expected from the generation process.

    \item In the {\ttfamily Bisect3D} dataset, the inliers follow the same distribution as in {\ttfamily Bisect}. The outliers are aligned along the bisector of the subspace spanned by Feature 0, Feature 1, and Feature 2, with the direction vector $v = [1, 1, 1, 0, 0, 0]$. Noise is added to the remaining three dimensions to simulate variability. This dataset is very similar to {\ttfamily Bisect3D\_Skewed}, presented in \ref{sec:Evaluation}, since also in this case anomalies are aligned along three features. The main difference lays in the vector $v$ used for the generation. In {\ttfamily Bisect3D} $v$ has the same amplitude across all the three anomalous directions, making outliers deviate from normal points by the same amount along Feature 0,1 and 2. Contrarily in {\ttfamily Bisect3D\_Skewed} the outlyingness of the different variables is different. As a result in the Score Plot of {\ttfamily Bisect3D}, depicted in Figure \ref{fig:Feat-imp-Bisect-3d}, the importance scores of the top 3 features are almost identical while there is a clear distinction in the Score Plot of {\ttfamily Bisect3D\_Skewed}, outlined in Figure \ref{fig:Bisect3D_prop_score_bars}.

    \item Finally in {\ttfamily Bisect6D} anomalies are distributed along all the six features composing the input space while inliers are distributed as usual with the addition of some white noise to induce diversity. Consequently the direction vector is defined as $v = [1, 1, 1, 1, 1, 1]$. This dataset was produced to observe the behavior of \approach in an extreme scenario where all dimensions are affected by anomalous points. The outcome is that all six features are assigned more or less the same importance score and thus the final feature ranking is decided by the stochasticity of the algorithm. For a more detailed description of the results we refer the interested reader to \url{https://github.com/alessioarcudi/ExIFFI}. 
\end{itemize}
}

\rebuttal{For clarity the pseudocode for the generation of the synthetics datasets employed in this paper is provided in \ref{alg:syn_data}}. 

\begin{algorithm}[ht!]
\caption{Synthetic Dataset Generation with Direction Vector $u$ for Outliers} \label{alg:syn_data}
\SetKwInOut{Input}{Input}
\SetKwInOut{Output}{Output}

\Input{Number of samples to generate $n$, Number of inliers $n_I$, Number of outliers $n_O$, Number of features $p$, Radius of the inliers ball $r$,Radius of the outlier distance $d$, Number of anomalous features $k$, Bounds for anomalies $[\text{min}, \text{max}]$, Direction vector $v \in \mathbb{R}^k$}
\Output{Generated synthetic dataset $\mathcal{D} \in \mathbb{R}^{n \times p}$}

$\mathcal{D}_I \gets \text{empty list}$\;
$\mathcal{D}_O \gets \text{empty list}$\;
\tcp{Initialize empty lists for inliers and outliers}

\tcp{--- Inliers Generation ---}
\While{$\text{len}(\mathcal{D}_I) < n_I$}{
    \tcp{Sample a random $p$-dimensional vector from a Uniform distribution in $[-r,r]$}
    $\text{inlier\_point} \sim \mathcal{U}_p([-r,r])$\;
    
    \tcp{Check if the L2 norm of the point is less than or equal to $r$}
    \If{$(||\text{inlier\_point}||^2) \leq r$}{
        $\mathcal{D}_I \gets \mathcal{D}_I \cup \text{inlier\_point}$\;
    }
}

\tcp{--- Normalize the direction vector $v$ ---}
$u \gets v / ||v||$\;
\tcp{Normalize $v$ to obtain the unit direction vector $u$}

\tcp{--- Outliers Generation ---}
\For{$i \gets 1$ \textbf{to} $n_O$}{
    \tcp{Sample a scalar $x$ from a uniform distribution in $[\text{min}, \text{max}]$}
    $x \sim \mathcal{U}([\text{min}, \text{max}])$\;

    \tcp{Create an empty vector for the outlier point}
    $\text{outlier\_point} \gets \text{zeros}(p)$\;

    \tcp{Calculate the values of the first $k$ anomalous features based on direction $u$ and added noise}
    \For{$j \gets 1$ \textbf{to} $k$}{
        $\text{outlier\_point}[j] \gets d*u[j] + x*v[j] + \mathcal{N}(0,1)$\;
    }

    \tcp{Generate remaining $p-k$ noisy features from a Gaussian distribution}
    \For{$j \gets k+1$ \textbf{to} $p$}{
        $\text{outlier\_point}[j] \gets \mathcal{N}(0,1)$\;
    }
    
    \tcp{Add the generated outlier point to the dataset}
    $\mathcal{D}_O \gets \mathcal{D}_O \cup \text{outlier\_point}$\;
}

\tcp{Combine inliers and outliers to form the final dataset}
$\mathcal{D} \gets \mathcal{D}_I \cup \mathcal{D}_O$\;

\Return{$\mathcal{D}$}\;
\end{algorithm}

\subsubsection{Real-world Datasets}\label{sec:appendix-real-world-data}
\rebuttal{
Most of the Real-World datasets used in our experiments come from the widely used Outlier Detection DataSets (ODDS) library introduced by Rayana \cite{Rayana:2016}. Unlike synthetic datasets, these reflect the complexity of real-world scenarios. However, many of these datasets were originally designed for different purposes, such as multi-class classification, and later adapted for anomaly detection (AD) by undersampling the least-represented class to create outliers, while merging the remaining classes as normal data. This transformation can present challenges, as the undersampled minority class may not truly represent anomalies, complicating the evaluation of AD methods. Moreover, AD models designed to detect isolated anomalies may not always align with the provided labels, further complicating interpretability, as discussed in Section \ref{sec:real_perf}.

Many real-world datasets also lack detailed information on features and labels, making it difficult to evaluate model performance and interpret Feature Importance scores. An exception is the {\ttfamily Glass} dataset, detailed by Carletti et al. \cite{carletti2019explainable}, which allows for deeper analysis of feature influence on anomalies (see Section \ref{sec:real_perf}). The absence of such detailed information poses significant challenges to assessing models qualitatively and understanding feature impact on anomalies.

To address these limitations, two new datasets, {\ttfamily Diabetes} and {\ttfamily Moodify}, were added. These datasets provide detailed information on the semantics of their features and labels, allowing for more thorough evaluation of AD and interpretation results in alignment with domain knowledge.}

\rebuttal{Here we introduce the real-world datasets with details about the context of the features and labels}
\begin{itemize}
    \item {\ttfamily Annthyroid}: The {\ttfamily Annthyroid} dataset \cite{misc_thyroid_disease_102} is part of the UCI Machine Learning Repository. It is a three-class Classification dataset. The aim of the classification task associated with this dataset is to detect whether a patient is hypothyroid (i.e. the patient has an underactive thyroid) or not. For this scope, the three classes used refer to the functioning conditions of the thyroid that can be\textit{normal} (not hypothyroid), \textit{hyperfunctioning} (overactive thyroid) and \textit{subnormal}. The dataset originally contained 15 categorical and 6 numerical attributes. In order to adapt it to an Anomaly Detection task only the numerical attributes were considered and the \textit{hyperfunctioning} and \textit{subnormal} function classes were considered as part of the outliers while the \textit{normal} functioning samples are used to build the inlier class. 
    The six numerical features represent, respectively, the following quantities: TSH, T3, TT4, T4U, and FTI. 
        \begin{itemize}
            \item \textbf{TSH (Thyroid-Stimulating Hormone}: TSH is an hormone produced by the pituitary gland. An underactive thyroid (hypothyroidism) is associated with high levels of TSH while low levels of TSH are associated with an overactive thyroid (hyperthyroidism). 
            \item \textbf{T3 (Triiodothyronine)}: T3 is one of the thyroid hormones. It plays a crucial role in regulating the metabolism. It is important to measure the levels of this hormone, specifically in cases of hyperthyroidism. 
            \item \textbf{TT4 (Total Thyroxine)}: This quantity represents the total amount of the T4 thyroid hormone in the blood. T4 levels are associated to the overall thyroid hormone production. 
            \item \textbf{T4U (Thyroxine-Binding Globulin}: T4U measures the level of thyroxine-binding globulin, a protein binding to thyroid hormones in the blood. T4U levels are connected to the thyroid hormone availability in the body. 
            \item \textbf{FTI (Free Thyroxine Index}: Taking into account the T4 and T4U levels, the FTI provides an estimate of the amount of free thyroxine (T4) in the blood. It is used to asses the free, active thyroid hormone levels in the body. 
        \end{itemize}
        
    \item {\ttfamily Breast}: The {\ttfamily Breast} dataset \cite{misc_breast_cancer_wisconsin_(original)_15} is a Binary Classification dataset where the target is the presence of breast cancer or not in a patient. The peculiarity of the samples contained in this dataset is that they are formed by categorical features. Normally, the Anomaly Detection models described in this paper are not built to deal with categorical features but in this particular case the high number of levels (e.g. The age variable has nine levels: 10-19, 20-29, \dots; the tumor-size feature has levels equal to 0-4,5-9,10-14, \dots) characterizing the {\ttfamily Breast} dataset's attributes makes it possible to consider them as numerical features.
    The dataset is composed of samples coming from the clinical cases of Dr. Wolberg collected in a time span going from January 1989 to November 1991. 
    The dataset is composed by 9 features representing the following quantities: Clump Thickness, Uniformity of Cell Size, Uniformity of Cell Shape, Marginal Adhesion, Single Epithelial Cell Size, Bare Nuclei, Bland Chromatin, Normal Nucleoli and Mitoses. 
    
    \item {\ttfamily Cardio}: The {\ttfamily Cardio} \cite{misc_cardiotocography_193} dataset is part of the UCI Machine Learning Repository and its complete name is {\ttfamily Cardiotocoraphy} since it contains measurements of fetal heart rate (FHR) and uterine contraction (UC) on cardiotocograms \footnote{A cardiotogram is a medical test monitoring fetal hearth rate and uterine contractions during pregnancy. It is used to assess the health status of the fetus and the progress of labor during pregnancy and childbirth}. This dataset contains 3 classes that were assigned by expert obstetricians: \textit{normal}, \textit{supsect}, and \textit{pathologic}. In order to use this dataset for Anomaly Detection the \textit{pathologic} class was discarded and the \textit{pathologic} class was downsampled to 176 points to maintain unbalance with respect to the \textit{pathologic} class. There are 21 features. \footnote{The complete list of features names can be found at: https://archive.ics.uci.edu/dataset/193/cardiotocography}

    \item {\ttfamily Glass}: \ale{The {\ttfamily Glass} dataset \cite{misc_glass_identification_42} is originally used for multi-class classification to distinguish 7 different types of glasses. There are 9 features: the first one is the Refractive Index while the others measure the concentration of Magnesium (Mg), Silicon (Si), Calcium (Ca), Iron (Fe), Sodium (Na), Aluminum (Al), Potassium (K) and Barium (Ba) in the glass samples. The seven original glass groups were divided into two: Groups 1,2,3 and 4 represent Window Glasses while the remaining ones are non-window glasses: containers glass, tableware glass and headlamp glass. Among the non-window glasses, headlamp glasses were considered as outliers while Window Glasses are labeled as inliers in order to convert this dataset to be used for Anomaly Detection. According to some prior knowledge on the subject the Barium (Ba) and Potassium (K) concentration should be decisive in distinguishing between headlamp and window glasses. In fact Barium is considered the crucial element to perform the distinction between headlamp and window glasses since it is usually added to the category of headlamp glasses in order to improve their optical properties. Another potential key attribute is Potassium (K) that is frequently exploited to enhance strength and thermal resistance of window glasses.}

    \item {\ttfamily Ionosphere}: The {\ttfamily Ionosphere} dataset \cite{misc_ionosphere_52} contains measurements collected by a radar in Goose Bay, Labrador. The measurement system consisted of 16 high-frequency antennas with a transmitted power in the order of 6.4 kilowatts. The targets to measure were free electrons in the Ionosphere. Originally this is built as a Binary Classification dataset where the two classes are \textit{good} if the captured electrons show some kind of structure in the ionosphere and \textit{bad} otherwise. The signal is processed with an autocorrelation function depending on the time of a pulse and the pulse number. There are 17 pulse numbers and each of them is identified by a pair of features for a total of 34 features. 

    
    \item {\ttfamily Pima}: The {\ttfamily Pima} dataset \footnote{https://www.kaggle.com/datasets/uciml/pima-indians-diabetes-database} comes from the National Institute of Diabetes, Digestive and Kidney Diseases. It is a Binary Classification dataset whose aim is to predict whether a patient has diabetes or not. This dataset is the result of the application of some constraints to a larger dataset. In {\ttfamily Pima}, in fact, the data considered are obtained from female patients of at least 21 years of age coming from the Pima Indian heritage. 
    
    
    The dataset contains 8 features indicating some typical diagnostic measurements: Number of pregnancies, Plasma Glucose Concentration, Diastolic Blood Pressure, Triceps skin thickness, 2-hour serum if insulin, BMI (Body Mass Index), Diabetes Pedigree function, and Age.

    \item {\ttfamily Pendigits}: The {\ttfamily Pendigits} dataset collects data regarding handwritten digits produced by Forty-four human writers. The task associated with this dataset is the one of recognizing the correct written digit. The dataset is composed of 16 features and the number of objects per class (i.e. per digit) was reduced by a factor of 10 \% to increase the unbalance between classes in order to adapt the dataset as a benchmark for the evaluation of Anomaly Detection models.
    
    \item {\ttfamily Shuttle}: The {\ttfamily Shuttle} dataset \cite{misc_statlog_(shuttle)_148} describes radiator positions in a NASA space shuttle. The samples are characterized by 9 attributes. Originally, the dataset is used for Multi-Class Classification with the target variable containing seven possible classes, which are: Radiator Flow, Fpv Close, Fpv Open, High, Bypass, Bpv Close, and Bpv Open. Besides the normal Radiator Flow class about 20\% of the data points describe anomalous situations. To reduce the amount of anomalies the Radiator Flow class is used to form the inlier class while a stratified sampling procedure was applied to classes 2,3,5,6 and 7. Finally, data coming from class 4 were discarded.
    
    \item {\ttfamily Wine}: The {\ttfamily Wine} dataset \cite{misc_wine_109} is part of the UCI Machine Learning repository and it was originally created as a 3-class classification dataset. In fact, it contains data resulting from a chemical analysis of wines grown in the same region of Italy but obtained from 3 different cultivars. So the aim of a Classification model applied to this dataset would be to correctly predict the original culture of wine given its chemical properties. As it is usually done in these cases the dataset was adapted to test Anomaly Detection models considering the data from two cultures as inliers and the ones beholding to the last culture as the outliers. The dataset is composed by 13 features representing the following quantities: Alcohol, Malic Acid, Ash, Alcalinity of ash, Magnesium, Total phenols, Flavanoids, Nonflavanoid phenols, Proanthocyanins, Color Intensity, Hue, OD280/OD315 \footnote{These codes refers to two measurements of the optical density of a wine sample performed at two different wavelengths: 280 nm and 315 nm. These measurements are typically exploited to assess the protein content and the stability of wine} of diluted wines, and Proline. 
    
\item {\ttfamily Diabetes}: The {\ttfamily Diabetes} dataset \footnote{https://www.kaggle.com/datasets/iammustafatz/diabetes-prediction-dataset} is a Binary Classification dataset with medical data about patients used to predict whether they have diabetes or not. There are 4 categorical and 4 numerical variables. In order to use the \approach model only the numerical variables were considered: 

        \begin{itemize}
            \item Age: The age ranges between 0 and 80 and it can be an important factor since Diabetes is usually diagnosed in older adults.
            \item BMI (Body Mass Index): Measure of body fat based on weight and height. High BMI values are linked to a higher risk of diabetes.
            \item HbA1c-Level: The Hemoglobin A1c Level it's a measure of a person's average blood sugar level over the past 2-3 months. A high level of HbA1c is usually associated with high diabetes risk.
            \item blood-glucose-level: The amount of glucose in the bloodstream at a given time. High glucose levels are a key factor to detect Diabetes.
        \end{itemize}

        Finally, the target variable is a binary variable indicating the presence or absence of diabetes in the patient. Following the usual protocol for Real World Datasets, the inlier group will be represented by healthy patients while the ones affected by diabetes will be placed in the outlier group. 
    
    \item {\ttfamily Moodify}: Moodify is a recommendation app \footnote{More details on the Moodify project can be found at: https://github.com/orzanai/Moodify} that classifies Spotify songs according to the emotions transmitted to the users. The system is trained on the  {\ttfamily Moodify} dataset \footnote{https://www.kaggle.com/datasets/abdullahorzan/moodify-dataset} which contains features regarding the main characteristics of about 278.000 songs and the target is a 4 levels categorical variable with the following coding: 

    \begin{enumerate}
        \item \textit{Sad}
        \item \textit{Happy}
        \item \textit{Energetic}
        \item \textit{Calm}
    \end{enumerate}

    The label with the lowest frequency (15\%) in the dataset is \textit{Calm} so that was used to create the outliers group while the inlier group is formed by the songs that belong to the other three classes. 

    The dataset is composed of 11 numerical features describing different musical attributes of a song. Except for the Loudness variable, which expresses the overall loudness of the track in decibels (dB), all the features have values contained in the $[0,1]$ interval. The variable names are the following: Acousticness, Danceability, Energy, Instrumentalness, Liveness, Loudness, Speechiness, Valence, and Tempo. Other variables are Duration (ms) and spec-rate.
\end{itemize}

\subsection{Draw-backs of Isolation Forest}\label{sec:IFvsEIF}
Next, we delve deeper into the artifacts of the IF and how the EIF is able to avoid them.
The primary distinction among the mentioned models lies in their approach to determining splitting hyperplanes. In the case of the 
IF, it operates under the constraint of splitting the feature space along a single dimension. Consequently, the constructed splitting hyperplane is invariably orthogonal to one dimension while remaining parallel to the others.

\begin{figure}[!ht]
    \centering 
    \begin{subfigure}{0.45\textwidth}
    \centering
      \includegraphics[width=\linewidth]{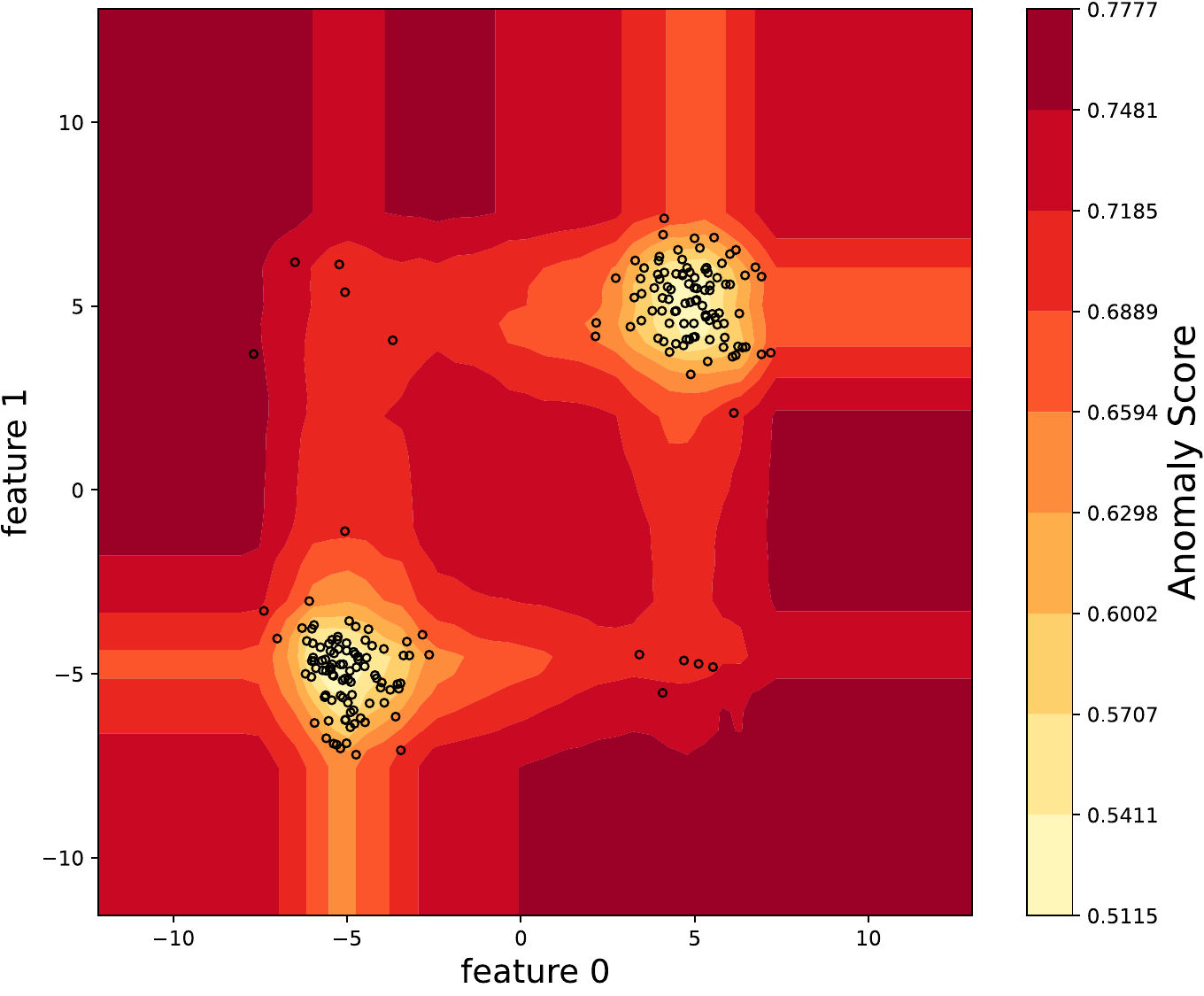}
      \caption{Anomaly Scoremap IF}
      \label{fig:anomaly scoremap IF}
    \end{subfigure}\hfil 
    \begin{subfigure}{0.45\textwidth}
    \centering
      \includegraphics[width=\linewidth]{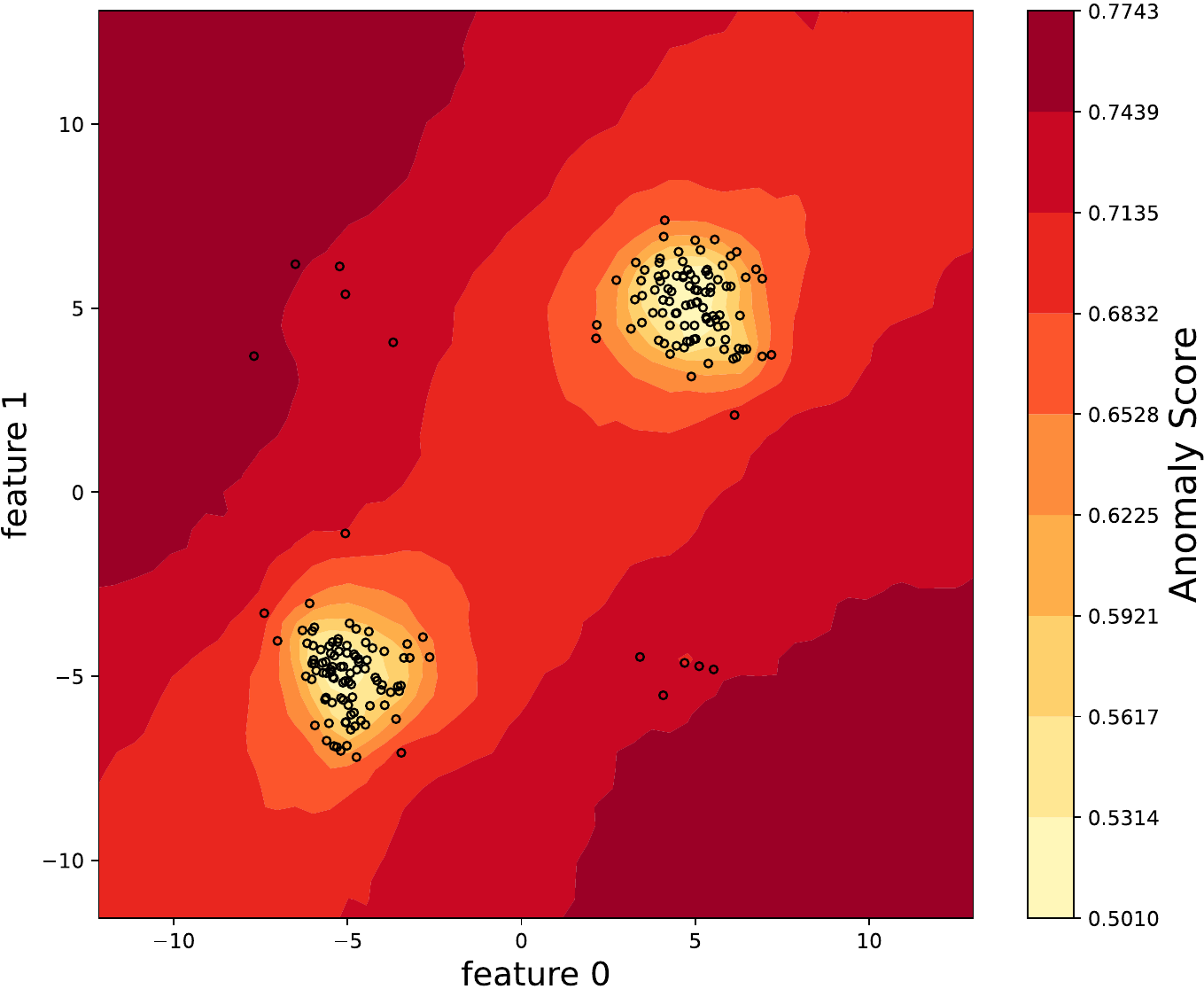}
      \caption{Anomaly Scoremap EIF}
      \label{fig:anomaly scoremap EIF}
    \end{subfigure}\hfil 
\caption{Scoremap showing the differences between IF and EIF in the two dimensional space dataset {\ttfamily Bimodal} described in Section \ref{sec:appendix-synthetic-data}. Figure inspired from \cite{8888179}.}\label{fig:scoremap 2 IF vs EIF}
\end{figure}

The EIF model instead relaxes this constraint, it allows multiple splits in the feature space along different dimensions, thus the splitting hyperplane can be oriented differently in each isolation tree. 
This relaxation helps to provide a more expressive way to capture complex data distributions while maintaining performances.
Moreover, as Hariri et al. pointed out in \cite{8888179}, the EIF avoids the creation of regions where the anomaly score is lower only due to the imposed constraints. In Figure \ref{fig:scoremap 2 IF vs EIF} we can observe that the effect of the constraint is twofold: (i) it generates bundles in the hyperplanes orthogonal to the main directions, and (ii) it creates "artifacts", i.e. low anomaly score zones in their intersections, as we can observe in the Scoremap \ref{fig:anomaly scoremap IF}.

As we progressively relax the IF constraint on splitting directions, these artifacts tend to disappear. We simulated the evolution of the anomaly score surface by allowing the hyperplane to split along a new feature in each step. Figure 
vividly demonstrates the gradual elimination of artifact-prone regions.

\begin{figure}[!ht] 
    \centering 
    \begin{subfigure}{0.33\textwidth}
    \centering
      \includegraphics[trim={0 0 3.5cm 0}, clip, width=\linewidth]{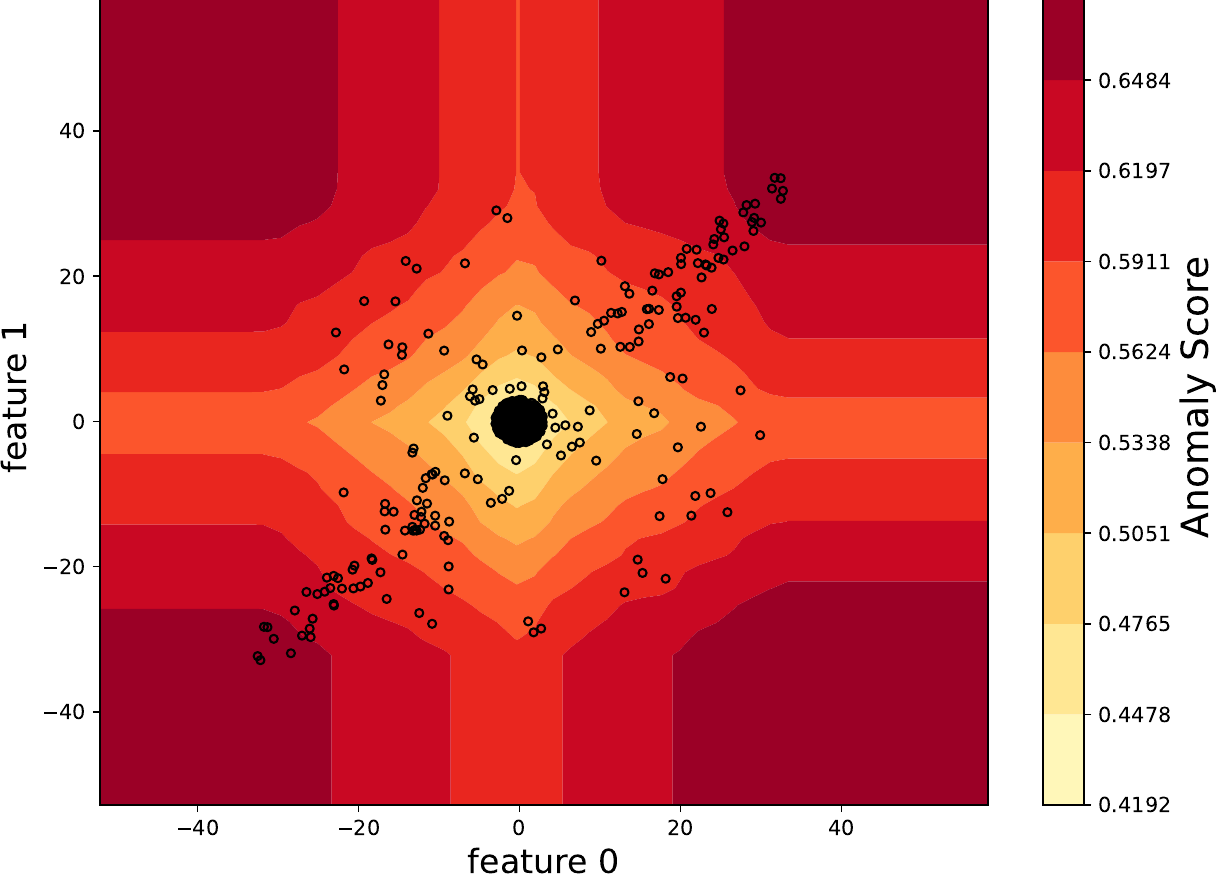}
      \caption{Anomaly Scoremap with 1 degree of freedom}
      \label{fig:scoremap1df}
    \end{subfigure}\hfil 
    \begin{subfigure}{0.33\textwidth}
    \centering
      \includegraphics[trim={0 0 3.5cm 0}, clip, width=\linewidth]{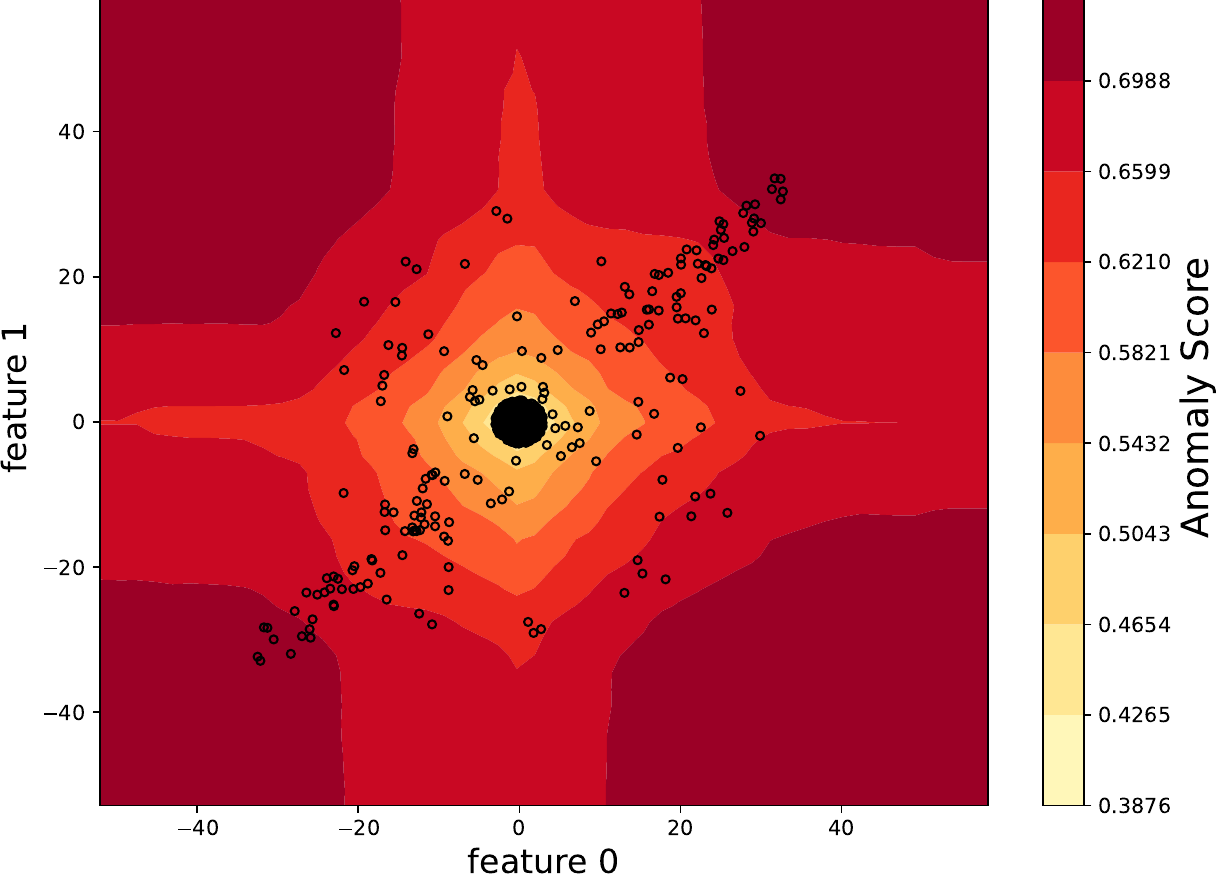}
      \caption{Anomaly Scoremap with 2 degree of freedom}
      \label{fig:scoremap2df}
    \end{subfigure}\hfil 
    \begin{subfigure}{0.33\textwidth}
    \centering
      \includegraphics[trim={0 0 3.5cm 0}, clip, width=\linewidth]{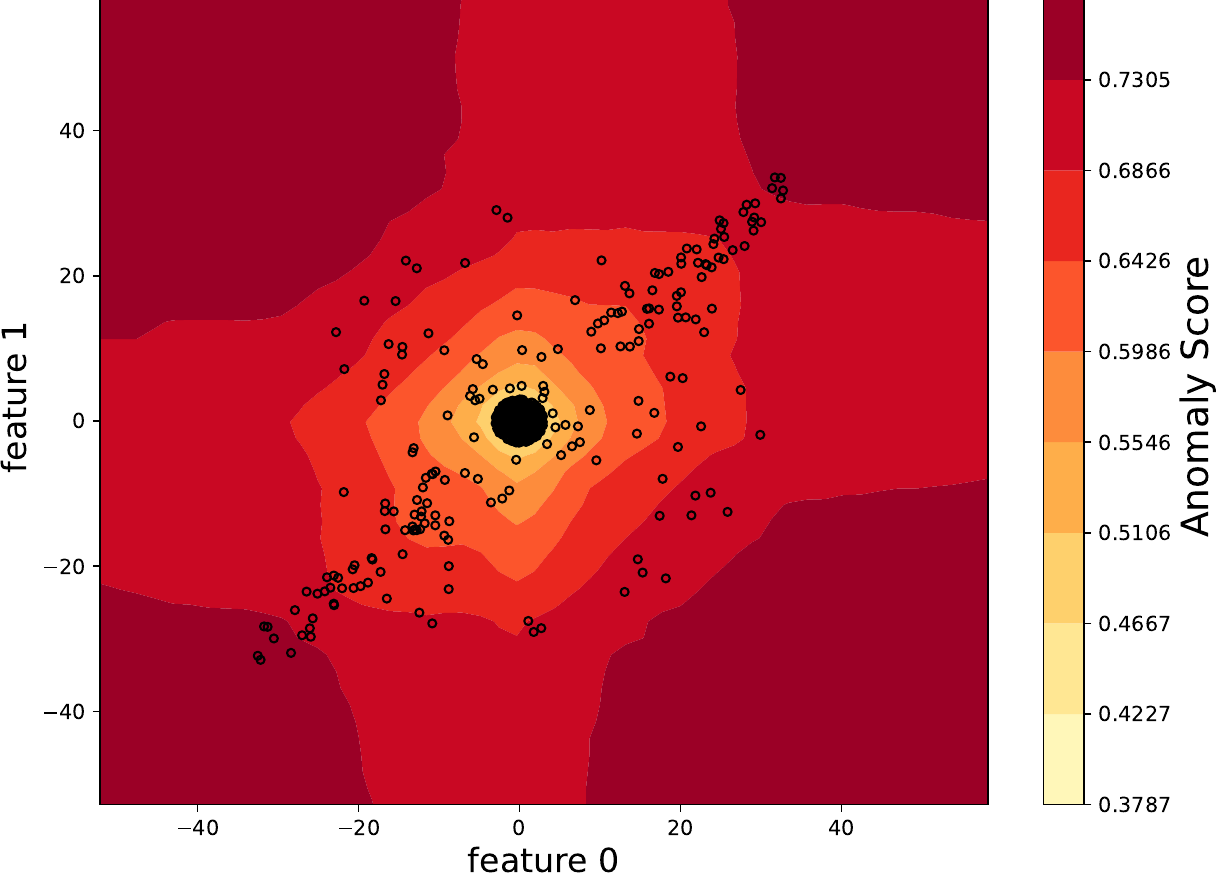}
      \caption{Anomaly Scoremap with 3 degree of freedom}
      \label{fig:scoremap3df}
    \end{subfigure}\hfil 
    \begin{subfigure}{0.33\textwidth}
    \centering
      \includegraphics[trim={0 0 3.5cm 0}, clip, width=\linewidth]{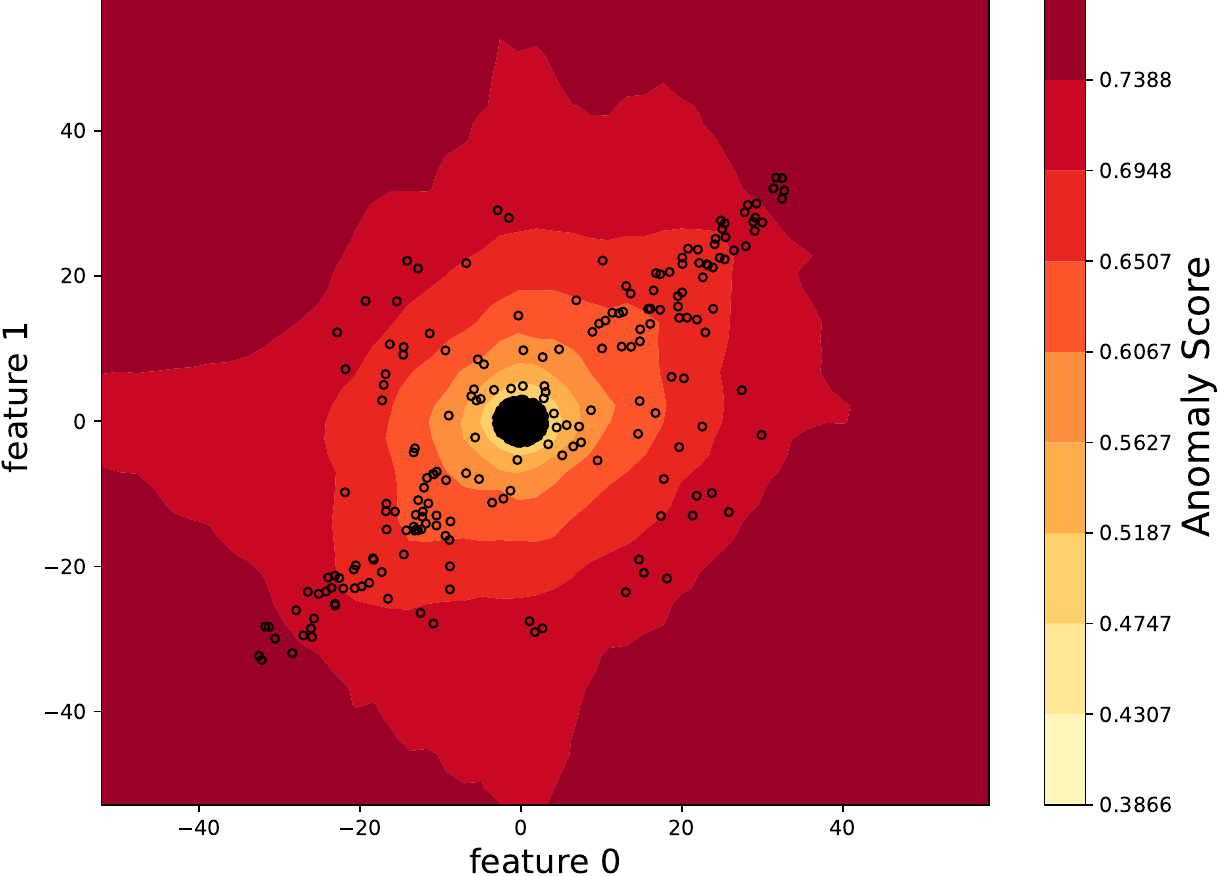}
      \caption{Anomaly Scoremap with 4 degree of freedom}
      \label{fig:scoremap4df}
    \end{subfigure}\hfil 
    \begin{subfigure}{0.33\textwidth}
    \centering
      \includegraphics[trim={0 0 3.5cm 0}, clip, width=\linewidth]{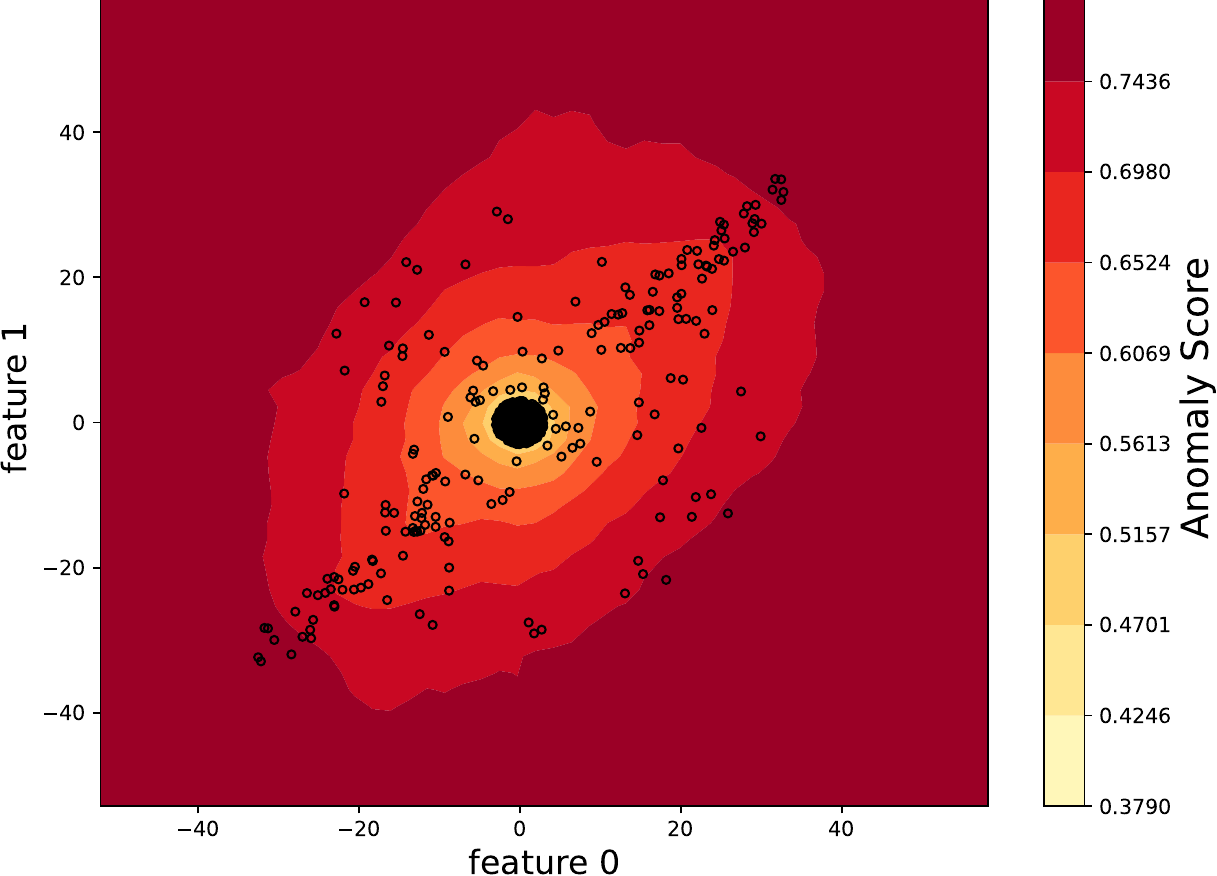}
      \caption{Anomaly Scoremap with 5 degree of freedom}
      \label{fig:scoremap5df}
    \end{subfigure}\hfil 
    \begin{subfigure}{0.33\textwidth}
    \centering
      \includegraphics[trim={0 0 3.5cm 0}, clip, width=\linewidth]{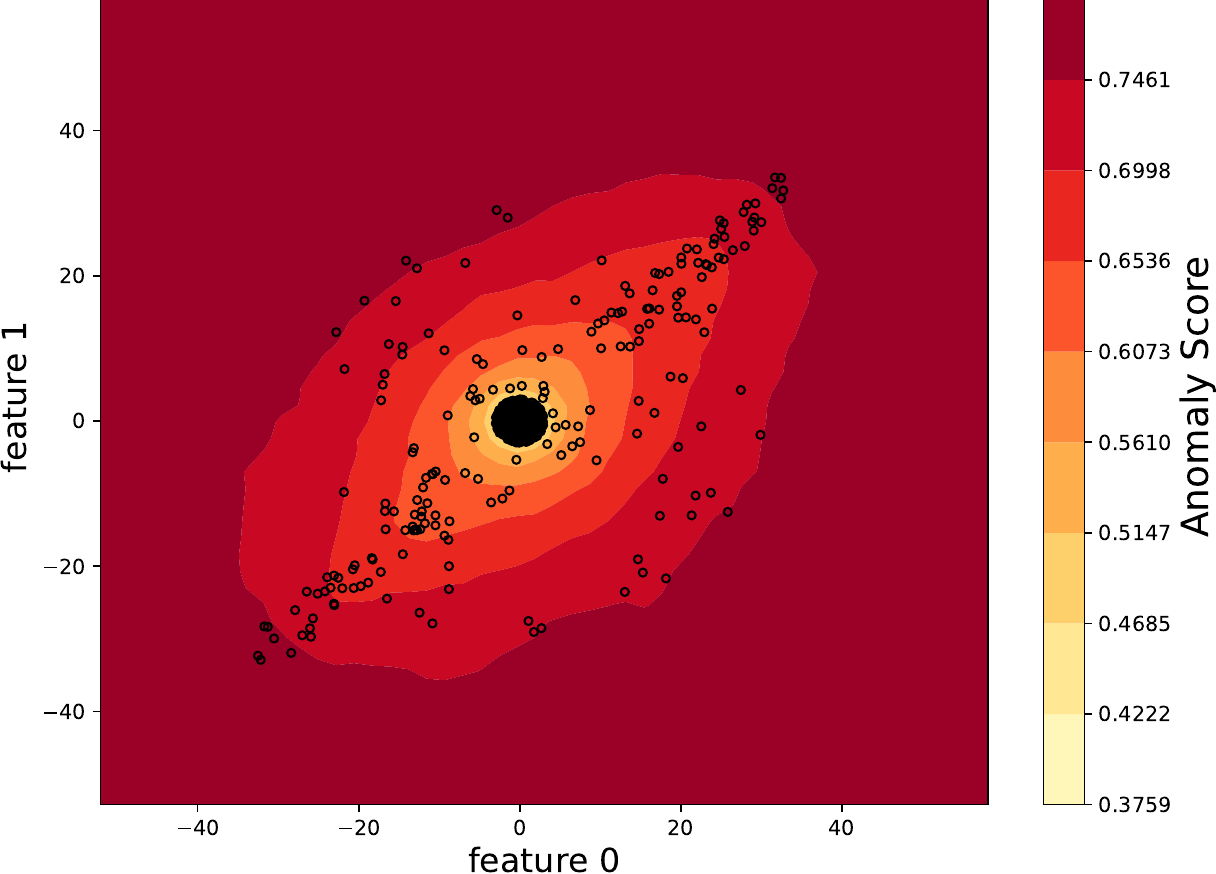}
      \caption{Anomaly Scoremap with 6 degree of freedom}
      \label{fig:scoremap6df}
    \end{subfigure}\hfil 
\caption{Scoremap resulting from adding one degree of freedom per time while searching for splitting hyperplanes. In this  way we move from the classic IF \ref{fig:scoremap1df} to the EIF \ref{fig:scoremap6df}. {The experiment is performed on the six dimensional space dataset {\ttfamily Bisect} described in Section \ref{sec:appendix-synthetic-data}.}}\label{fig:scoremap 6 IF vs EIF}
\end{figure}

\subsection{Non Depth-Based importance} \label{sec:non-depth}
In general, it is accurate to assert that the amount of feature importance attributed by a particular node $k$ to a given sample $x$, provided that the node effectively separates the samples, should be greater if the node is closer to the root of the tree, and the importance score should decrease as the node becomes deeper within the tree. This principle is derived from the same concept used to determine anomaly scores, which are higher for samples that on average are in leafs closer to the root of the tree.

Two distinct approaches exist for assigning importance scores to nodes: the one used in the \approach algorithm and the one proposed by the DIFFI algorithm. 
The DIFFI algorithm importance scores are assigned to nodes based on a weighting scheme that inversely relates the score to the depth of the node. This approach is parameterized to ensure that scores decrease as nodes become deeper within the tree.

The \approach algorithm is an alternative approach for assigning importance scores to nodes in a decision tree, which differs from the DIFFI algorithm by not using the inverse of the depth to decrease the score as the node becomes deeper in the tree. This stems from the fact that the maximum acceptable score for a node is directly linked to the number of elements it must partition. As the depth of the node increases, this number inevitably decreases. Therefore, we assert that it is redundant and potentially misleading to incorporate an additional parameter to weigh the importance score.

The underlying idea is that as the node becomes deeper in the tree, it is responsible for splitting fewer and fewer elements, and therefore, it should not be penalized by reducing its importance score because the structure of the algorithm itself decreases the importance on average after every step deeper in the tree. In contrast, a node that splits many elements closer to the root of the tree may be more important because it has a greater impact on the separation of the data, but its importance should not be guaranteed by an external factor.

The \approach algorithm adjusts the maximum score acceptable for a node based on the number of elements it needs to split, which allows for a fair comparison of the importance of nodes across different depths in the tree. This approach can help to avoid overestimating the importance of nodes that are close to the root and underestimating the importance of nodes that are deeper in the tree, a risk that the DIFFI algorithm does not take into account.

\begin{figure}[!ht] 
    \centering 
    \begin{subfigure}{0.45\textwidth}
      \includegraphics[trim={1cm 1cm 1.5cm 1.7cm}, clip, width=\linewidth]{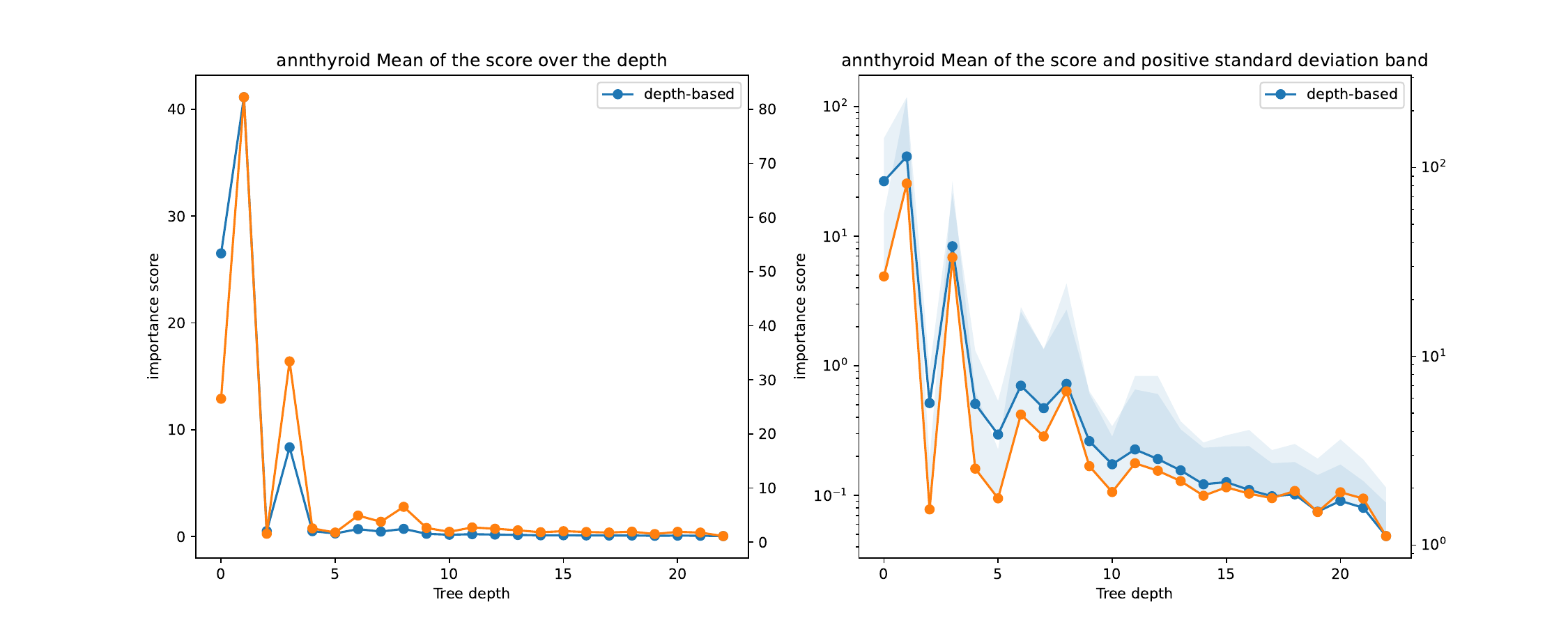}
        \caption{{\ttfamily Annthyroid}}\label{fig:depth_annthyroid}
    \end{subfigure}\hfil 
    \begin{subfigure}{0.45\textwidth}
      \includegraphics[trim={1cm 1cm 1.5cm 1.7cm}, clip, width=\linewidth]{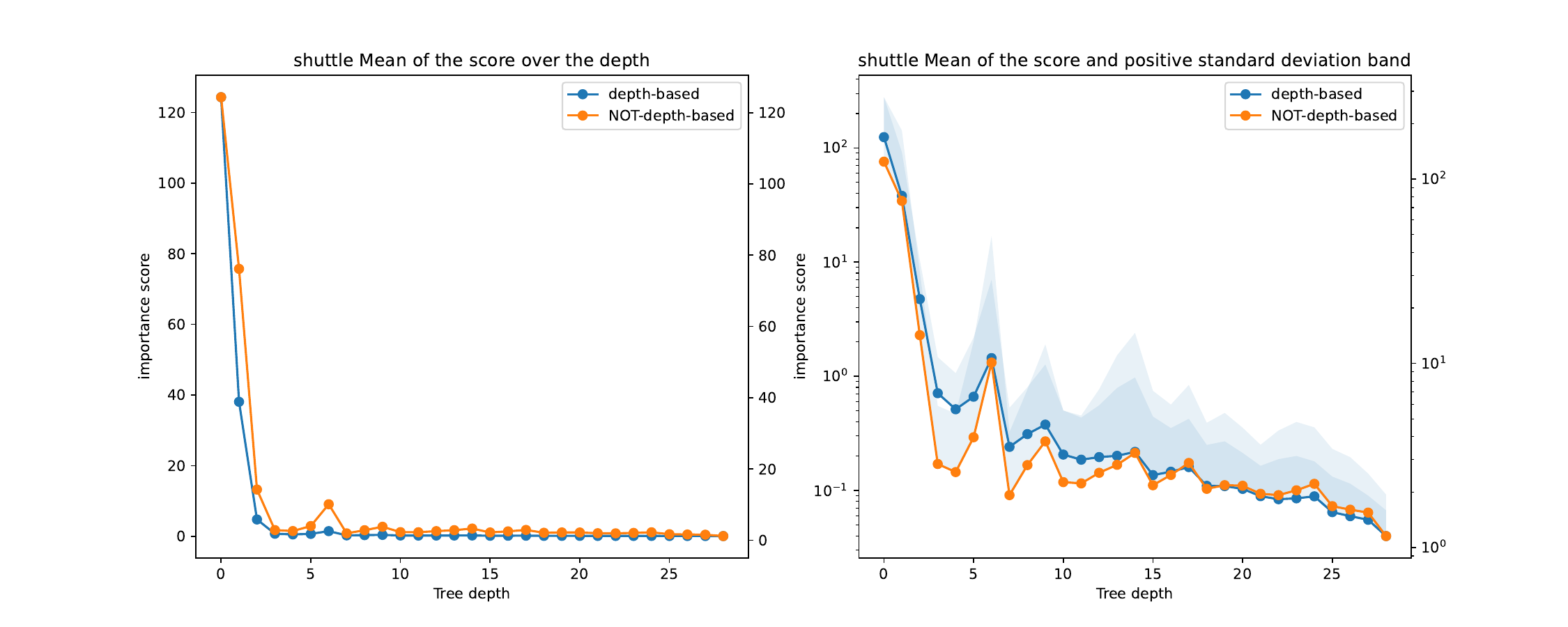}
        \caption{{\ttfamily Shuttle}}\label{fig:depth_shuttle}
    \end{subfigure}\hfil 
    \begin{subfigure}{0.45\textwidth}
      \includegraphics[trim={1cm 1cm 1.5cm 1.7cm}, clip, width=\linewidth]{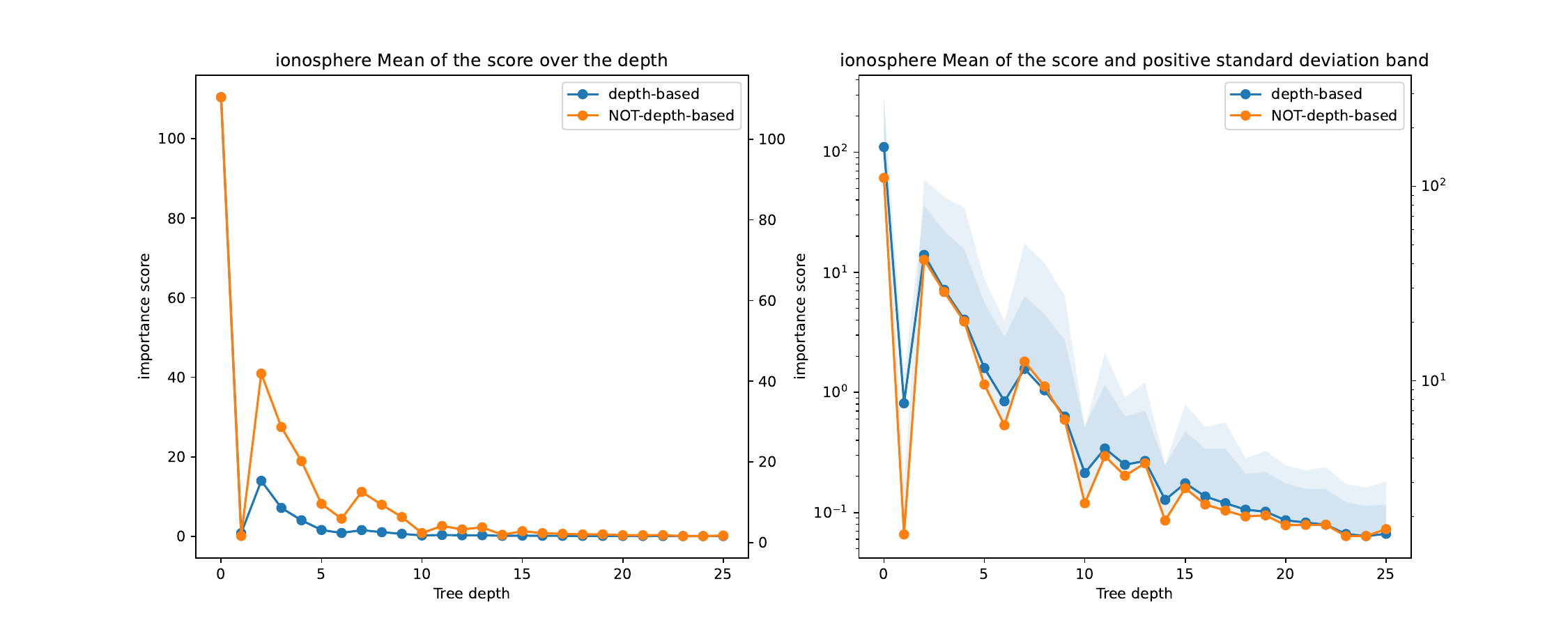}
    \caption{{\ttfamily Ionosphere}}\label{fig:depth_ionosphere}
    \end{subfigure}\hfil 
        \begin{subfigure}{0.45\textwidth}
      \includegraphics[trim={1cm 1cm 1.5cm 1.7cm}, clip, width=\linewidth]{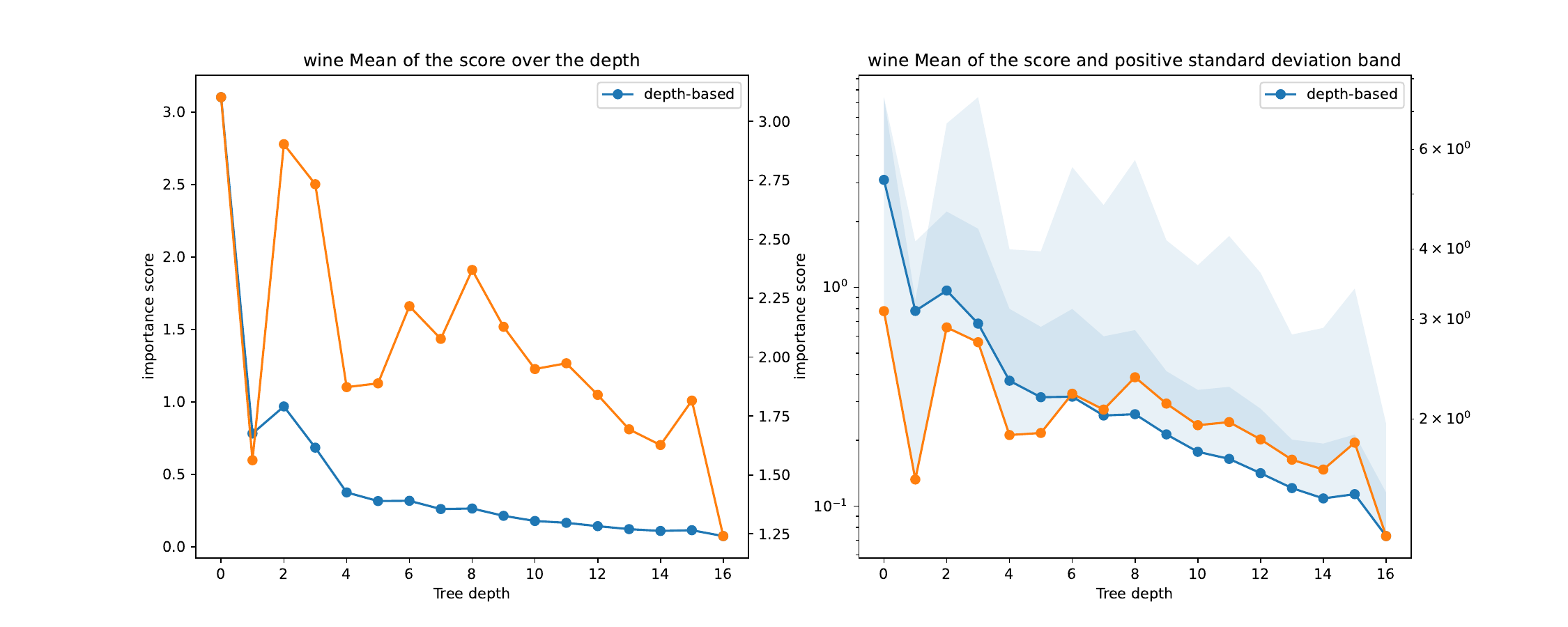}
    \caption{{\ttfamily Wine}}\label{fig:depth_wine}
    \end{subfigure}\hfil 

\caption{Analysis of the average scores of the nodes related to their depth.}\label{fig:depth}
\end{figure}

We conducted an analysis of the \approach algorithm with and without the depth parameter, in order to identify differences in the evolution of the score at different depths of the trees. To achieve this, we plotted the average score of the nodes at various levels of depth in the forests, as shown in Figure \ref{fig:depth}.

The results of this analysis indicate that the score generally decreases without the use of a depth parameter. Although the \approach algorithm without the depth parameter produced higher scores, the overall shape of the graph is still descending as nodes are evaluated deeper in the tree. This suggests that the \approach algorithm is effective in assigning importance scores to nodes, even when the depth parameter is not utilized.

In summary, our analysis suggests that the \approach algorithm is a robust method for assigning importance scores to nodes in decision trees, with or without the use of a depth parameter.

\subsection{\EIFplus: the effects of $\eta$}\label{sec:eta}

\ale{In this section the results of the ablation study on the $\eta$ hyperparameter of the \EIFplus model for the datasets covered in this Appendix are reported. }

\ale{The ablation study consists in evaluating the performance of the \EIFplus model, through the Average Precision metric, for 25 linearly spaced values between 0.5 and 5 of the parameter $\eta$, whose interpretation is given in \ref{sec:EIFplus}. }

\begin{figure}[!ht] 
    \centering 
    \begin{subfigure}{0.33\textwidth}
    \centering
      \includegraphics[width=\linewidth]{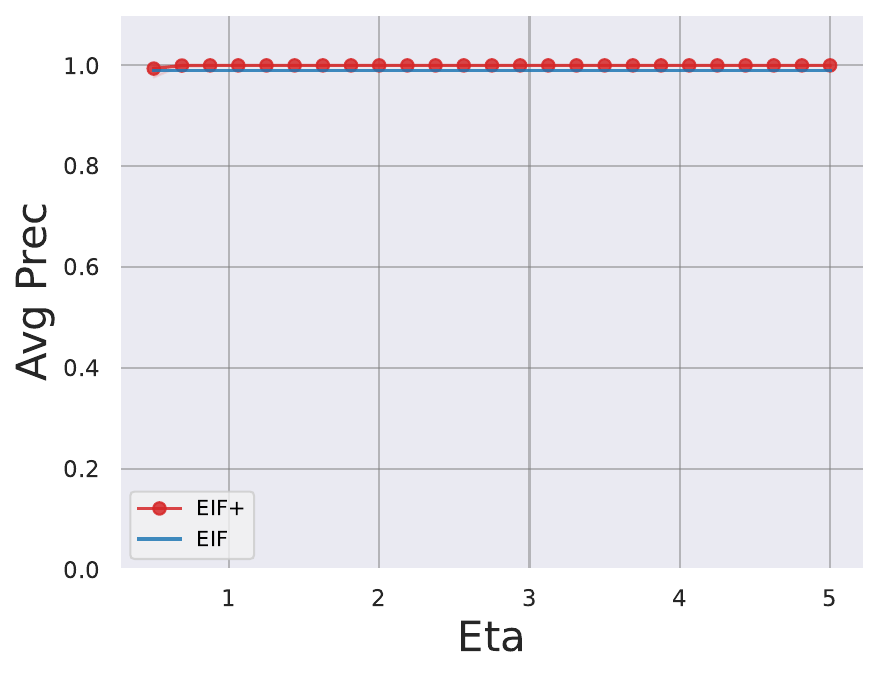}
      \caption{{\ttfamily Bisect}}
      \label{fig:ablation-Bisect}
    \end{subfigure}\hfil 
    \begin{subfigure}{0.33\textwidth}
    \centering
      \includegraphics[width=\linewidth]{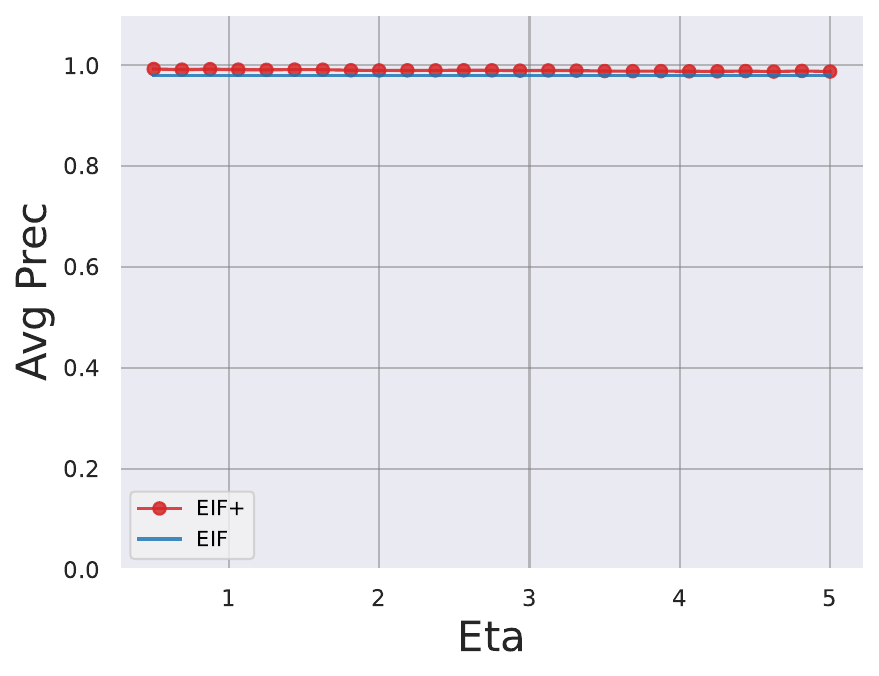}
      \caption{{\ttfamily Breastw}}
      \label{fig:ablation-Breastw}
    \end{subfigure}\hfil 
    \begin{subfigure}{0.33\textwidth}
    \centering
      \includegraphics[width=\linewidth]{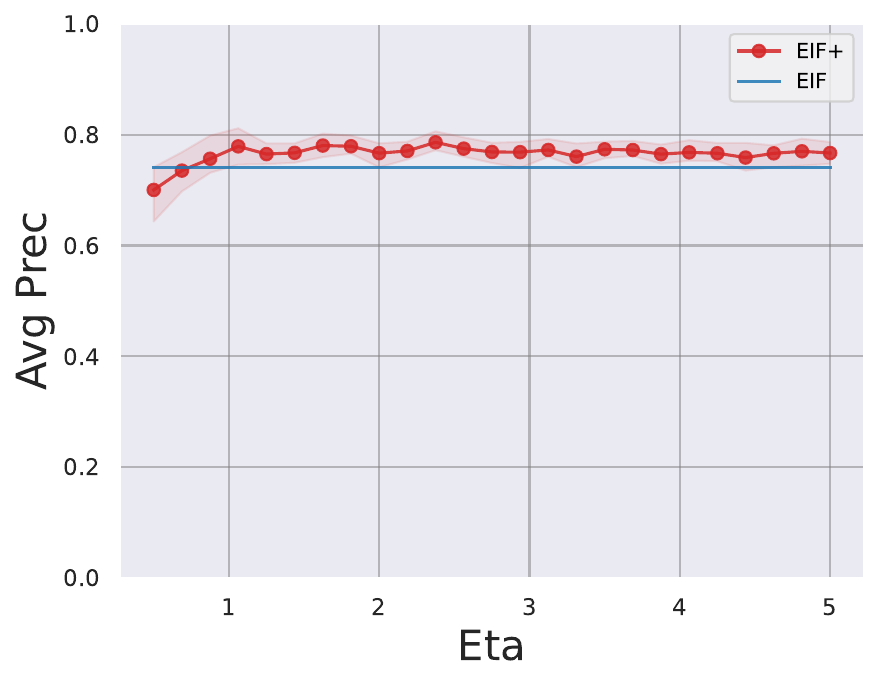}
      \caption{{\ttfamily Cardio}}
      \label{fig:ablation-Cardio}
    \end{subfigure}\hfil 
    \begin{subfigure}{0.33\textwidth}
    \centering
      \includegraphics[width=\linewidth]{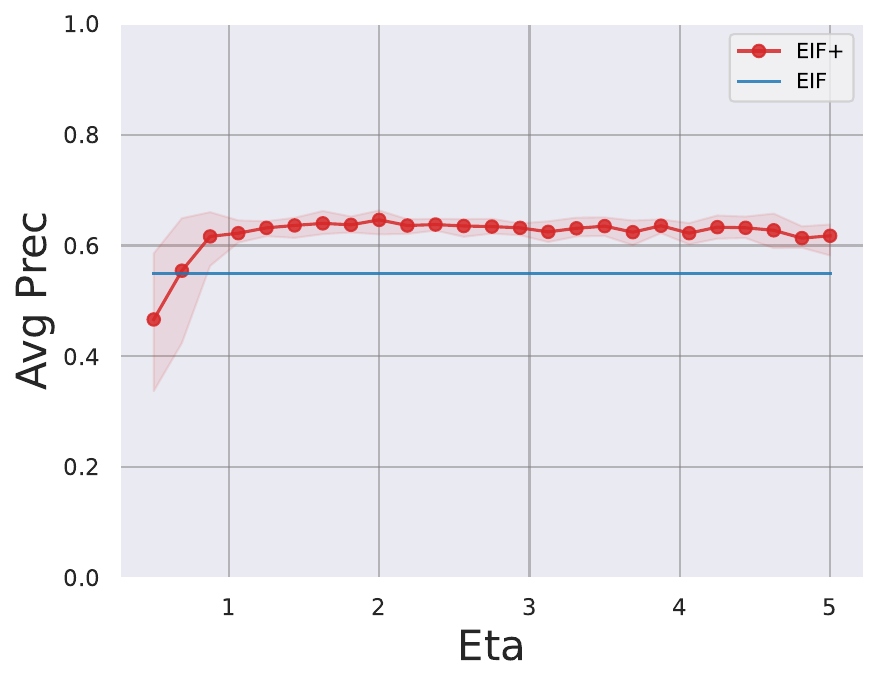}
      \caption{{\ttfamily Diabetes}}
      \label{fig:ablation-Diabetes}
    \end{subfigure}\hfil 
    \begin{subfigure}{0.33\textwidth}
    \centering
      \includegraphics[width=\linewidth]{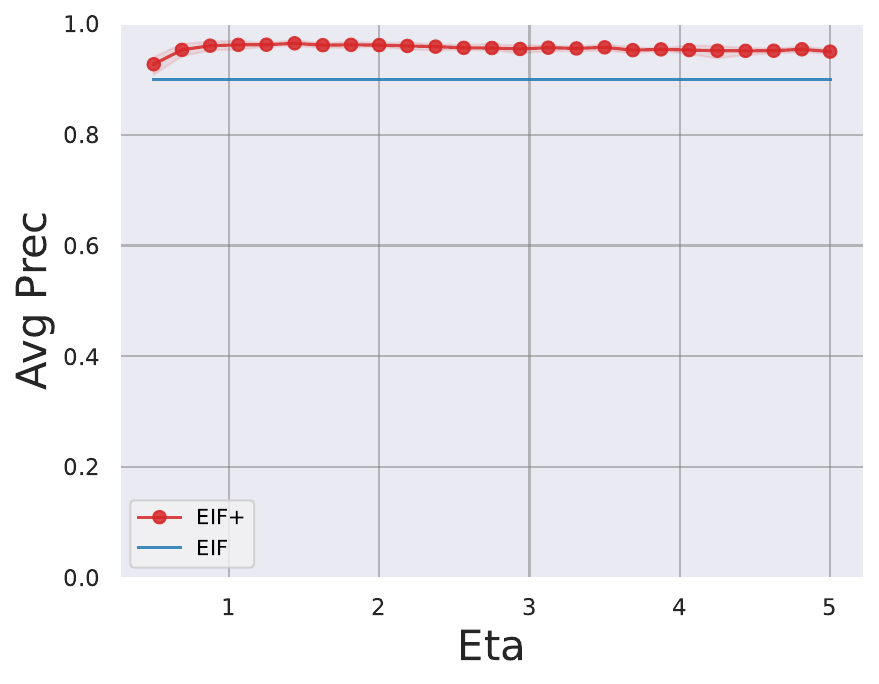}
      \caption{{\ttfamily Ionosphere}}
      \label{fig:ablation-Ionosphere}
    \end{subfigure}\hfil 
    \begin{subfigure}{0.33\textwidth}
    \centering
      \includegraphics[width=\linewidth]{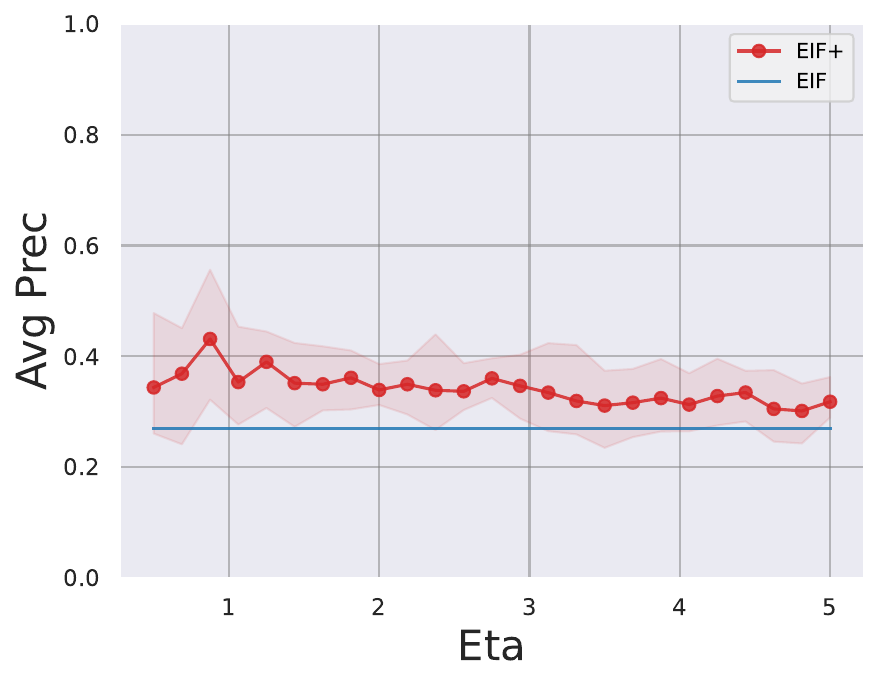}
      \caption{{\ttfamily Pendigits}}
      \label{fig:ablation-Pendigits}
    \end{subfigure}\hfil 
    \begin{subfigure}{0.33\textwidth}
    \centering
      \includegraphics[width=\linewidth]{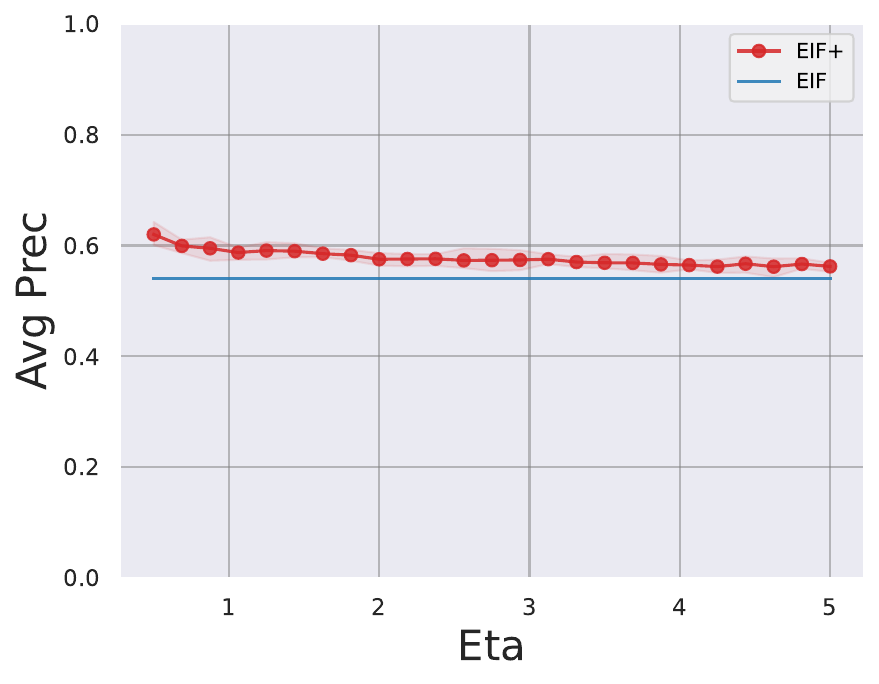}
      \caption{{\ttfamily Pima}}
      \label{fig:ablation-Pima}
    \end{subfigure}\hfil 
    \begin{subfigure}{0.33\textwidth}
    \centering
      \includegraphics[width=\linewidth]{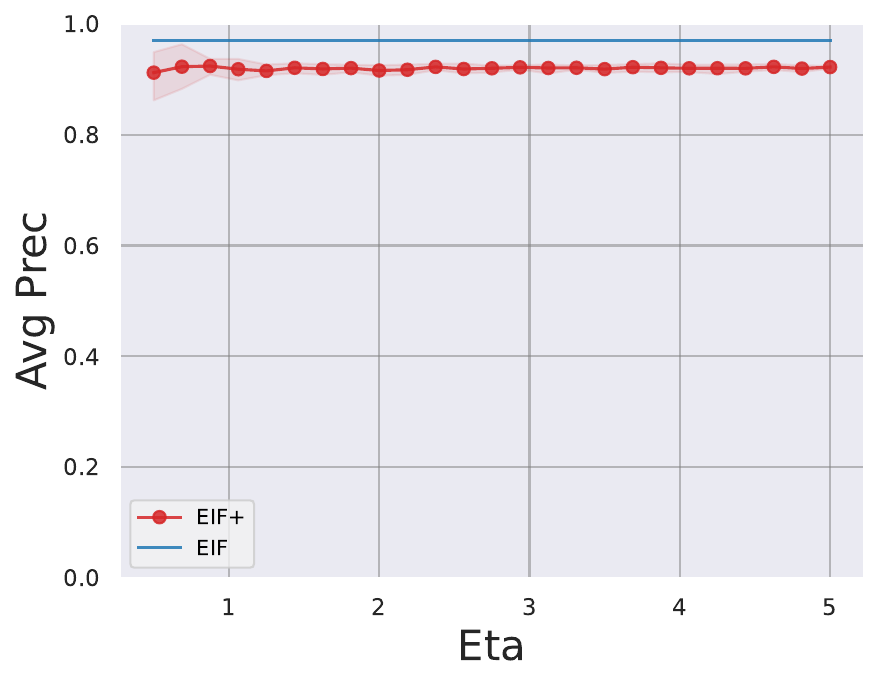}
      \caption{{\ttfamily Shuttle}}
      \label{fig:ablation-Shuttle}
    \end{subfigure}\hfil 
    \begin{subfigure}{0.33\textwidth}
    \centering
      \includegraphics[width=\linewidth]{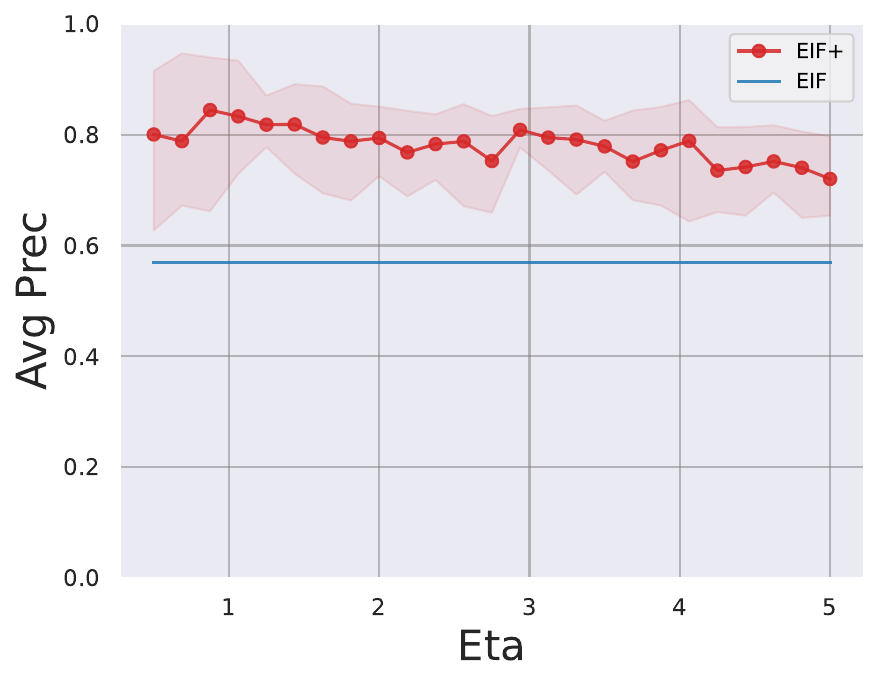}
      \caption{{\ttfamily Wine}}
      \label{fig:ablation-Wine}
    \end{subfigure}\hfil 
\caption{Exploration of how average precision changes with variations in the $\eta$ parameter.}\label{fig:ablation_appendix}
\end{figure}

\ale{After a careful observation of the plots collected in Figure \ref{fig:ablation_appendix} it is possible to conclude that for the {\ttfamily Bisect}, {\ttfamily Breastw}, {\ttfamily Ionosphere} and {\ttfamily Shuttle} the Average Precision of \EIFplus is not affected by the value of the parameter $\eta$. In fact, as it can be noticed comparing to the blue line inserted in the plot to represent the Average Precision value achieved by the EIF model, in these sets of data the improvement achieved by \EIFplus on its counterpart EIF is minimal since both models already achieve almost perfect precision scores. As a consequence the model remain unchanges independelty on the $\eta$ value since anomalies are detectable utilizing the traditional oblique cuts employed by EIF.}

\ale{For other datasets, where the data distribution is intricated and EIF-like partitions are not as effective as the novel partition approach introduced with \EIFplus, the effect of $\eta$ is more visible in the sense that there are some contained oscillations on the Average Precision values as the hyperparameter is varied. }

\ale{Concluding, comparing the diverse plots produced it is possible to infer that the optimal value for the parameter analysed in this section is highly dependent on the specific dataset structure and thus the suggested approach is to perform an hyperparameter optimization through a validation set in order to find the optimal value of $\eta$. }

\subsection{Explanations comparison through Normalized Discounted Cumulative Gain} \label{sec:ndcg_exp}

\rebuttal{In Section \ref{sec:imp-tab} different interpretation algorithm for Anomaly Detection were compared thorugh the newly introduced $AUC_{FS}$ metric. In this section we consider the employment of another quantitative metric for the evaluation of the effectiveness of different XAD methods: the Normalized Discounted Cumulative Gain (NDCG). The NDCG measure was originally introduced for the evaluation of the rank quality in the field of Information Retrieval \cite{NDCG}. In this experiment NDCG is re adapted to perform the comparison between the feature ranking produced by the computation of the Global Feature Importance scores with a dataset's ground truth feature ranking. For this reason this evaluation was conducted solely on the synthetic datasets, the only ones equipped with ground truth anomalies and feature rankings as explained in \ref{sec:Evaluation}. 

In order to obtain the NDCG score to evaluate how close a feature ranking produced by an interpretation algorithm is to the ideal one, the Discounted Cumulative Gain (DCG) of the raw GFI scores will be compared with a relevance vector. 

The DCG is computed considering the GFI values of the different features together with a discount factor that takes into account the position of a feature in the list. Considering a dataset $\mathcal{X}$ composed by $p$ features and a GFI feature vector $GFI=[I_1,\dots,I_p]$, the DCG score for dataset $\mathcal{X}$ can be defined as follows:

$$
    DCG = \sum_{i=0}^p \frac{I_i}{log_2(i+1)}
$$

Finally to obtain the NDCG score, DCG is normalized by the Ideal DCG (IDCG) (i.e. the DCG score of the relevance vector representing the ideal feature ranking):

$$
    NDCG = \frac{DCG}{IDCG}
$$

The relevance vector has to be crafted in order to represent the ground truth importance score ranking, which may not be unique. In this specific study, in fact, there are multiple correct rankings for each different synthetic datasets. Indeed, since normal features are sampled as noise their order in the feature ranking is not relevant. The aim of the evaluation through the NDCG score is to assess the ability of a XAD model to correctly rank the crafted anomalous features in the top positions. 

For example, in the {\ttfamily Xaxis} dataset an explanation can be considered effective if it is able to place Feature 0 in the first position of the ranking, while in {\ttfamily Bisect3D\_Skewed} a good interpretation is the one that places Feature 0,1 and 2 in the top 3 spots, in this precise order. For this reason the relevance vectors for these two datasets can be constructed, respectively, as follows: $v=[1,0,0,0,0,0]$, $v=[4,3,2,0,0,0]$. 

As it can be noticed from Table \ref{tab:ndcg_tab} most of the interpretation models considered are able to correctly rank the abnormal features, leading to a perfect NDCG score of 1. There are a few exceptions represented by the DIFFI and ECOD interpretability algorithms on the {\ttfamily Xaxis} dataset which obtain lower NDCG scores. This results comes without a surprise, in fact the struggle of the IF and ECOD model on the {\ttfamily Xaxis} dataset are already deeply documented in Sections \ref{sec:Evaluation} and in Tables \ref{tab:ComprehensiveConsolidated}\ref{tab:AUC_FS_tab}\ref{tab:corr_tab}
}


\begin{table}[ht!]
\caption{Normalized Discounted Cumulative Gain score for different interpretation algorithms in Scenario I (S1) and Scenario II (S2) across the synthetic datasets introduced in this study. Scores for the proposed algorithm (i.e. \approach) are enclosed in grey cells.}
\label{tab:ndcg_tab}
\centering
\resizebox{0.5\textwidth}{!}{
\begin{tabular}{|l|l|m{0.9cm}|m{0.9cm}|m{0.9cm}|m{0.9cm}|m{0.9cm}|}
\hline
\multicolumn{2}{|l|}{} & \multicolumn{5}{c|}{Synthetic}\\
\hline
\rotatebox[origin=c]{90}{Scenario} & \rotatebox[origin=c]{90}{Model} &
 \rotatebox[origin=c]{90}{{\ttfamily Xaxis}} & \rotatebox[origin=c]{90}{{\ttfamily Bisec}} & \rotatebox[origin=c]{90}{{\ttfamily Bisec3D}} & \rotatebox[origin=c]{90}{{\ttfamily Bisec6D}} & 
 \rotatebox[origin=c]{90}{{\ttfamily Bisec3D\_Skewed}}\\
\hline
\multirow{8}{*}{S1} & DIFFI & 1.0& 1.0& 1.0& 0.9& 1.0\\
& IF\_ExIFFI & \cellcolor{lightgray} 0.35& \cellcolor{lightgray} 1.0& \cellcolor{lightgray} 1.0& \cellcolor{lightgray} 0.99& \cellcolor{lightgray} 1.0\\
& EIF\_ExIFFI & \cellcolor{lightgray} 1.0& \cellcolor{lightgray} 1.0& \cellcolor{lightgray} 1.0& \cellcolor{lightgray} 0.99& \cellcolor{lightgray} 1.0\\
& \EIFplus\_ExIFFI & \cellcolor{lightgray} 1.0& \cellcolor{lightgray} 1.0& \cellcolor{lightgray} 1.0& \cellcolor{lightgray} 0.99& \cellcolor{lightgray} 1.0\\
& IF\_RF&  1.0&  1.0&  1.0&  0.99&  1.0\\
& EIF\_RF &  1.0&  1.0&  1.0&  0.99&  1.0\\
& \EIFplus\_RF &  1.0&  1.0&  1.0&  0.99&  1.0\\
& ECOD & 0.35& 0.69& 1.0& 0.99& 1.0\\
\hline
\multirow{8}{*}{S2} & DIFFI & 0.63& 0.91& 0.96& 0.67& 1.0\\
& IF\_ExIFFI & \cellcolor{lightgray} 0.35& \cellcolor{lightgray} 1.0& \cellcolor{lightgray} 1.0& \cellcolor{lightgray} 0.99& \cellcolor{lightgray} 1.0\\
& EIF\_ExIFFI & \cellcolor{lightgray} 1.0& \cellcolor{lightgray} 1.0& \cellcolor{lightgray} 1.0& \cellcolor{lightgray} 0.99& \cellcolor{lightgray} 1.0\\
& \EIFplus\_ExIFFI & \cellcolor{lightgray} 1.0& \cellcolor{lightgray} 1.0& \cellcolor{lightgray} 1.0& \cellcolor{lightgray} 0.99& \cellcolor{lightgray} 1.0\\
& IF\_RF&  1.0&  1.0&  1.0&  0.99&  1.0\\
& EIF\_RF &  1.0&  1.0&  1.0&  0.99&  1.0\\
& \EIFplus\_RF &  1.0&  1.0&  1.0&  0.99&  1.0\\
& ECOD & 0.35& 0.69& 1.0& 0.99& 1.0\\
\hline
\end{tabular}}
\end{table}

\end{document}